\documentclass[journal]{IEEEtran}
\usepackage{graphicx}
\usepackage{color}
\usepackage{cite}

\ifCLASSINFOpdf
   \usepackage{graphicx}
   \DeclareGraphicsExtensions{.pdf,.jpg,.png}
\else
\fi

\usepackage[cmex10]{amsmath}

\def\log{ { \rm log} }

\usepackage{url}

\usepackage{amsmath}
\usepackage[linesnumbered,ruled,vlined]{algorithm2e}

\begin{document}

\title{Characterness: An Indicator of Text in the Wild}

\author{Yao Li,
        Wenjing Jia,
        Chunhua Shen, Anton van den Hengel
        \thanks{Y. Li, C. Shen and A. van den Hengel
        are with the University of Adelaide, Australia.
        W. Jia is with the University of Technology, Sydney,
        Australia. }%
\thanks{W. Jia's contribution was made when visiting the University of
Adelaide.
}
\thanks{All correspondence should be addressed to C. Shen
(chhshen@gmail.com).}%
}

\markboth{Manuscript}%
{Li \MakeLowercase{\textit{et al.}}: Characterness: An Indicator of Text in the Wild}

\maketitle

\begin{abstract}
Text in an image provides vital information for interpreting its
contents, and text in a scene can aide with a variety of tasks from
navigation, to obstacle avoidance, and odometry.
Despite its value,
however, identifying general text in images remains a challenging
research problem.
Motivated by the need to consider the widely varying forms of natural
text, we propose a bottom-up approach to the problem which reflects
the `characterness' of an image region.  In this sense our approach
mirrors the move from saliency detection methods to measures of
`objectness'.
In order to measure the characterness we develop three novel cues that are
tailored for character detection, and a Bayesian method for their integration.
Because text is made up of sets of characters, we then design a Markov random
field (MRF) model so as to exploit the inherent dependencies between characters.

We experimentally demonstrate the effectiveness of our
characterness cues as well as the advantage of Bayesian multi-cue
integration.
The proposed text detector outperforms state-of-the-art methods on a few  benchmark scene text detection datasets.
We also show that our measurement of `characterness' is superior than state-of-the-art
saliency detection models when applied to the same task.
\end{abstract}

\IEEEpeerreviewmaketitle

\section{Introduction}
\label{sec:introduction}
Human beings find the identification of text in an image almost effortless, and largely involuntarily.
As a result, much important information is conveyed in this form, including navigation instructions
(exit signs, and route information, for example), and warnings (danger signs \emph {etc.}), amongst a host of others.
Simulating such an ability for machine vision system has been an active topic
in the vision and document analysis community.
Scene text detection serves as an important preprocessing step for end-to-end scene text recognition which has manifested itself in various forms, including navigation, obstacle avoidance, and odometry to name a few.
Although some breakthrough have been made, the accuracy of the state-of-the-art scene text detection algorithms still lag behind human performance on the same task.

Visual attention, or visual saliency, is fundamental to the human visual system, and alleviates the need to process the otherwise vast amounts of incoming visual data.
As such it has been a well studied problem within multiple disciplines, including cognitive psychology, neurobiology, and computer vision.
In contrast with some of the pioneering saliency detection models~\cite{itti1998model,DBLP:conf/nips/BruceT05,DBLP:conf/nips/HarelKP06,hou2007saliency,
DBLP:conf/iccv/JuddEDT09} which has achieved reasonable accuracy in predicting human eye fixation,
recent
work has focused on
\emph{salient object detection}~\cite{DBLP:conf/mm/ZhaiS06,achanta2009frequency,DBLP:conf/eccv/WeiWZ012,
DBLP:conf/iccv/FengWTZS11,cheng2011global,liu2011learning,DBLP:conf/iccv/KleinF11,
DBLP:conf/cvpr/AlexeDF10,perazzi2012saliency,shen2012unified,jiang2011automatic,
DBLP:conf/cvpr/GofermanZT10,DBLP:conf/iccv/ChangLCL11,DBLP:conf/eccv/RahtuKSH10}.
The aim of salient object detection
is to highlight the whole attention-grabbing object with well-defined boundary.
Previous saliency detection
approaches can be broadly classified into either local
~\cite{DBLP:conf/eccv/RahtuKSH10,jiang2011automatic,DBLP:conf/iccv/KleinF11}
or global~\cite{achanta2009frequency,
DBLP:conf/iccv/FengWTZS11,cheng2011global,perazzi2012saliency,shen2012unified,
DBLP:conf/cvpr/GofermanZT10} methods.
Saliency detection has manifested itself in various forms, including image retargeting~\cite{DBLP:conf/cvpr/DingXY11, sun2011scale}, image classification~\cite{DBLP:conf/cvpr/SharmaJS12}, image segmentation~\cite{DBLP:conf/iccv/WangXZH11}.
Our basic motivation is the fact that
\emph{text attracts human attention}, even when amongst
a cluttered background.
This has been shown by a range of authors including
Judd \emph{et al.}~\cite{DBLP:conf/iccv/JuddEDT09} and Cerf \emph{et al.}~\cite{cerf2009faces} who verified that humans tend to focus on text in natural scenes.
Previous work
\cite{DBLP:conf/icpr/SunLS10,
DBLP:conf/das/ShahabSDU12,DBLP:conf/das/MengS12,DBLP:conf/icdar/UchidaSKF11}
has also demonstrated that saliency detection models can be used in early stages of scene
text detection.
In \cite{DBLP:conf/icpr/SunLS10}, for example, a saliency map obtained from Itti \emph{et al.}
~\cite{itti1998model} was used to find regions of interest.
Uchida \emph{et al.}~\cite{DBLP:conf/icdar/UchidaSKF11} showed that
using
both SURF and saliency features achieved superior character recognition performance over using SURF features alone.
More recently, Shahab \emph{et al.}~\cite{DBLP:conf/das/ShahabSDU12} compared the performance of four different saliency detection models at scene text detection.
Meng and Song~\cite{DBLP:conf/das/MengS12} also adopted the saliency framework of~\cite{liu2011learning} for scene text detection.

While the aforementioned approaches
have demonstrated that saliency detection models facilitate scene text detection, they share
a common inherent limitation, which is that they are distracted by other salient objects in the scene.
The approach we propose here differs from these existing methods in that
we propose a text-specific saliency detection model (\emph{i.e.} a characterness model) and demonstrate its robustness when applied to scene text detection.

Measures of  `objectness' \cite{DBLP:conf/cvpr/AlexeDF10} have
built upon the saliency detection in order  to
identify windows within an image that are likely to contain an
object of interest.
Applying an objectness measure in a sliding-window approach thus allows the identification of interesting objects, rather than regions.
This approach has been shown to be very useful as a pre-processing step for a wide range of problems including occlusion boundary detection~\cite{DBLP:journals/ijcv/HoiemEH11}, semantic segmentation~\cite{DBLP:conf/cvpr/ArbelaezHGGBM12}, and training
object class detectors~\cite{DBLP:conf/cvpr/PrestLCSF12}.

We propose here a similar approach to text detection, in that we
seek to develop a method which is capable of identifying
individual, bounded units of text, rather than areas with
text-like characteristics.
The unit in the case of text is the
character, and much like the `object', it has a particular set of
characteristics, including a closed boundary.
In contrast to the objects of~\cite{DBLP:conf/cvpr/AlexeDF10}, however, text is made up of a set of inter-related characters.
Therefore, effective text detection should be able to compensate for, and exploit these dependencies between
characters.
The object detection method of~\cite{DBLP:conf/cvpr/AlexeDF10} is similar to that proposed here in as much as it is based on a Bayesian framework combining a number of visual cues, including one which represents the boundary of the object, and one which measures the degree to which a putative object differs from the background.

In contrast to saliency detection algorithms
which either attempt to identify pixels or rectangular image windows that
attract the eye,
our focus here is instead on identifying individual characters within non-rectangular regions.
As characters represent the basic units of text, this renders our method
applicable in a wider variety of circumstances
than saliency-based paragraph detection,
yet more specific.
When integrating the three new characterness cues developed, instead of simple
linear combination, we use a Bayesian approach to model the joint
probability that a candidate region represents a character.
The probability distribution of cues on both characters and non-characters are obtained from training samples.
In order to model and exploit the inter-dependencies between characters we use
the graph cuts~\cite{DBLP:journals/pami/BoykovVZ01} algorithm to carry out inference over an MRF designed for the purpose.

To the best our of knowledge, we are the first to present a saliency detection model which
measures the
characterness of image regions.
This text-specific saliency detection model is less likely to be distracted by other objects which are
usually considered as salient in general saliency detection models.
Promising experimental results on benchmark datasets
demonstrate that our characterness approach outperforms the state-of-the-art.

\section{Related work}
\label{sec:state-of-the-art}

\subsection{Scene text detection}

Existing scene text detection approaches generally fall into one of three categories, namely,
texture-based approaches, region-based approaches, and hybrid
approaches.

Texture-based
approaches~\cite{DBLP:conf/icdar/LeeLLYK11,DBLP:conf/cvpr/ChenY04}
extract distinct texture properties from sliding windows, and use a
classifier (\emph{e.g.}, AdaBoost~\cite{DBLP:conf/icdar/LeeLLYK11,DBLP:conf/cvpr/ChenY04})
to detect individual instances.
Some widely used features include
Histograms of Gradients (HOGs), Local Binary Patterns (LBP), Gabor
filters and wavelets.
Foreground regions on various scales are then merged to generate final text regions.
Yet, there is something profoundly unsatisfying about texture-based approaches.
Firstly, the brute force nature of window classification is not particularly appealing.
Its computational complexity is proportional to the product of the number of scales.
Secondly, it is unclear which
features have contributed most for removing various interfering
backgrounds.
Thirdly, an extra post processing step, \emph{i.e.} text extraction, is needed before text recognition.

Region-based approaches~\cite{DBLP:conf/cvpr/EpshteinOW10,DBLP:conf/icpr/ZhangK10a,DBLP:conf/accv/ZhangK10,DBLP:journals/tip/YiT11,
DBLP:conf/icip/ChenTSCGG11,DBLP:conf/icpr/LiL12,DBLP:conf/icip/Li13}, on the other hand, first extract potential characters through edge detection~\cite{DBLP:conf/cvpr/EpshteinOW10,DBLP:conf/icpr/ZhangK10a,DBLP:conf/accv/ZhangK10}, color clustering~\cite{DBLP:journals/tip/YiT11} or Maximally Stable Extremal Region (MSER) detection~\cite{DBLP:conf/icip/ChenTSCGG11,DBLP:conf/icpr/LiL12}.
After that, low level features based on geometric and
shape constraints are used to reject non-characters.
As a final step, remaining regions are clustered into lines through
measuring the similarities between them.
A typical example of region-based approaches was proposed by
Epshtein \emph{et al.}~\cite{DBLP:conf/cvpr/EpshteinOW10}, where a local
image operator, called Stroke Width Transform (SWT), assigned each
pixel with the most likely stroke width value, followed by a
series of rules in order to remove non-characters.
All region-based approaches, although not particularly computationally
demanding, involve many parameters that need to be tuned manually.
An advantage of region-based approaches is that their results can
be sent to Optical Character Recognition (OCR) software for
recognition directly, without the extra text extraction step.

Hybrid approaches~\cite{DBLP:conf/accv/NeumannM10,DBLP:conf/cvpr/NeumannM10,
DBLP:journals/tip/PanHL11,DBLP:conf/cvpr/Yao,DBLP:journals/tip/YiT12,koo2013scene}
are a combination of texture-based and region-based approaches.
Usually, the initial step is to extract potential characters,
which is the same as region-based approaches.
Instead of utilizing low level cues, these methods extract
features from regions and exploit classifiers to decide whether a
particular region contains text or not, and this is considered as a
texture-based step.
In~\cite{DBLP:conf/cvpr/Yao}, potential characters were extracted
by the SWT initially.
To reject non-characters two random forest classifiers
were trained using two groups of features
(component and chain level) respectively.

\subsection{Saliency detection}
The underlying hypothesis of existing saliency detection algorithms is the same:
\emph {the contrast between salient object and background is high}.
Contrast can be computed via various features, such as intensity~\cite{DBLP:conf/iccv/KleinF11}, edge density~\cite{DBLP:conf/cvpr/AlexeDF10},
orientation~\cite{DBLP:conf/iccv/KleinF11}, and most commonly color~\cite{DBLP:conf/cvpr/AlexeDF10,DBLP:conf/eccv/RahtuKSH10,
DBLP:conf/iccv/KleinF11,liu2011learning,jiang2011automatic,DBLP:conf/iccv/FengWTZS11,
cheng2011global,DBLP:conf/iccv/ChangLCL11,DBLP:conf/cvpr/GofermanZT10,perazzi2012saliency,
shen2012unified,DBLP:conf/eccv/WeiWZ012}.
The measurement of contrast also varies, including discrete form of Kullback-Leibler divergence~\cite{DBLP:conf/iccv/KleinF11},
intersection distance~\cite{DBLP:conf/iccv/FengWTZS11},
$\chi^2$ distance~\cite{DBLP:conf/cvpr/AlexeDF10,liu2011learning,DBLP:conf/iccv/ChangLCL11,
jiang2011automatic}, Euclidean distance~\cite{DBLP:conf/cvpr/GofermanZT10}.
As no prior knowledge about the size of salient objects is available,
contrast is computed at multiple scales in some methods~\cite{DBLP:conf/cvpr/GofermanZT10,DBLP:conf/eccv/RahtuKSH10,
jiang2011automatic,DBLP:conf/iccv/KleinF11,liu2011learning}.
To make the final saliency map smoother, spatial information is also commonly adopted
in the computation of contrast~\cite{DBLP:conf/cvpr/GofermanZT10,cheng2011global,DBLP:conf/iccv/FengWTZS11,
jiang2011automatic,perazzi2012saliency}.

The large amount of literature on saliency detection can be broadly classified into two classes in terms of the scope of contrast computed.
\emph{Local methods}
~\cite{DBLP:conf/eccv/RahtuKSH10,jiang2011automatic,DBLP:conf/iccv/KleinF11}
estimate saliency of an image patch according to its contrast against its surrounding patches.
They assume that patches whose contrast values are higher against their surrounding counterparts should
be salient.
As computing local contrast at one scale tends to only highlight boundaries of the salient object rather than the whole object, local methods are
always performed at multiple scales
\begin{figure*}[t]

\centering
\begin{tabular}{@{}c}
\includegraphics[width=1\linewidth]{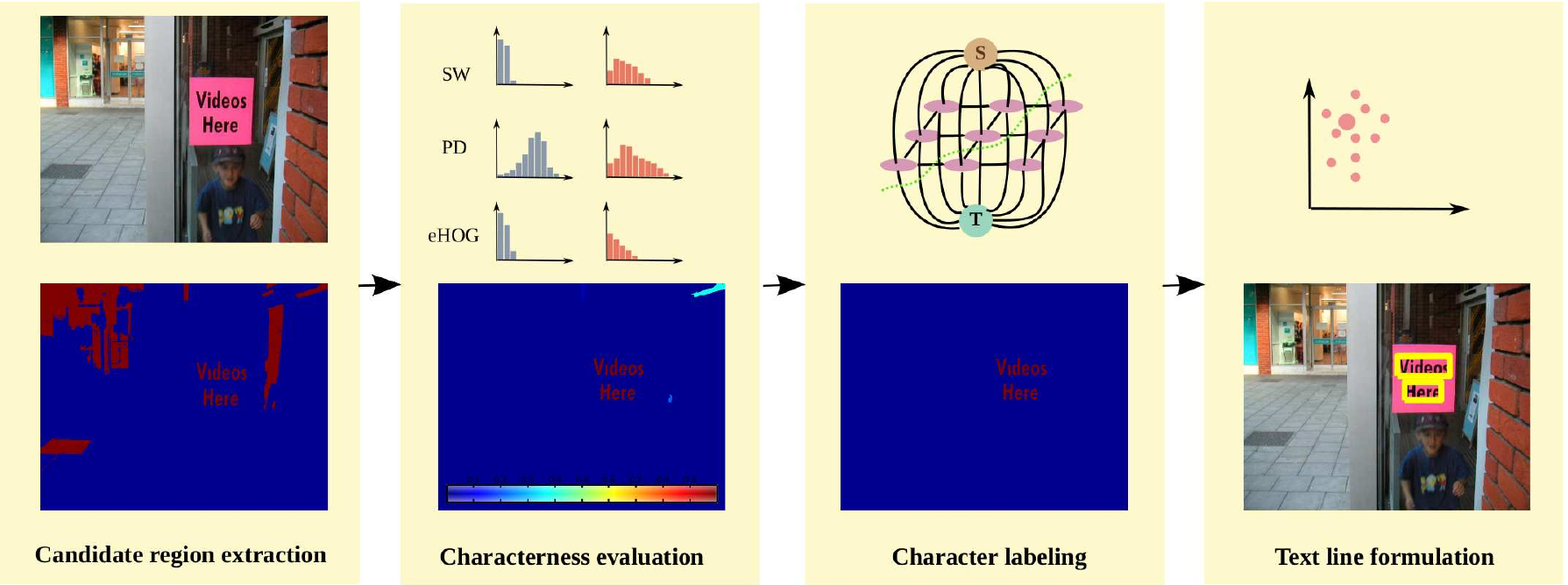}\\
\end{tabular}
\vspace{-3mm}
\caption{Overview of our scene text detection approach.
The characterness model consists of the first two phases.
}
\label{fig:pipeline}
\end{figure*}

\emph{Global methods},
e.g., \cite{hou2007saliency,achanta2009frequency,
DBLP:conf/iccv/FengWTZS11,cheng2011global,perazzi2012saliency,shen2012unified,
DBLP:conf/cvpr/GofermanZT10}, on the other hand, take the entire image into account when estimating saliency of a particular patch.
The underlying hypothesis is that globally rare features correspond to high saliency.
Color contrast over the entire image is computed in~\cite{cheng2011global}.
Shen and Wu~\cite{shen2012unified} stacked features extracted from all patches
into a matrix and then solved a low-rank matrix recovery problem.
Perazzi \emph{et al.}~\cite{perazzi2012saliency} showed that global saliency can be estimated by high-dimensional Gaussian filters.

\subsection{Contributions}
Although previous work~\cite{DBLP:conf/icpr/SunLS10,
DBLP:conf/das/ShahabSDU12,DBLP:conf/das/MengS12,DBLP:conf/icdar/UchidaSKF11}
has demonstrated that existing saliency detection models can facilitate scene text detection,
none of them has designed a saliency detection model tailored for scene text detection.
We argue that adopting existing saliency detection models directly to scene text detection~\cite{DBLP:conf/icpr/SunLS10,
DBLP:conf/das/ShahabSDU12,DBLP:conf/das/MengS12,DBLP:conf/icdar/UchidaSKF11} is inappropriate, as
general saliency detection models are likely to be distracted by non-character objects in the scene that are also salient.
In summary, contributions of our work comprise the following.
\begin{enumerate}
\item We propose a
text  detection
model which reflects the `characterness' (\emph{i.e.} the probability of
representing
a character) of image regions.
\item We design
an energy-minimization approach to character labeling, which encodes both individual characterness and pairwise similarity in a unified framework.
\item We evaluate ten state-of-the-art saliency detection models for the measurement of `characterness'.
To the best of our knowledge, we are the first to evaluate state-of-the-art saliency detection models for reflecting `characterness' in this large quantity.
\end{enumerate}

\section{overview}

Fig.~\ref{fig:pipeline} shows an overview of the proposed scene text detection approach.
Specifically, Sec.~\ref{sec:characterness} describes the characterness model, in which
perceptually homogeneous regions are extracted by a
modified
MSER-based region detector. %
Three novel
characterness cues are then computed, each of which independently
models the probability of the region forming a character (Sec.~\ref{subec:CharacternessEvaluation}).
These cues are then fused in a Bayesian
framework, where Naive Bayes is used to model the
joint probability.
The posterior probability reflects the `characterness' of the corresponding image patch.

In order to consolidate the characterness responses
we design a
character labeling method
in Sec. \ref{subsec:MRF}.
An MRF, minimized by graph
cuts~\cite{DBLP:journals/pami/BoykovVZ01}, is
used to combine
evidence from multiple per-patch characterness estimates into evidence for a single character or
compact group of characters .
Finally, verified characters are grouped to readable text lines via a clustering scheme (Sec. \ref{subsec:grouping}).

Two phases of experiments are conducted separately
in order to evaluate the
characterness model and scene text detection approach as a whole.
In the first phase (Sec. \ref{sec:CharacternessEvaluation}), we compare the proposed characterness model with ten state-of-the-art saliency detection algorithms on the characterness evaluation task, using evaluation criteria typically adopted in saliency detection.
In the second phase (Sec. \ref{sec:SceneTextDetectionEvaluation}), as in conventional scene text detection algorithms, we use
the  bounding boxes of detected text lines in order to
compare against
state-of-the-art scene text detection approaches.

\section{The Characterness Model}
\label{sec:characterness}
\subsection{Candidate region extraction}
\label{subsec:eMSER}
MSER~\cite{DBLP:conf/bmvc/MatasCUP02} is an effective region detector which
has been applied in various vision tasks, such as tracking
~\cite{DBLP:conf/cvpr/DonoserB06}, image matching~\cite{DBLP:conf/iccv/ForssenL07}, and scene text detection~\cite{DBLP:conf/accv/NeumannM10,DBLP:conf/icdar/NeumannM11,
DBLP:conf/icip/ChenTSCGG11,DBLP:conf/accv/Tsai,koo2013scene} amongst others.
Roughly speaking, for a gray-scale image, MSERs are those which have a boundary which remains relatively unchanged
over a set of different intensity thresholds.
The MSER
detector is thus particularly well suited to identifying
regions with almost uniform intensity surrounded by contrasting background.

\begin{figure}[t]
\begin{center}
\begin{tabular}{@{}c@{}c@{}c}
\includegraphics[width=0.25\linewidth]{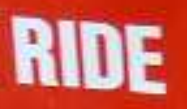} \ &
\includegraphics[width=0.25\linewidth]{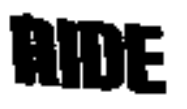} \ &
\includegraphics[width=0.25\linewidth]{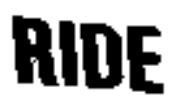} \ \\
\includegraphics[width=0.25\linewidth]{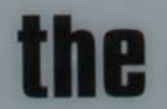} \ &
\includegraphics[width=0.25\linewidth]{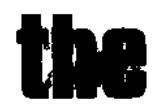} \ &
\includegraphics[width=0.25\linewidth]{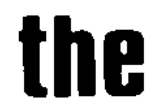} \ \\
{(a) Original text} & {(b) Original MSER} & {(c) Our results}\ \\
\end{tabular}
\end{center}
\vspace{-3mm}
\caption{Cases that the original MSER fails to extract the characters while the modified eMSER succeeds. }
\label{fig:MSER}
\end{figure}

For the task of scene text detection, although the original  MSER algorithm is able to
detect characters in most cases, there are some characters that are either missed or incorrectly connected (Fig. \ref{fig:MSER} (b)).
This
tends to degrade the performance of the following steps in the scene text detection algorithms.
To address this problem, Chen \emph{et
al.}~\cite{DBLP:conf/icip/ChenTSCGG11} proposed to prune out MSER pixels
which were located outside the boundary detected by Canny edge
detector.
Tsai \emph{et al.}~\cite{DBLP:conf/accv/Tsai} performed judicious parameter selection and multi-scale analysis of MSERs.
Neumann and Matas extended MSER to MSER++~\cite{DBLP:conf/icdar/NeumannM11}
and later Extremal Region (ER)~\cite{DBLP:conf/cvpr/NeumannM10}.
In this paper, we use the edge-preserving MSER algorithm from our earlier work~\cite{DBLP:conf/icip/Li13} (\emph{c.f.} Algorithm \ref{algo:eMSER}).

\textbf{Motivation.} As illustrated in some previous work~\cite{DBLP:journals/ijcv/MikolajczykTSZMSKG05,DBLP:conf/iccv/ForssenL07}, the MSER detector is sensitive to blur.
We have observed through empirical testing that
this may be attributed to the
large quantities of \emph{mixed pixels} (pixels lie between dark background and bright regions, and \emph{vice versa}) present along
character boundaries.
We notice that these mixed pixels usually have larger gradient
amplitude than other pixels.
We thus propose to incorporate the gradient amplitude so as to produce
edge-preserving MSERs (see Fig.~\ref{fig:MSER}(c)).
\begin{algorithm}
\DontPrintSemicolon %
\KwIn{A color image, and required parameters}
\KwOut{Regions contain characters and non-characters}
Convert the color image to an intensity image $I$. \;
Smooth $I$ using a guided filter ~\cite{DBLP:conf/eccv/HeST10}. \;
Compute the gradient amplitude map $\nabla I$, then normalize it to $[0,255]$. \;
Get a new image $I^* = I + \gamma \nabla I$ (resp. $I^* = I - \gamma \nabla I$). \;
Perform MSER algorithm on $I^*$ to extract dark regions on the bright background (resp. bright regions on the dark background).\;
\caption{Edge-preserving MSER (eMSER)}
\label{algo:eMSER}
\end{algorithm}

\subsection{Characterness evaluation}
\label{subec:CharacternessEvaluation}
\subsubsection{Characterness cues}
Characters attract human attention
because their appearance differs from that of their
surroundings.
Here, we propose three novel cues to measure the unique properties of characters.
\vspace{3mm}
\newline
\textbf{Stroke Width (SW).}
Stroke width has been a widely exploited feature for text
detection~\cite{DBLP:conf/cvpr/EpshteinOW10,DBLP:conf/accv/ZhangK10,
DBLP:conf/cvpr/Yao,DBLP:journals/tip/YiT12}.
In particular, SWT~\cite{DBLP:conf/cvpr/EpshteinOW10} computes the
length of a straight line between two edge pixels in the
perpendicular
direction,
which is
used
as a preprocessing step for later algorithms~\cite{DBLP:conf/cvpr/Yao,DBLP:conf/bmvc/Ali12,DBLP:conf/accv/Pan12}.
In~\cite{DBLP:journals/tip/YiT12}, a stroke is defined as a connected image region with uniform color and half-closed boundary.
Although this assumption is not supported by many uncommon typefaces, stroke width remains a valuable cue.

\begin{figure}[t]
\begin{center}
\begin{tabular}{@{}c@{}c@{}c}
\includegraphics[width=0.3\linewidth]{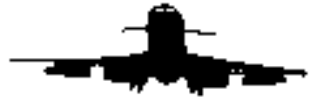} \ &
\includegraphics[width=0.3\linewidth]{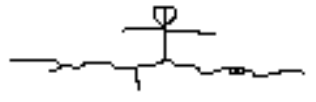} \ &
\includegraphics[width=0.3\linewidth]{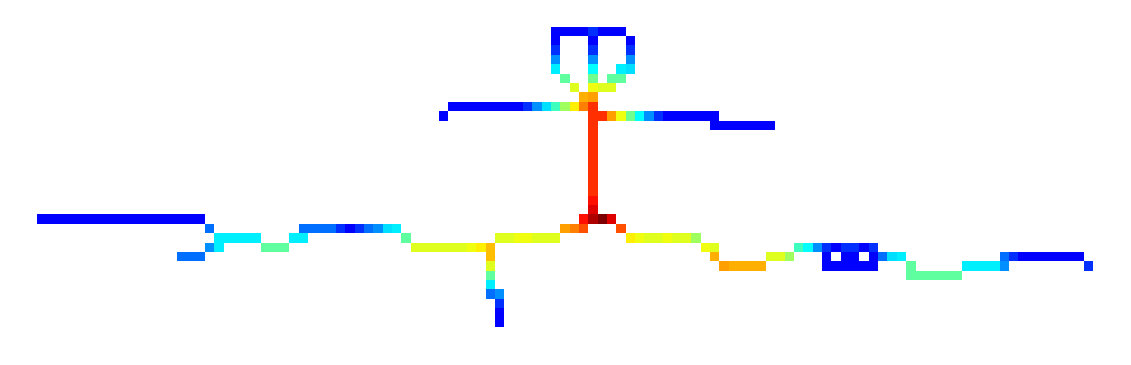} \ \\
\vspace{0.5mm}
\includegraphics[width=0.3\linewidth]{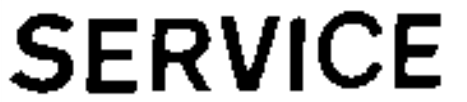} \ &
\includegraphics[width=0.3\linewidth]{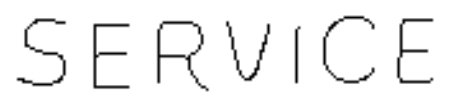} \ &
\includegraphics[width=0.3\linewidth]{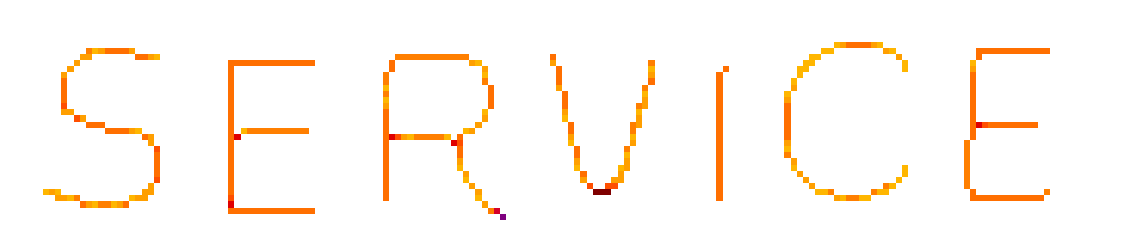} \ \\
{(a) Detected regions} & {(b) Skeleton} & {(c) Distance transform}\ \\
\end{tabular}
\end{center}
\vspace{-3mm}
\caption{Efficient stroke width computation~\cite{DBLP:conf/icpr/LiL12} (best viewed in color). Note the color variation of non-characters and characters on (c). Larger color variation indicates larger stroke width variance. }
\label{fig:strokewidth}
\end{figure}

Based on the efficient stroke width computation method we have developed earlier~\cite{DBLP:conf/icpr/LiL12} (\emph{c.f.} Algorithm \ref{algo:stroke width}), the stroke width cue of region $r$ is defined as:
\begin{equation}
{\rm{SW}}(r) = \frac{{\rm Var}(l)}{{\rm E}(l)^2} ,
\end{equation}
where is ${\rm E}(l)$ and ${\rm Var}(l)$ are stroke width mean and variance (\emph{c.f.} Algorithm \ref{algo:stroke width}).
In Fig.~\ref{fig:strokewidth} (c),
we use color to visualize the stroke width of
exemplar characters and non-characters, where larger
color variation indicates larger stroke width variance and {\it vice
versa}.
\begin{algorithm}
\DontPrintSemicolon %
\KwIn{A region $r$}
\KwOut{Stroke width mean ${\rm E}(l)$ and variance ${\rm Var}(l)$ }
Extract the skeleton $S$ of the region. \;
For each pixel $p \in S$, find its shortest path to the region boundary via distance transform. The corresponding length $l$ of the path is defined as the stroke width. \;
Compute the mean ${\rm E}(l)$ and variance ${\rm Var}(l)$. \;
\caption{Efficient computation of stroke width}%
\label{algo:stroke width}
\end{algorithm}
\vspace{3mm}
\newline
\textbf{Perceptual Divergence (PD).} As stated in Sec. \ref{sec:state-of-the-art}, color contrast is a widely adopted measurement of saliency. %
For the task of scene text detection, we observed that, in order to ensure reasonable readability of text to a human, the color of text in natural scenes is typically distinct from that of the surrounding area.
Thus, we propose the PD cue to measure the perceptual divergence of a region $r$ against its surroundings, which is defined as:
\begin{equation}
\label{eq:PD}
{\rm PD}(r)=\sum_{R,G,B}\sum_{j=1}^b h_j(r)\log\frac{h_j(r)}{h_j(r^*)},
\end{equation}
where the term $\int_xp(x)\log\frac{p(x)}{q(x)}$ is the Kullback-Leibler divergence (KLD) measuring the dissimilarity of two probability distributions in the information theory.
Here we take advantage of its discrete form~\cite{DBLP:conf/iccv/KleinF11},
and replace the probability distributions $p(x)$, $q(x)$ by the color histograms
of two regions $h(r)$ and $h(r^*)$ ($r^*$ denotes the region outside $r$ but within its bounding box) in a sub-channel respectively.
$\{j\}_1^b$ is the index of histogram bins.
Note that the more different the two histograms are, the higher the \rm{PD} is.

In~\cite{DBLP:conf/icpr/LiuS08a}, the authors quantified the perceptual divergence as the  overlapping areas between the normalized intensity histograms. However, using the intensity channel only ignores valuable color information, which will lead to a reduction in the measured perceptual divergence between distinct colors with the same intensity.
In contrast, all three sub-channels (\emph{i.e.}, R, G, B) are utilized in the computation of perceptual divergence in our approach.
\vspace{3mm}
\newline
\textbf{Histogram of Gradients at Edges (eHOG).}
The Histogram of Gradients (HOGs)~\cite{DBLP:conf/cvpr/DalalT05} is an
effective feature descriptor which captures the distribution of
gradient magnitude and orientation.
Inspired by~\cite{DBLP:conf/icpr/ZhangK10a}, we
propose a characterness cue based on the gradient
orientation at edges of a region, denoted by eHOG.
This cue aims to exploit the fact that the edge pixels of characters typically
appear in pairs with opposing gradient directions~\cite{DBLP:conf/icpr/ZhangK10a}\footnote{Let us assume the gradient orientation of an edge
pixel $p$ is ${\theta_p}$. If we follow the ray along this
direction or its inverse direction, we would possibly find another
edge pixel $q$,
whose gradient orientation, denoted by ${\theta_q}$, is
approximately opposite to $p$, \emph{i.e.}, $|\theta_p-\theta_q| \approx \pi$, as edges of a
character are typically closed.}. %
\begin{figure}[t]
\begin{center}
\begin{tabular}{@{}c@{}c}
\includegraphics[width=0.4\linewidth]{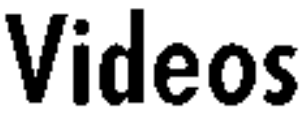} \ &
\includegraphics[width=0.4\linewidth]{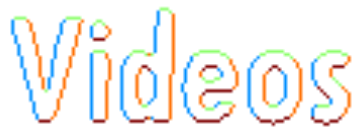} \ \\
\end{tabular}
\end{center}
\vspace{-3mm}
\caption{Sample text (left) and four types of edge points
represented in four different colors (right). Note that the number
of edge points in blue is roughly equal to that in orange, and so for green and crimson. }
\label{fig:HOG}
\end{figure}

Firstly, edge pixels of a region $r$ are extracted by the
Canny edge detector. Then, gradient orientations
$\theta$ of those pixels are quantized into four types, \emph{i.e.},
Type 1: $0 < \theta \le \pi /4$
or $7\pi /4 < \theta \le 2\pi $, Type
2: $\pi /4 < \theta \le 3\pi /4$, Type 3: $3\pi /4 < \theta \le
5\pi /4$, and Type 4: $5\pi /4 < \theta \le 7\pi /4$.
An example demonstrating the four types of edge pixels for text
is shown in Fig.~\ref{fig:HOG} (right), where four different
colors are used to depict the four types of edge pixels.
As it shows, we can expect that the number of edge pixels in
Type 1 should be close to that in Type 3, and so for Type 2 and
Type 4.
Based on this observation, we define the eHOG cue as:
\begin{equation}
{\rm eHOG(r)} = \frac{\sqrt {{{({w_1}(r) - {w_3}(r))}^2} + {{({w_2}(r) - {w_4}(r))}^2}}}{\sum\nolimits_{i = 1}^4 {{w_i}(r)}}   ,
\end{equation}
where $w_i(r)$ denotes the number of edge pixels in Type $i$
within region $r$, and the denominator $\sum\nolimits_{i = 1}^4 {{w_i}(r)}$ is for the sake of scale invariance.

\subsubsection{Bayesian multi-cue integration}
The aforementioned
cues measure the characterness
of a region $r$ from different perspectives.
\rm{SW and eHOG} distinguish characters from non-characters on the basis of their differing intrinsic structures.
\rm{PD} exploits surrounding color information.
Since they are complementary and obtained independently of each other, we argue
that combining them in the same framework outperforms any of the cues individually. %
Following the Naive Bayes model, we assume that each cue is conditionally independent.
Therefore, according to Bayes' theorem, the posterior
probability that a region is a character (its characterness score) can be computed as:
\begin{eqnarray*}
p(c|\Omega ) & = & \frac {p(\Omega|c) p(c)}{p(\Omega)}
\nonumber\\
&=& \frac {p(c) \mathop \prod \nolimits_{cue \in \Omega
}p(cue|c)} {\sum_{k \in \{c,b\}} %
p(k) \prod
\nolimits_{cue \in \Omega} p(cue|k)} ,
\label{EQ:Bayes}
\end{eqnarray*}
where $\Omega  = \{ {\rm SW,PD, eHOG} \}$,
and $p(c)$ and $p(b)$ denote the prior probability of characters and background respectively, which we determine on the basis of relative frequency.
We model the observation likelihood $p(cue|c)$ and $p(cue|b)$ via distribution of cues on positive and negative samples respectively, with details provided as follows.
\vspace{3mm}
\newline
\textbf{Learning the Distribution of Cues.}
In order to learn the distribution of the proposed cues, we use the training set of text segmentation task in the ICDAR 2013 robust reading competition (challenge 2).
To our knowledge, this is is the only benchmark dataset with pixel-level ground truth so far.
This dataset contains 229 images harvested from natural scenes.
We randomly selected 119 images as the training set, while the rest 100 images were treated as the test set for characterness evaluation in our experiment (Sec. \ref{subec:CharacternessEvaluation}).
\begin{itemize}
\item To learn the distribution of cues on positive samples, we directly compute the three cues on characters, as pixel-level ground truth is provided.
\item To learn the distribution of cues on negative samples, eMSER algorithm is applied twice to each training image. After erasing ground truth characters, the rest of the extracted regions are considered as negative samples on which we compute the three cues.
\end{itemize}
\begin{figure}[t]
\centering
\begin{tabular}{@{}c@{}c}
\includegraphics[width=0.45\linewidth]{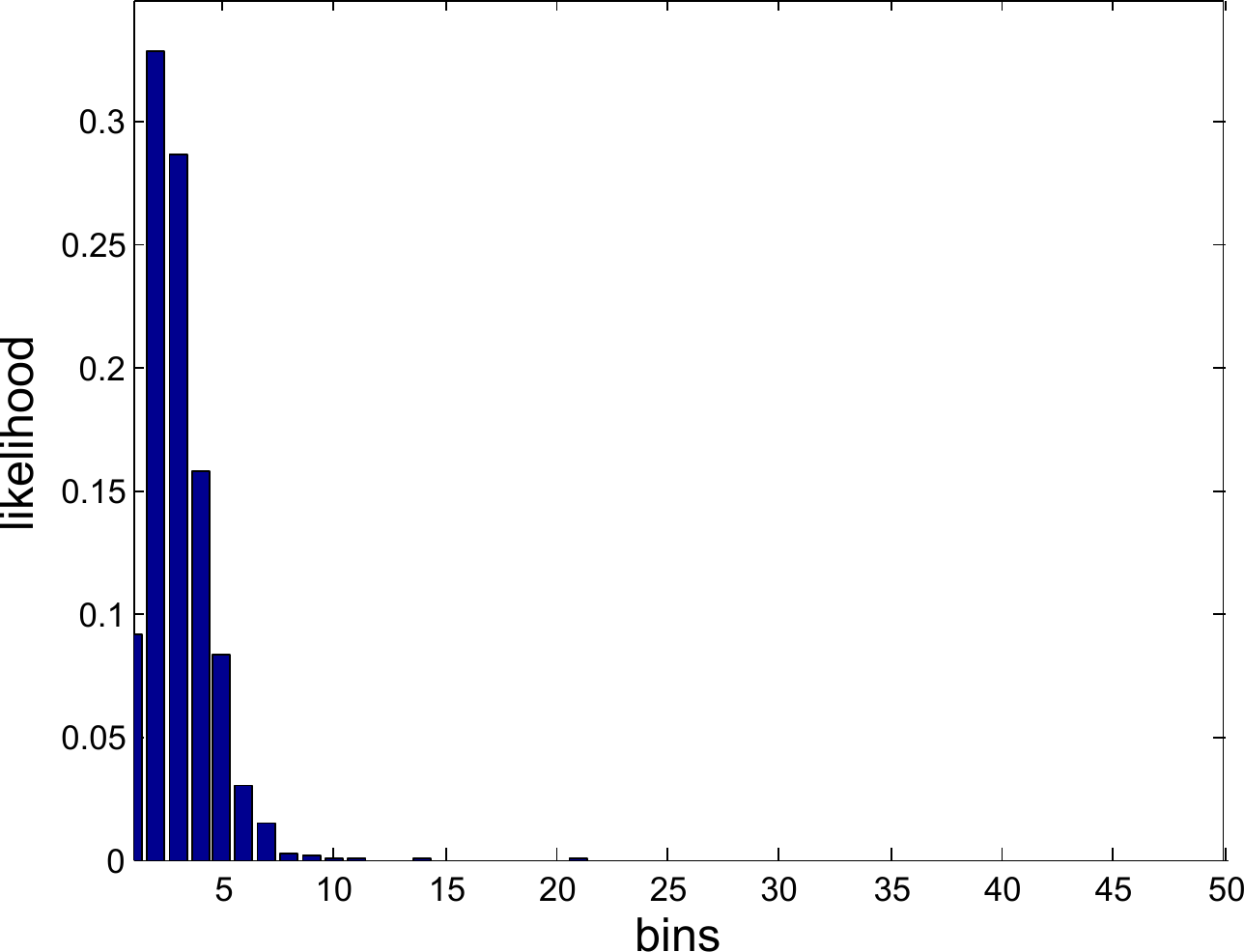} &
\includegraphics[width=0.45\linewidth]{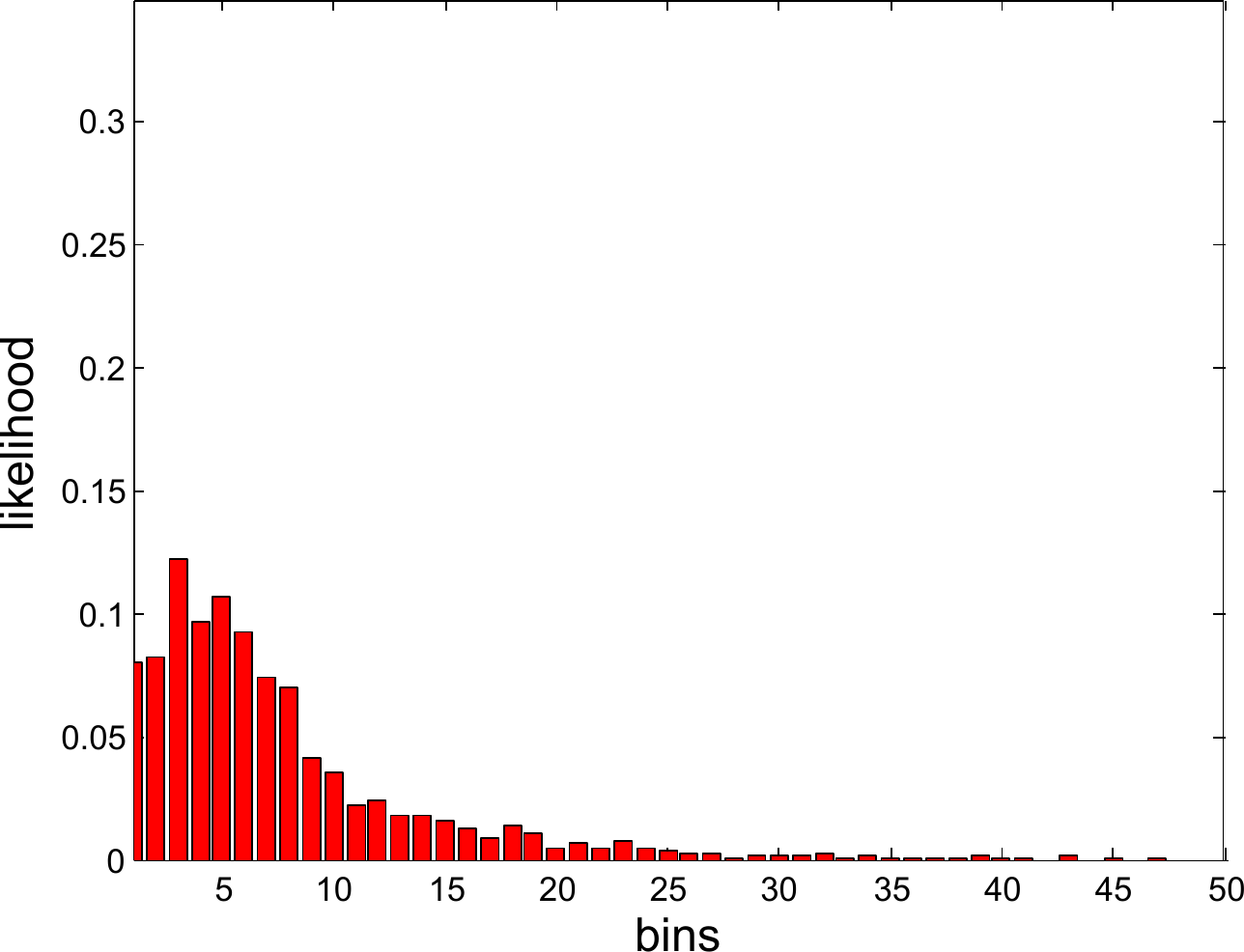}\\
\includegraphics[width=0.45\linewidth]{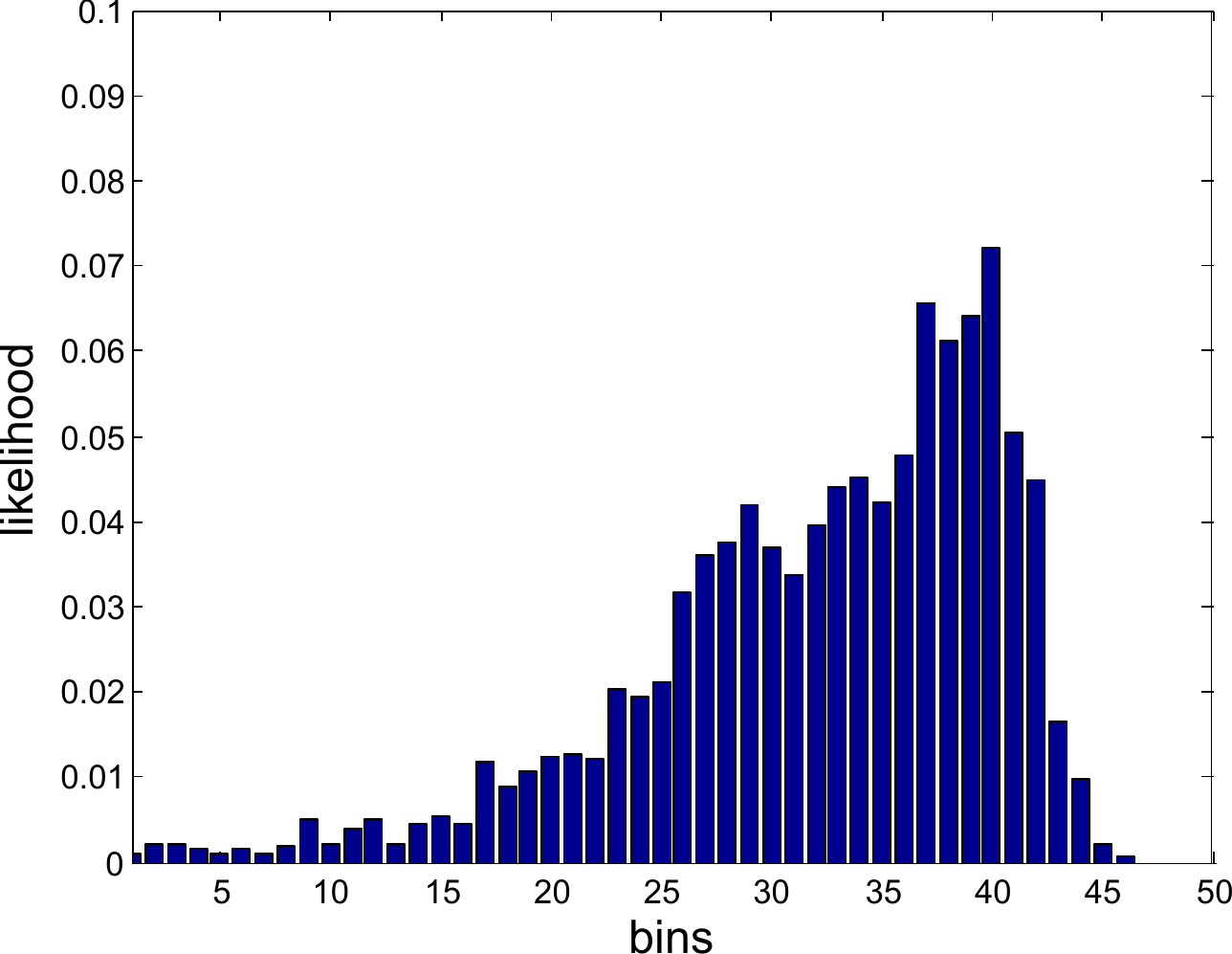}&
\includegraphics[width=0.45\linewidth]{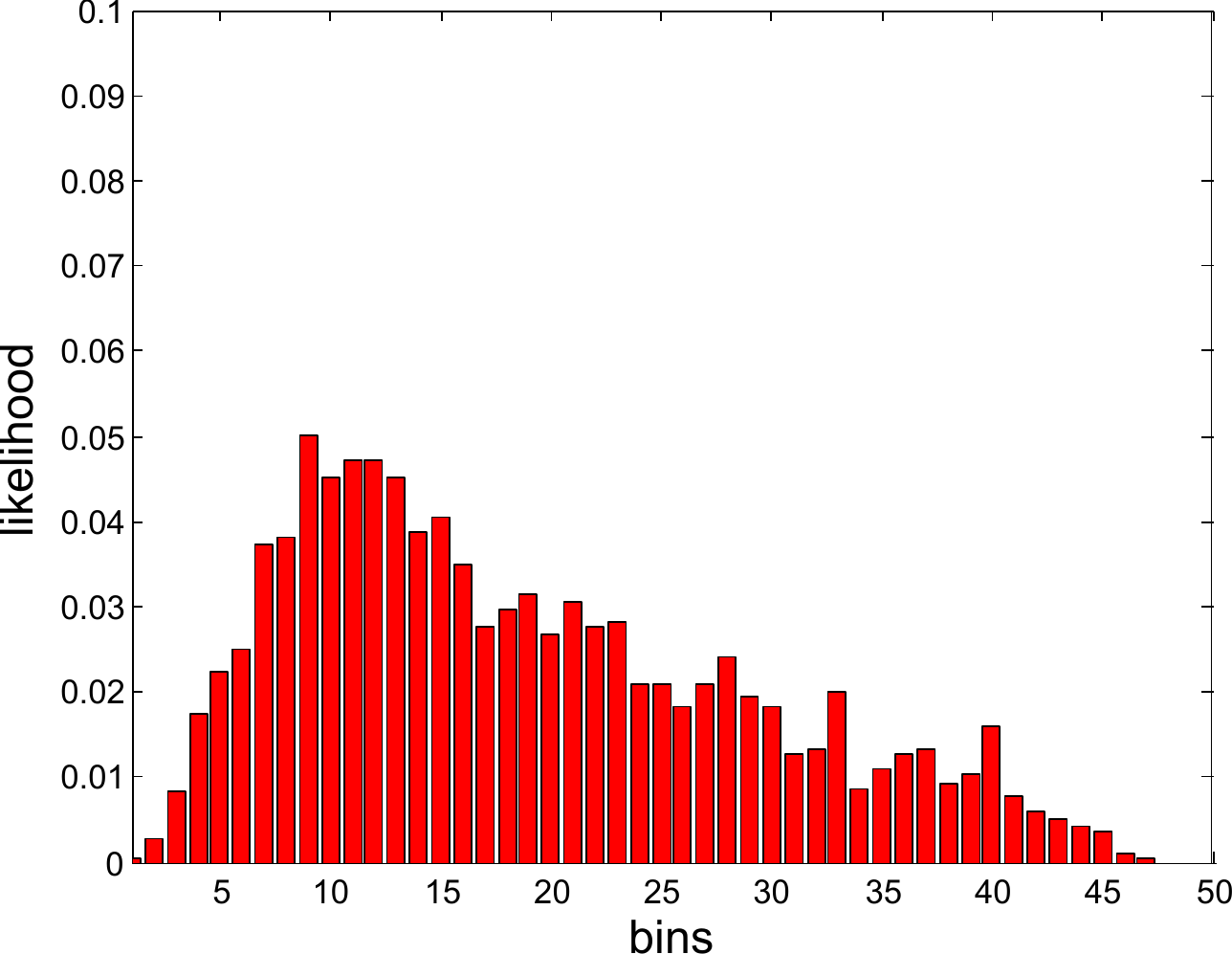}\\
\includegraphics[width=0.45\linewidth]{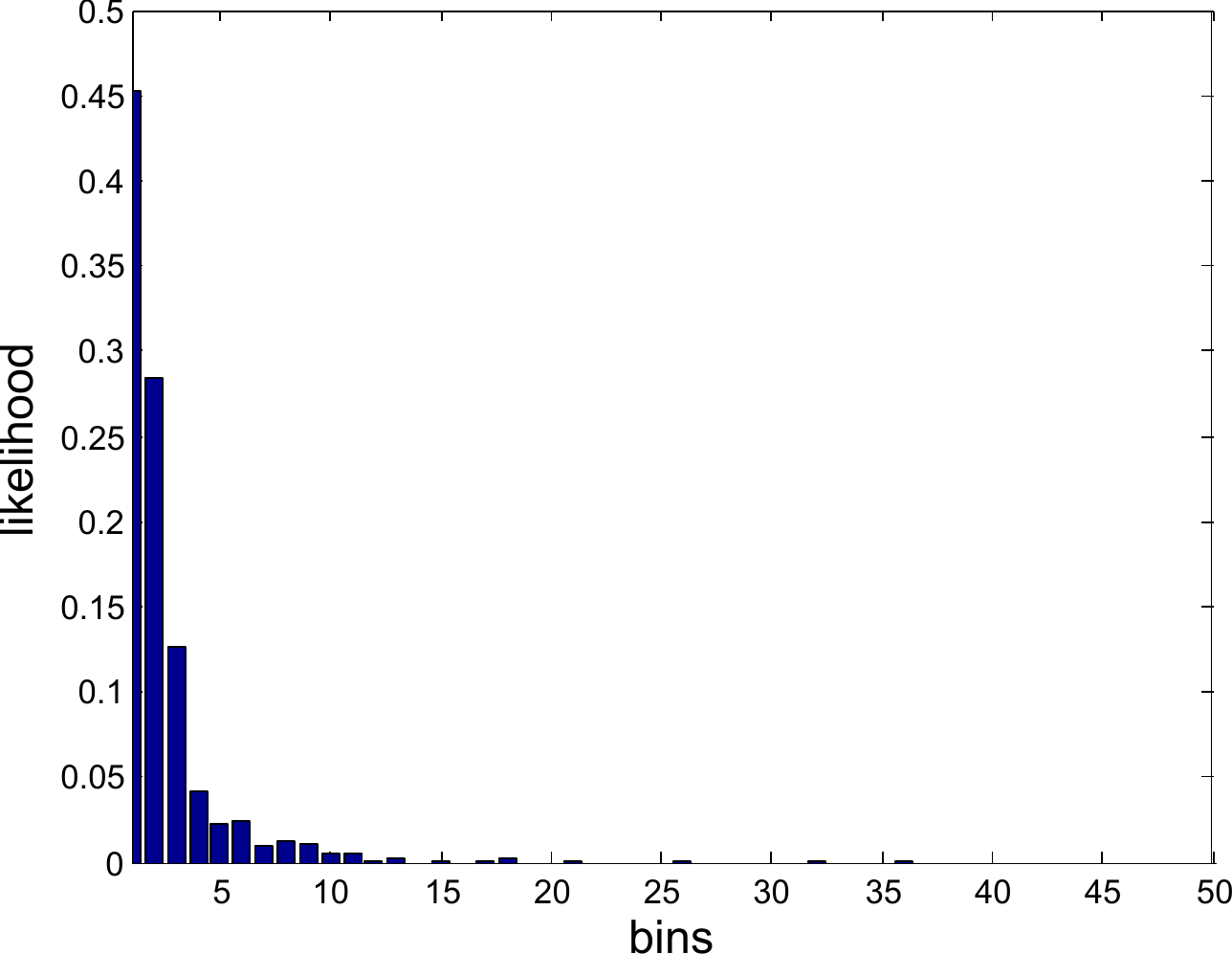}&
\includegraphics[width=0.45\linewidth]{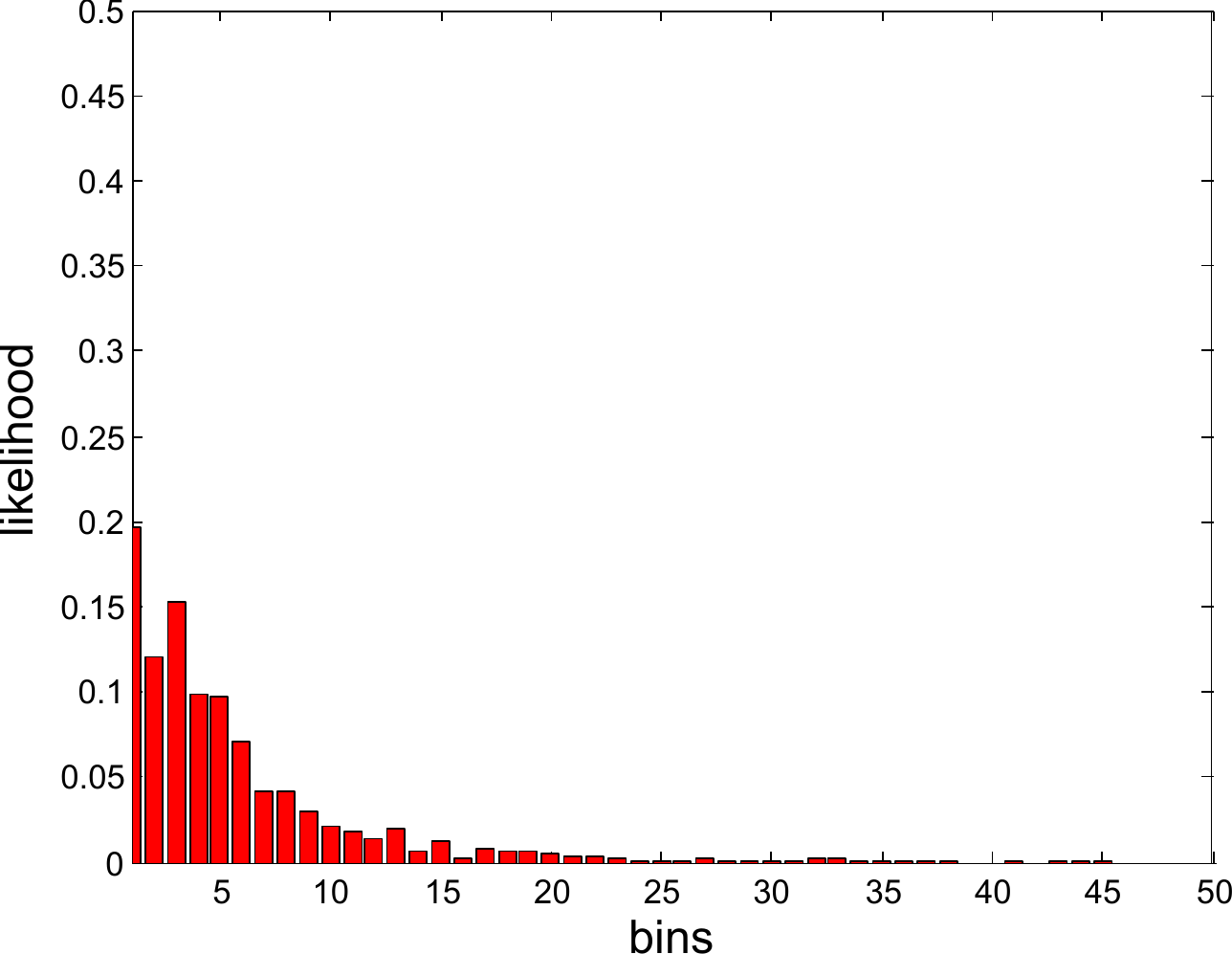}\\
\end{tabular}
\caption{%
Observation likelihood of characters (blue) and non-characters (red) on three characterness cues \emph{i.e.}, SW (top row), PD (middle row), and eHOG (bottom row). Clearly, for all three cues, observation likelihoods of characters are quite different from those of non-characters, indicating that the proposed cues are effective in distinguishing them. Notice that 50 bins are adopted.}
\label{fig:likelihood}
\end{figure}

Fig. \ref{fig:likelihood} shows the distribution of the three cues via normalized histograms.
As it shows, for both SW and eHOG, compared with non-characters, characters are more likely to have relatively smaller values (almost within the first 5 bins). For the distribution of PD, it is clear that characters tend to have higher contrast than that of non-characters.

\section{Labeling and Grouping}
\label{sec:TextDetection}
\subsection{Character labeling}
\label{subsec:MRF}
\subsubsection{Labeling model overview}
We cast the task of separating characters from non-characters as a binary labeling
problem.
To be precise, we construct a standard graph $\mathcal{G}=(\mathcal{V},\mathcal{E})$,
where $\mathcal{V}=\{v_{i}\}$ is the vertex set corresponding to the candidate
characters, and $\mathcal{E}=\{e_j\}$ is the edge set corresponding to the interaction between vertexes.\footnote{In our work, we consider the edge between two vertexes (regions) exists only if the
Enclidean distance between their centroids is smaller than the minimum of
their characteristic scales. Characteristic scale is defined as the
sum of the length of major axis and minor axis~\cite{DBLP:conf/cvpr/Yao}.}
Each $v_i\in \mathcal{V}$ should be labeled as either character, \emph{i.e.}, $l_i=1$, or non-character, \emph{i.e.}, $l_i=0$.
Therefore, a labeling set $\mathcal{L}=\{l_i\}$ represents the separation of characters from non-characters.
The optimal labeling $\mathcal{L^*}$ can be found by minimizing an energy function:
\begin{equation}
\mathcal{L^*}=\arg\min_{\mathcal{L}} E(\mathcal{L}),
\end{equation}
where $E(\mathcal{L})$ consists of the sum of two potentials:
\begin{align}
E(\mathcal{L})=U(\mathcal{L})+V(\mathcal{L}) \\
U(\mathcal{L})=\sum_i u_i(l_i)\\
V(\mathcal{L})=\sum_{ij\in\mathcal{E}}v_{ij}(l_i,l_j),
\end{align}
where $u_i(l_i)$ is the unary potential which determines the cost of assigning the label $l_i$ to $v_i$.
$v_{ij}(l_i,l_j)$ is the pairwise potential which reflects the cost of assigning different labels to $v_i$ and $v_j$.
This model is widely adopted in image segmentation algorithms~\cite{DBLP:conf/iccv/BoykovJ01,DBLP:conf/cvpr/KuttelF12}.
The optimal $\mathcal{L^*}$ can be found efficiently using graph-cuts~\cite{DBLP:journals/pami/BoykovVZ01} if the pairwise potential is submodular.
\subsubsection{The design of unary potential}
characterness score of extracted regions is encoded in the design of unary potential directly:
\begin{equation}
u_i(l_i)=
\begin{cases}
p(c|\Omega) & l_i=0 \\
1-p(c|\Omega) & l_i=1.
\end{cases}
\end{equation}
\subsubsection{The design of pairwise potential}
As characters typically appear in homogeneous groups, the degree to which properties of a putative character (stroke width and color, for example) match
those of its neighbors is an important indicator.
This clue plays an important role for human vision to distinguish characters
from cluttered background and can be exploited to design the pairwise potential.
In this sense, similarity between extracted regions is measured by the following two cues.

\textbf{Stroke Width Divergence (SWD).} To measure the stroke width divergence between two extracted regions $r_1$ and $r_2$, we leverage on stroke width histogram.
In contrast with Algorithm~\ref{algo:stroke width} where only pixels on the skeleton are taken into account, distance transform is applied to all pixels within the region to find length of shortest path.
Therefore, the stroke width histogram is defined as the histogram of shortest length.
Then, \rm{SWD} is measured as the discrete KLD (\emph{c.f.} Equ.\ref{eq:PD}) of two stroke
width histograms.

\textbf{Color Divergence (CD).}
The color divergence of two regions is the distance of their average color (in the LAB space) measured by L2 norm.

The aforementioned two cues measure divergence between two regions from two distinct prospectives. Here, we combine them efficiently to produce the unified divergence (UD):
\begin{equation}
\rm{UD}(r_1,r_2) = \beta\rm{SWD}(r_1,r_2)+(1-\beta)\rm{CD}(r_1,r_2),
\end{equation}
where the coefficient $\beta$ specifies the relative weighting of the two
divergence.
Without losing generality, in our experiments we set $\beta=0.5$ so that the
two divergence are equally weighted.
We make use of the unified divergence to define the pairwise potential as:
\begin{equation}
v_{ij}(l_i,l_j)=[l_i\neq l_j](1-\tanh (\rm{UD}(r_i,r_j))),
\end{equation}
where $[\cdot]$ is the Iverson bracket. In other words, the more similar the color and stroke width of the
two vertexes are, the less likely they are assigned with
different labels.

\subsection{Text line formulation}
\label{subsec:grouping}
The goal of this step, given a set of characters identified in the previous step, is to group them
into readable \emph{lines} of text.
A comparable step is carried out in some existing text detection approaches~\cite{DBLP:conf/cvpr/EpshteinOW10,DBLP:conf/cvpr/Yao},
but the fact that these methods have many parameters which must be tuned to adapt to individual data means that
the adaptability of these methods to various data sets remains
unclear.
We thus introduce a mean shift based clustering scheme for
text line extraction.

Two features exploited in mean shift
clustering are characteristic scale and major orientation~\cite{DBLP:conf/cvpr/Yao}.
Note that both features are normalized. %
Clusters with at least two elements are retained for further processing.

Within each cluster a bottom-up grouping method is performed, with the goal that only
characters within the same line of text will be assigned the same label.
In order to achieve this goal all regions are set as unlabeled initially.
For an unlabeled region, if another unlabeled region is
nearby (less than the average of their skeleton length), both are
given the same label
and the angle of the line connecting their centroids is taken
as the text line angle.
On the other hand, for a labeled region, if another unlabeled region is nearby
and the angle between them is similar to that of the text line (less
than 30 degrees), the latter is assigned the label of the former,
and the angle of the text line is updated.
\begin{figure*}[t]
\centering
\begin{tabular}{@{}c@{}c@{}c}
\includegraphics[width=0.3\linewidth]{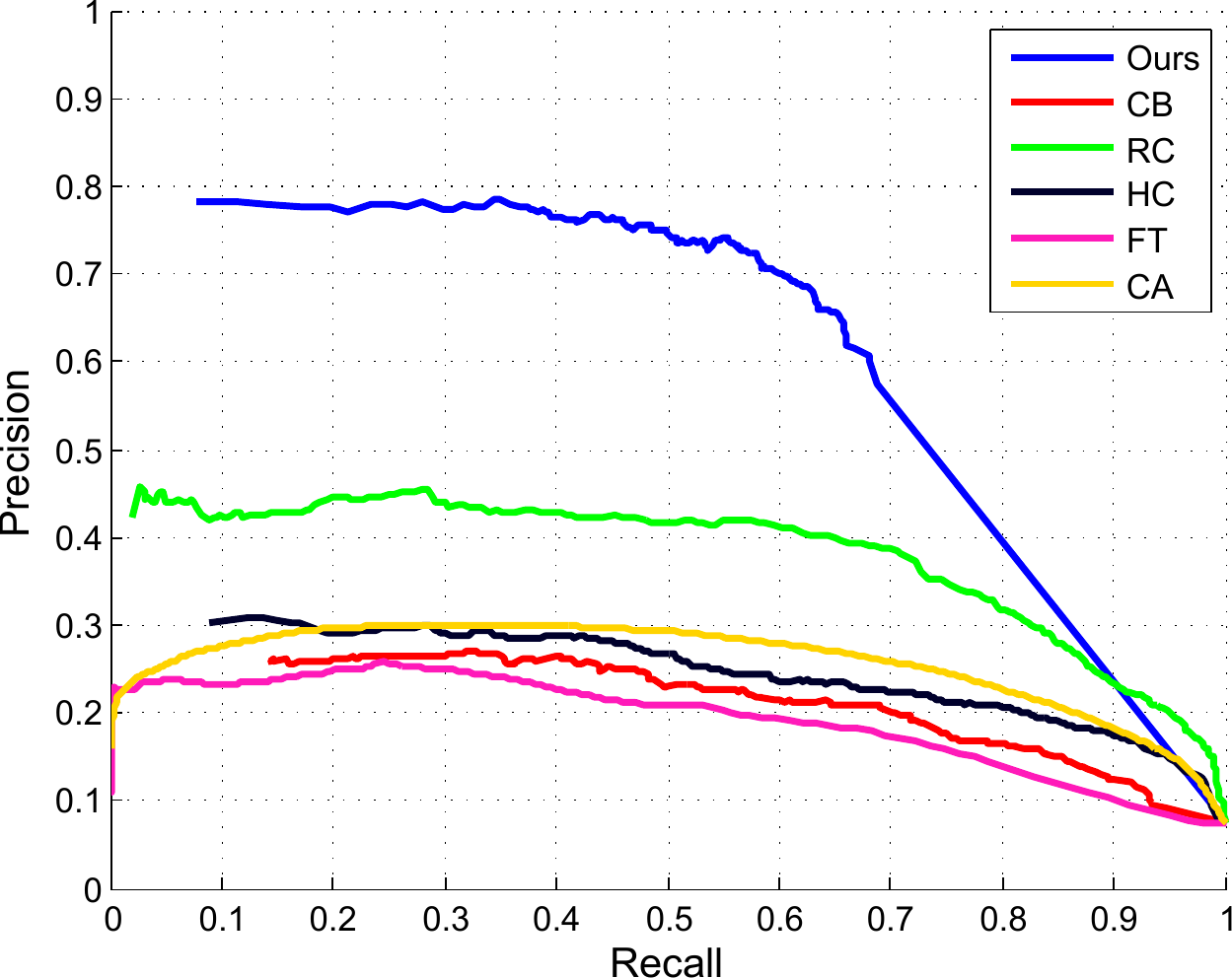}&
\includegraphics[width=0.3\linewidth]{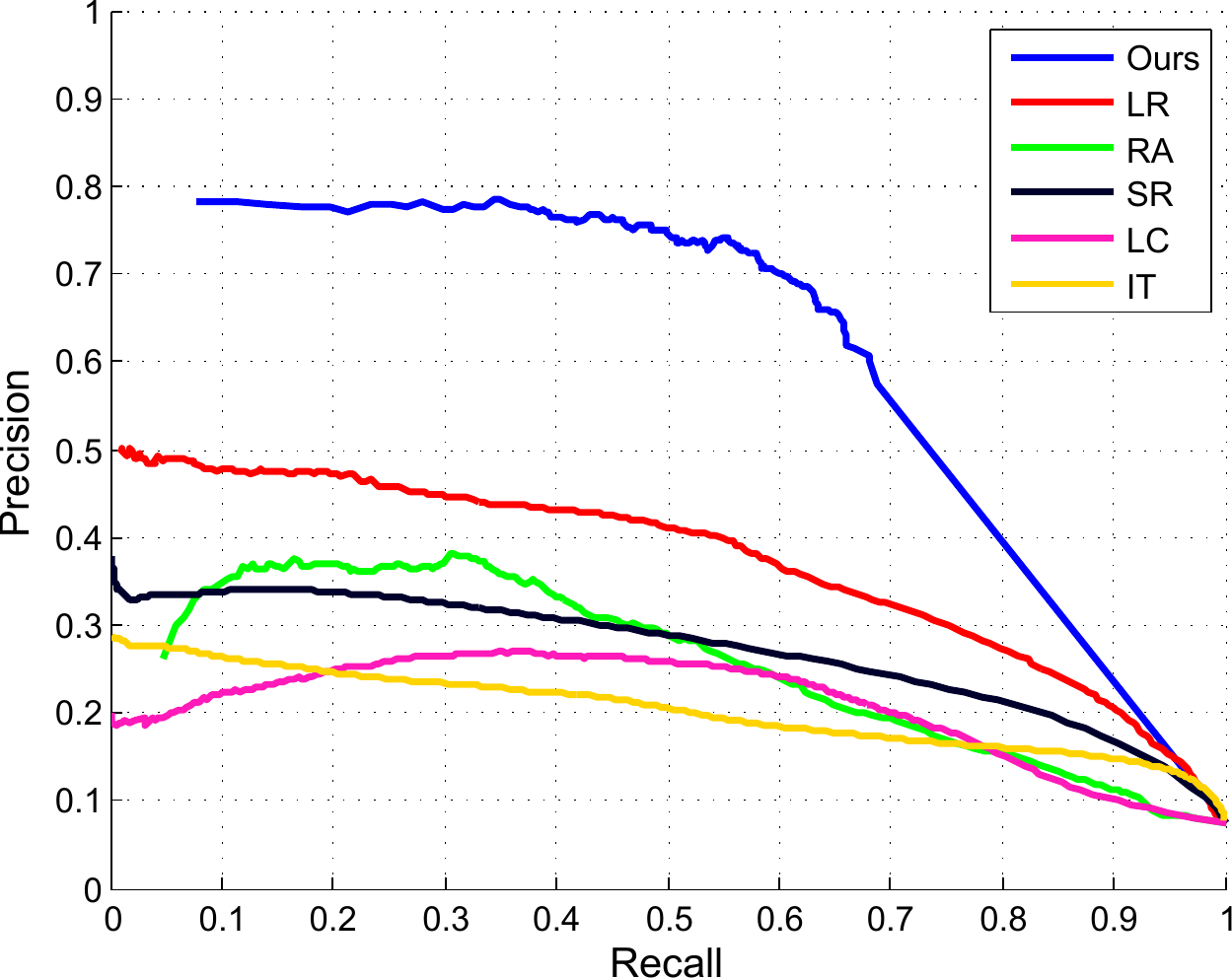}&
\includegraphics[width=0.3\linewidth]{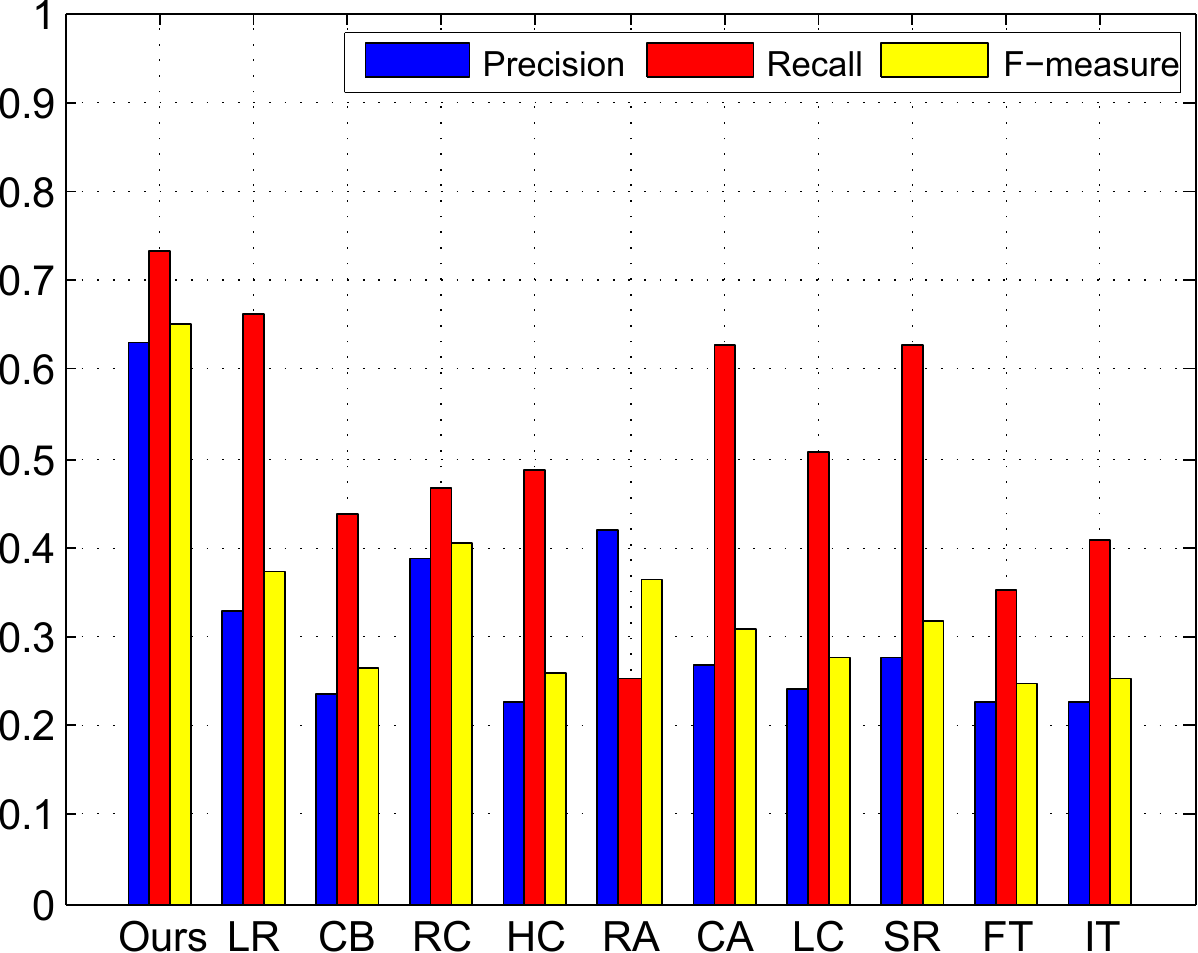}\\
\end{tabular}
\caption{Quantitative precision-recall curves (left, middle) and F-measure (right) performance of all the eleven approaches. Clearly, our approach achieves
significant improvement compared with state-of-the-art saliency detection models for
the measurement of 'characterness'.}
\label{fig:PRcurve}
\end{figure*}
\begin{table*}[t]
\caption{Quantitative performance of all the eleven approaches in
VOC overlap scores.}
\hspace{-0.3cm}
\centering
\begin{tabular}{ |c|c|c|c|c|c|c|c|c|c|c| }
\hline
Ours &  LR~\cite{shen2012unified} & CB~\cite{jiang2011automatic} &  RC~\cite{cheng2011global} & HC~\cite{cheng2011global} &  RA~\cite{DBLP:conf/eccv/RahtuKSH10}  & CA~\cite{DBLP:conf/cvpr/GofermanZT10} & LC~\cite{DBLP:conf/mm/ZhaiS06} & SR~\cite{hou2007saliency} & FT~\cite{achanta2009frequency} & IT~\cite{itti1998model}\\
\hline
{\bf 0.5143} & 0.2766 & 0.1667 & 0.2717 & 0.2032 & 0.1854 & 0.2179 & 0.2112 & 0.2242 & 0.1739 & 0.1556\\
\hline
\end{tabular}

\label{tal:overlapratio}  \vspace{-0.29cm}
\end{table*}
\section{Proposed Characterness Model Evaluation}
\label{sec:CharacternessEvaluation}
To demonstrate the effectiveness of the proposed characterness model, we follow
the evaluation of salient object detection algorithm.
Our characterness map is normalized to [0,1], thus treated as saliency map.
Pixels with high saliency value (\emph{i.e.,} intensity) are likely to belong to salient objects (characters in our scenario) which catch human attention.

We qualitatively and quantitatively compare the proposed `characterness' approach with ten existing saliency detection models: the classical Itti's
model (IT)~\cite{itti1998model}, the spectral residual approach (SR)~\cite{hou2007saliency}, the frequency-tuned approach (FT)~\cite{achanta2009frequency}, context-aware saliency (CA)~\cite{DBLP:conf/cvpr/GofermanZT10}, Zhai's method (LC)~\cite{DBLP:conf/mm/ZhaiS06}, histogram-based saliency (HC)~\cite{cheng2011global}, region-based saliency (RC)~\cite{cheng2011global}, Jiang's method (CB)~\cite{jiang2011automatic}, Rahtu's method (RA)~\cite{DBLP:conf/eccv/RahtuKSH10} and more recently low-rank matrix decomposition (LR)~\cite{shen2012unified}.
Note that CB, RC and CA are considered as the best salient object detection models in the benchmark work~\cite{DBLP:conf/eccv/BorjiSI12}.
For SR and LC, we use the implementation from~\cite{cheng2011global}. For the rest approaches, we use the publicly available implementations from the original authors.
To the best of our knowledge, we are the first to evaluate the state-of-the-art saliency detection models for reflecting characterness in this large quantity.

Unless otherwise specified, three parameters in Algorithm~\ref{algo:eMSER} were set as follows: the
delta value ($\Delta$) in the MSER was set to 10, and the local window
radius in the guided filter was set to 1, $\gamma=0.5$.
We empirically found that these parameters work well for different datasets.

\subsection{Datasets} For the sake of
more precise evaluation of `characterness', we need pixel-level ground truth of characters.\footnote{Dataset of pixel-level ground truth is also adopted in~\cite{DBLP:conf/das/ShahabSDU12}. However, it is not publicly available.}
As mentioned in Sec. \ref{subec:CharacternessEvaluation}, to date, the only benchmark dataset with pixel-level ground truth is the training set of text segmentation task in the ICDAR 2013 robust reading competition (challenge 2) which consists of 229 images.
Therefore, we randomly selected 100 images of this dataset here for evaluation (the other 119 images have been used for learning the distribution of cues in the Bayesian framework).

\subsection{Evaluation criteria}
For a given saliency map, three criteria are adopted
to evaluate the quantitative performance of
different approaches: precision-recall (PR) curve,
F-measure and VOC overlap score.
In all three cases we generate a binary segmentation mask of the saliency map at a threshold $T$.

To obtain the PR curve, we first get 256 binary segmentation masks from the saliency map using threshold $T$ ranging from 0 to 255, as
in~\cite{achanta2009frequency,cheng2011global,shen2012unified,perazzi2012saliency}.
For each segmentation mask, precision and recall rate are obtained by comparing it with the ground truth mask.
Therefore, in total 256 pairs of precision and recall rates are utilized to plot the PR curve.
In contrast with the computation of the PR curve, to get F-measure~\cite{achanta2009frequency}, $T$ is fixed as the twice of the mean saliency value of the image to get precision rate $P$ and recall rate $R$.
F-measure is computed as
$((\beta^{2}+1)P\cdot R)/(\beta^{2}P + R)$.
We set $\beta^2=0.3$ as that in~\cite{achanta2009frequency}.

The VOC overlap score~\cite{RosenfeldW11} is defined
as $\frac{|S \cap S'|}{|S \cup S'|}$.
Here, $S$ is the ground truth mask,
and $S'$ is the our segmentation mask
obtained by binarizing
the saliency map using the same adaptive threshold $T$
during the calculation of F-measure.

The resultant PR curve (resp. F-measure, VOC overlap score) of a dataset is generated by averaging PR curves (resp. F-measure, VOC overlap score) of all images in the dataset.

\subsection{Comparison with other approaches}
\begin{figure*}
\centering
\begin{tabular}{@{}c@{}c@{}c@{}c@{}c@{}c@{}c@{}c@{}c@{}c@{}c@{}c@{}c}
{\scriptsize Image}
& {\scriptsize GT}
& {\scriptsize Ours}
& {\scriptsize LR\cite{shen2012unified}}
& {\scriptsize CB\cite{jiang2011automatic}}
& {\scriptsize RC\cite{cheng2011global}} &
{\scriptsize HC\cite{cheng2011global}} & {\scriptsize RA\cite{DBLP:conf/eccv/RahtuKSH10}}
&{\scriptsize CA\cite{DBLP:conf/cvpr/GofermanZT10}} & {\scriptsize LC\cite{DBLP:conf/mm/ZhaiS06}} & {\scriptsize SR\cite{hou2007saliency}}
& {\scriptsize FT\cite{achanta2009frequency}}
& {\scriptsize IT\cite{itti1998model}} \\
\includegraphics[width=0.07\linewidth]{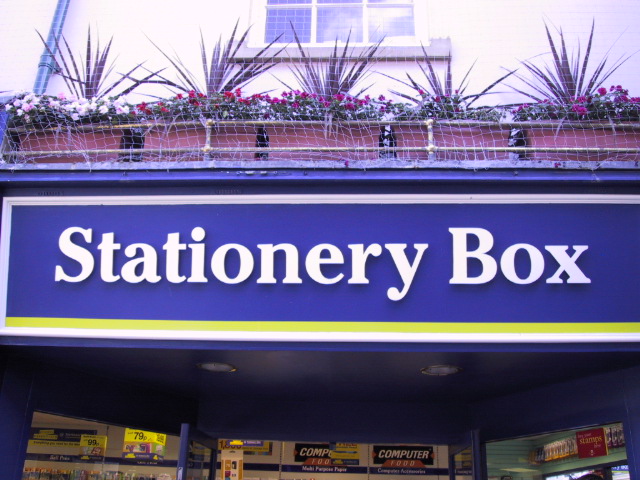} \ &
\includegraphics[width=0.07\linewidth]{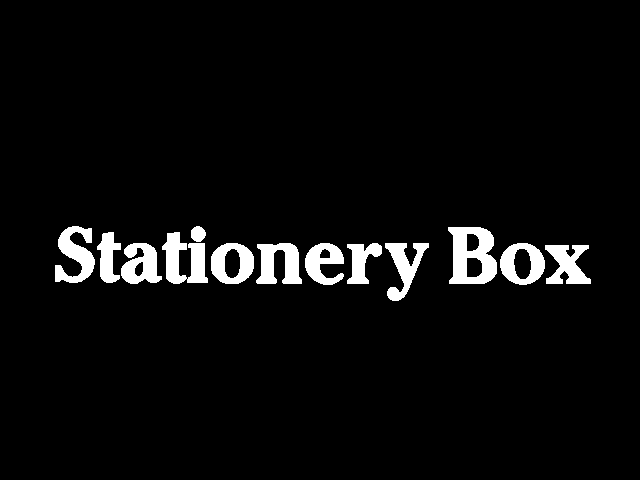} \ &
\includegraphics[width=0.07\linewidth]{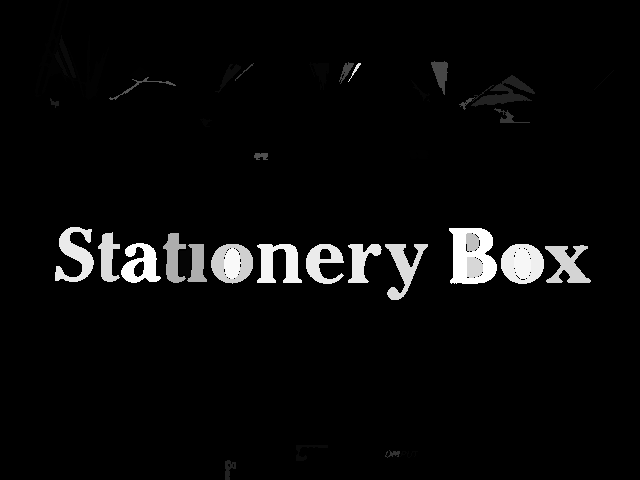} \ &
\includegraphics[width=0.07\linewidth]{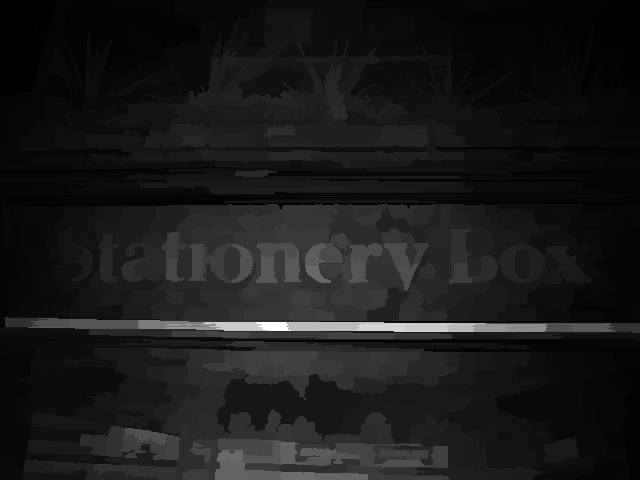} \ &
\includegraphics[width=0.07\linewidth]{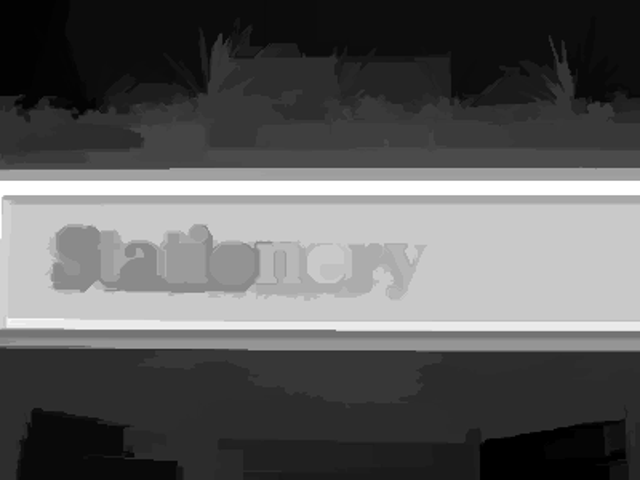} \ &
\includegraphics[width=0.07\linewidth]{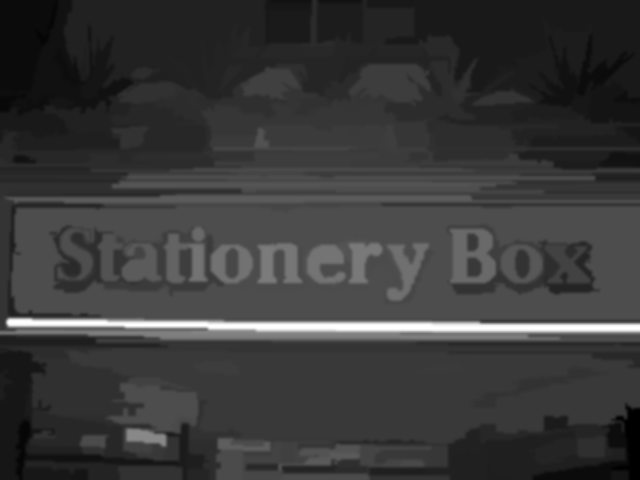} \ &
\includegraphics[width=0.07\linewidth]{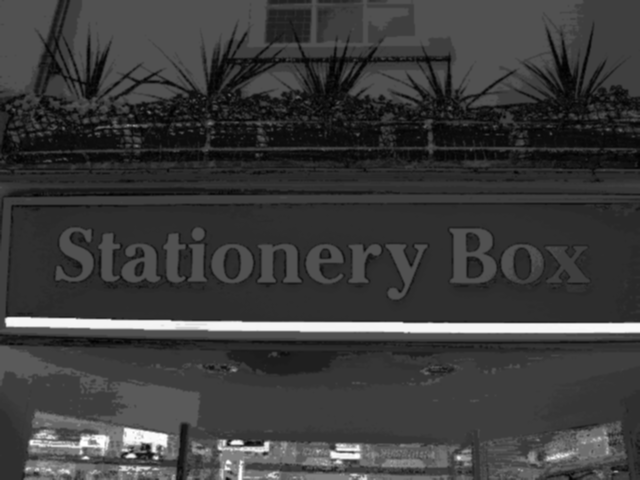} \ &
\includegraphics[width=0.07\linewidth]{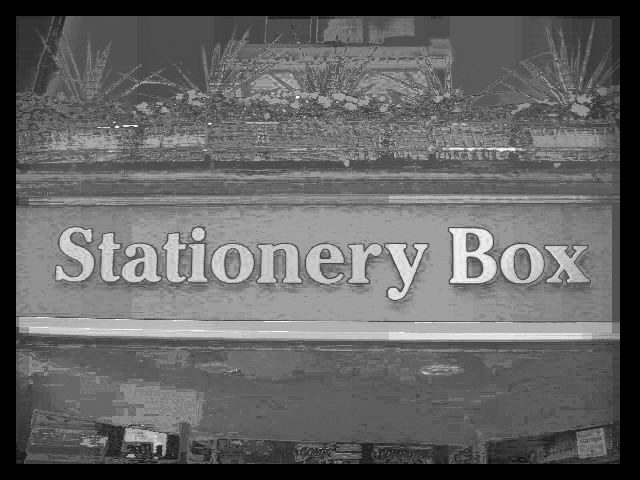} \ &
\includegraphics[width=0.07\linewidth]{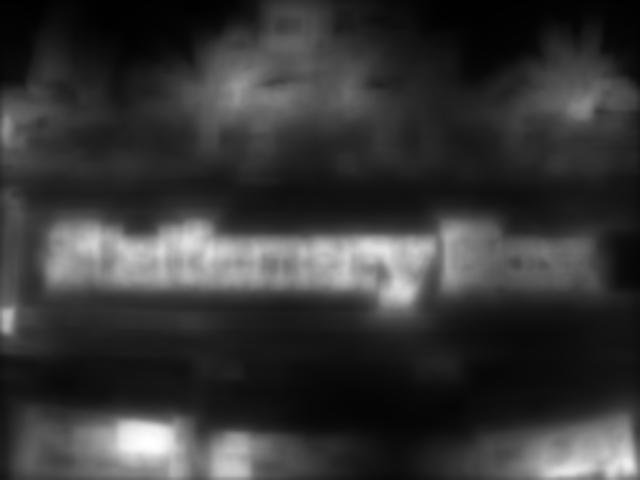} \ &
\includegraphics[width=0.07\linewidth]{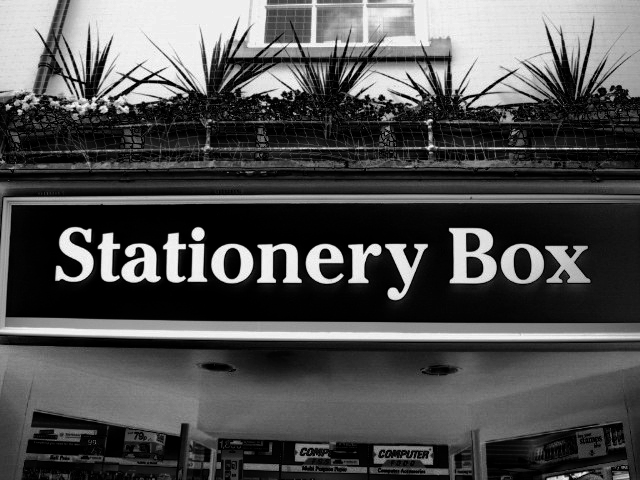} \ &
\includegraphics[width=0.07\linewidth]{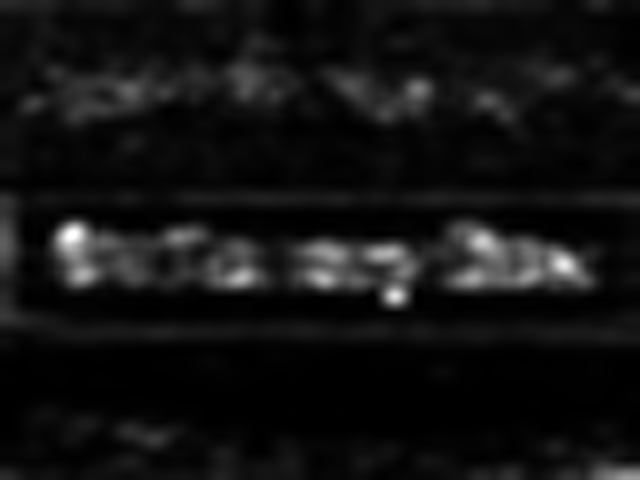} \ &
\includegraphics[width=0.07\linewidth]{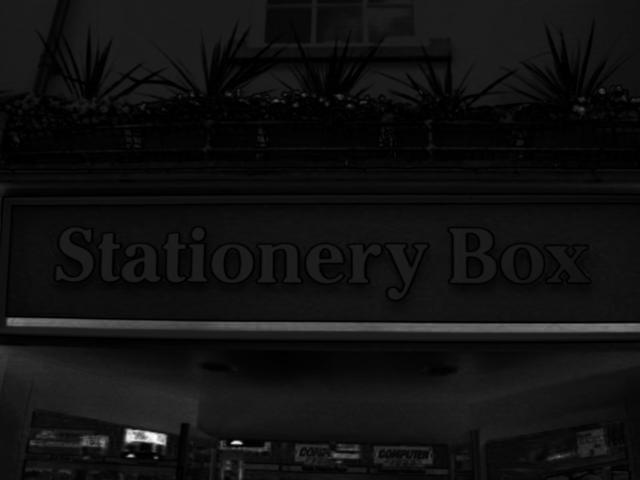} \ &
\includegraphics[width=0.07\linewidth]{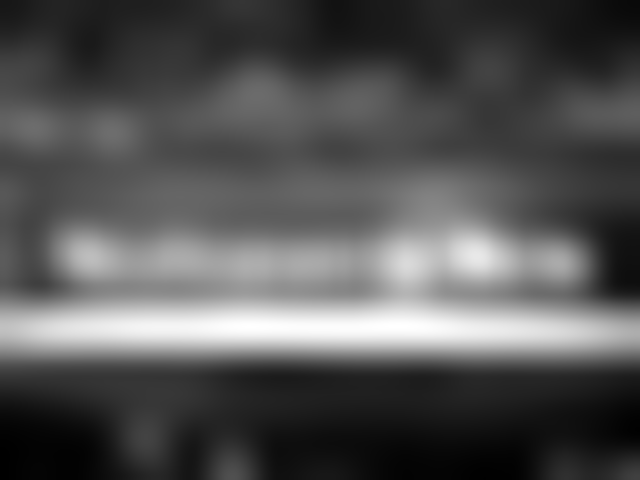} \\
\includegraphics[width=0.07\linewidth]{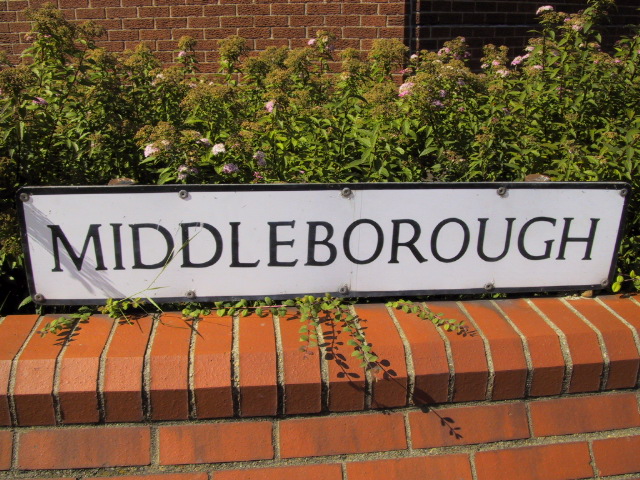} \ &
\includegraphics[width=0.07\linewidth]{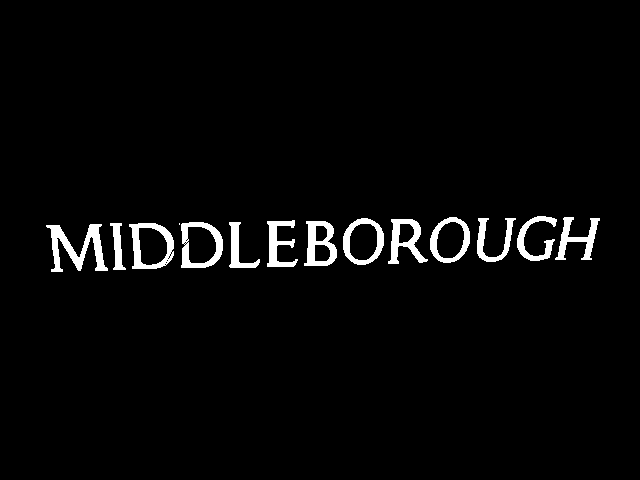} \ &
\includegraphics[width=0.07\linewidth]{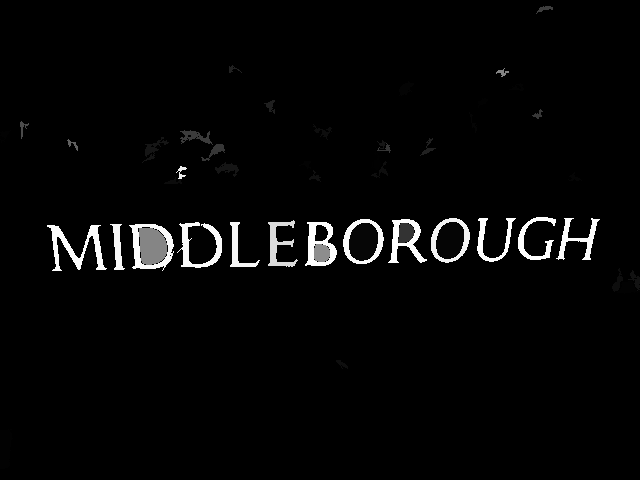} \ &
\includegraphics[width=0.07\linewidth]{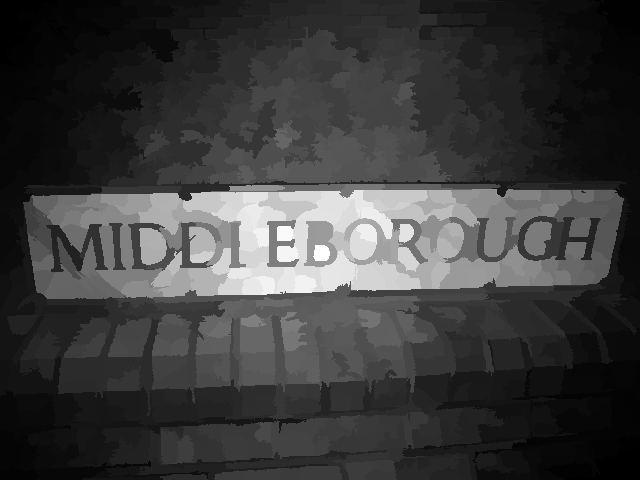} \ &
\includegraphics[width=0.07\linewidth]{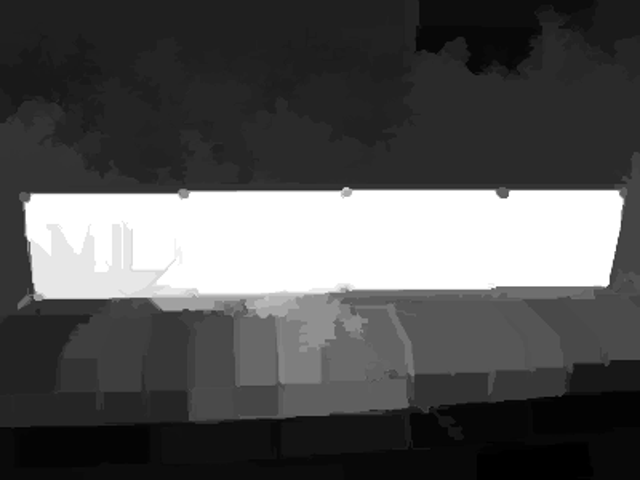} \ &
\includegraphics[width=0.07\linewidth]{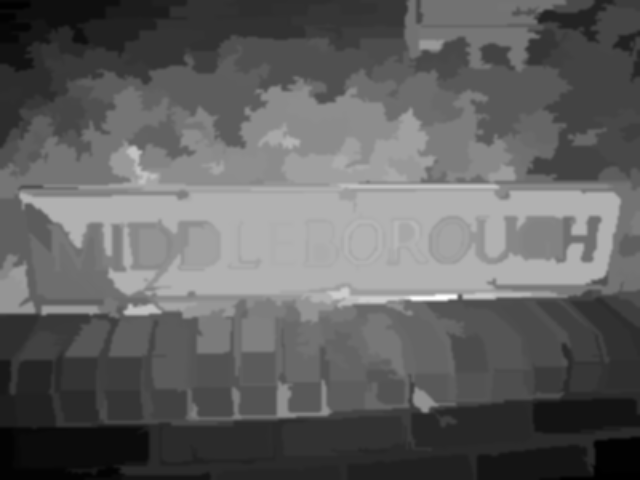} \ &
\includegraphics[width=0.07\linewidth]{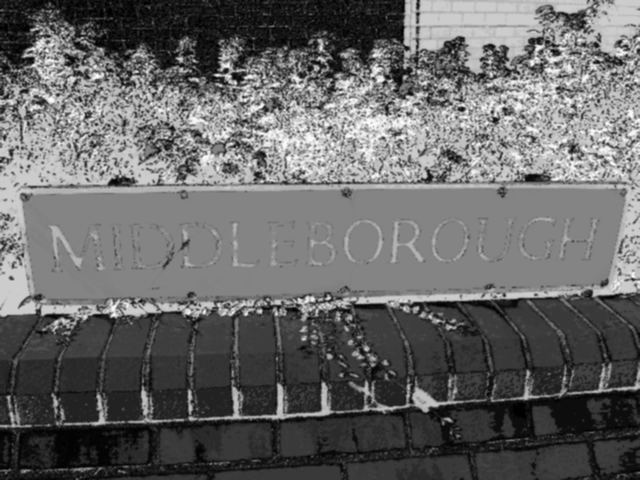} \ &
\includegraphics[width=0.07\linewidth]{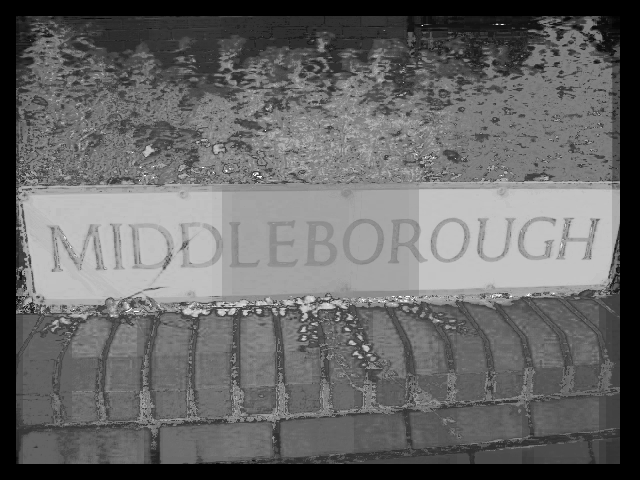} \ &
\includegraphics[width=0.07\linewidth]{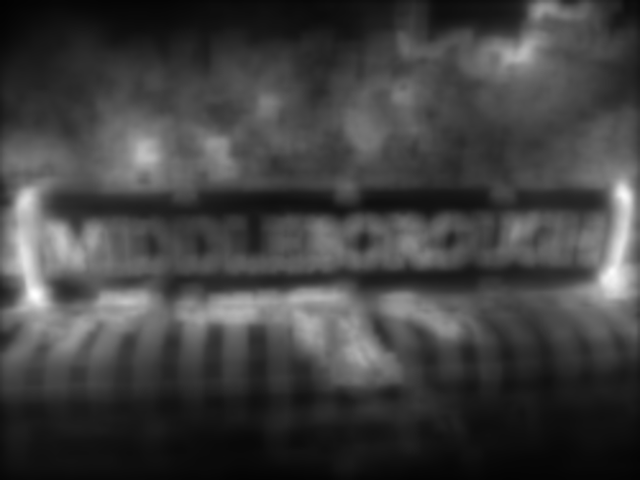} \ &
\includegraphics[width=0.07\linewidth]{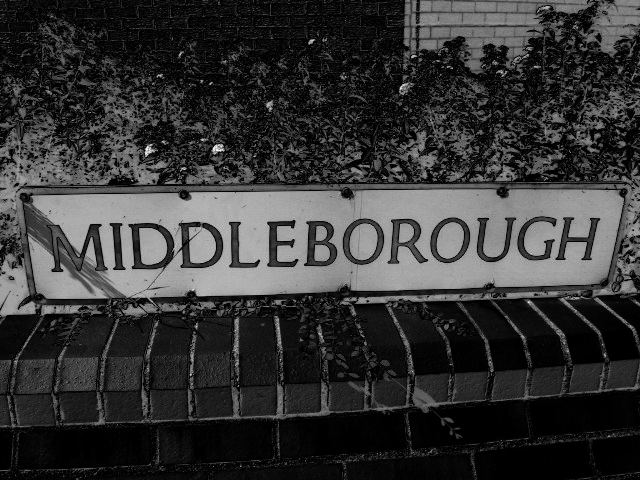} \ &
\includegraphics[width=0.07\linewidth]{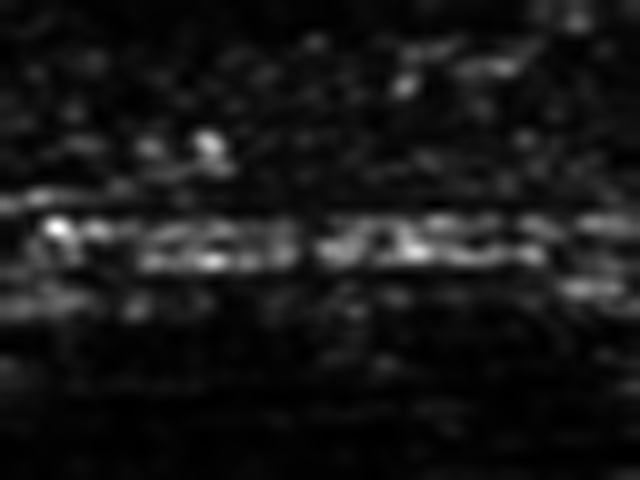} \ &
\includegraphics[width=0.07\linewidth]{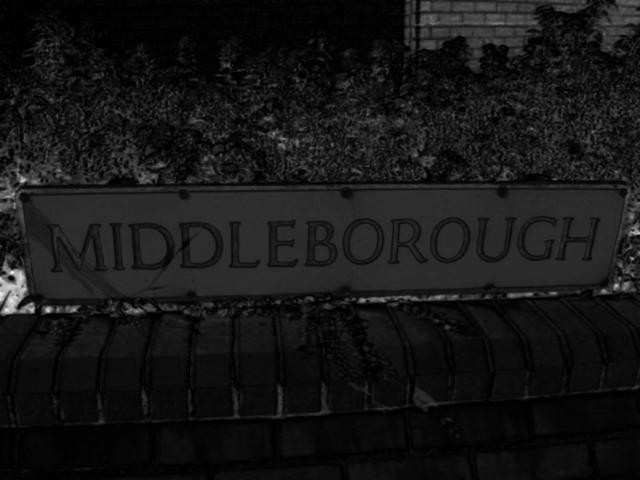} \ &
\includegraphics[width=0.07\linewidth]{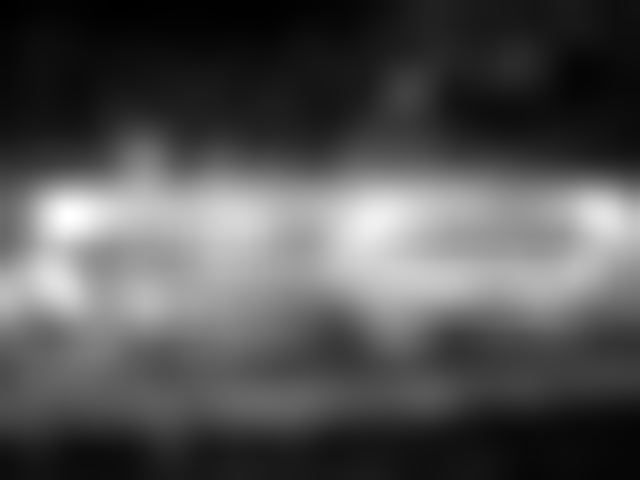} \\
\includegraphics[width=0.07\linewidth]{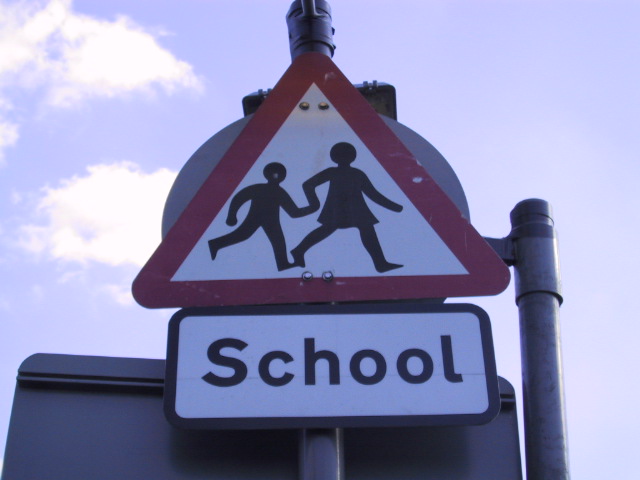} \ &
\includegraphics[width=0.07\linewidth]{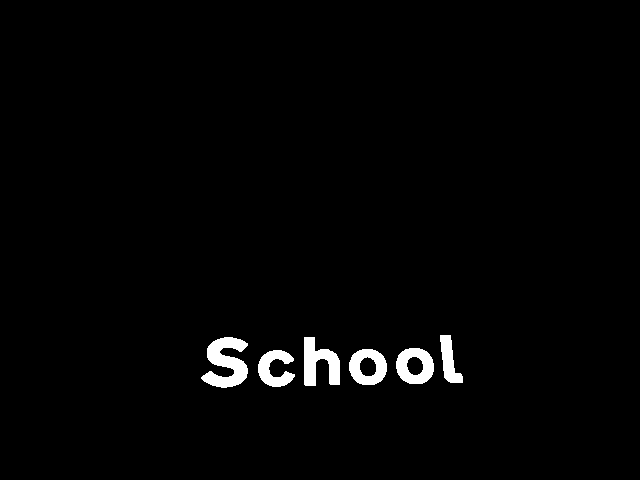} \ &
\includegraphics[width=0.07\linewidth]{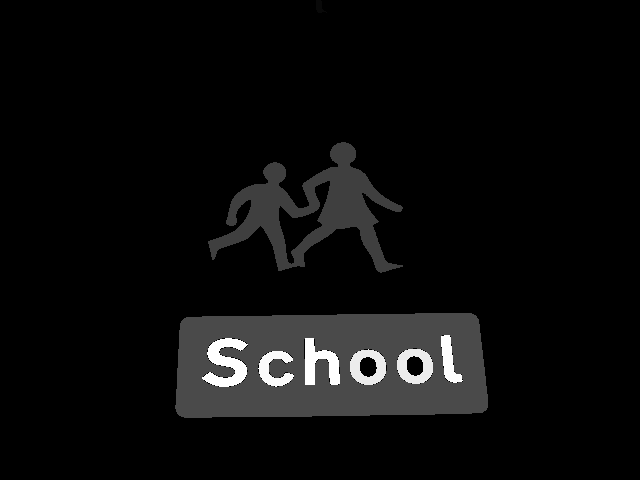} \ &
\includegraphics[width=0.07\linewidth]{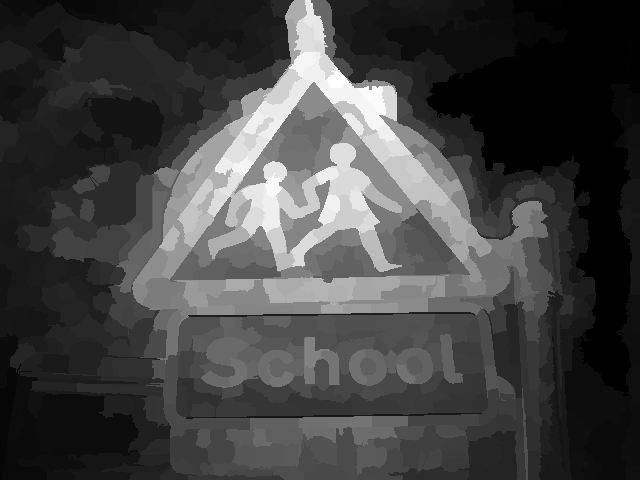} \ &
\includegraphics[width=0.07\linewidth]{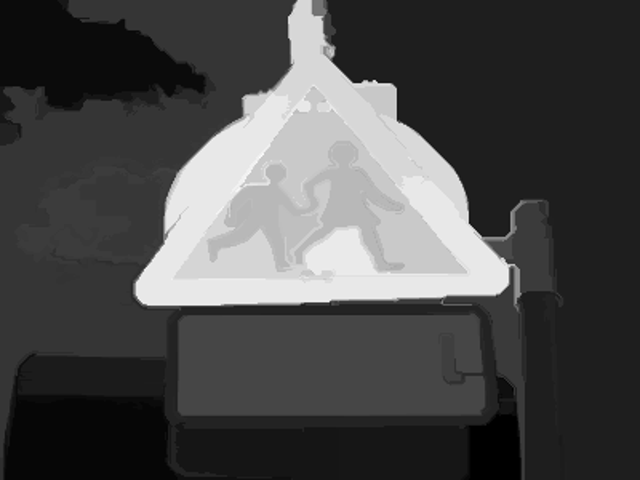} \ &
\includegraphics[width=0.07\linewidth]{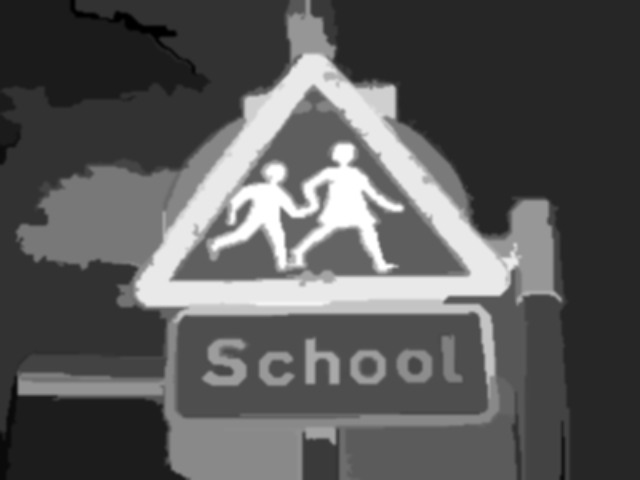} \ &
\includegraphics[width=0.07\linewidth]{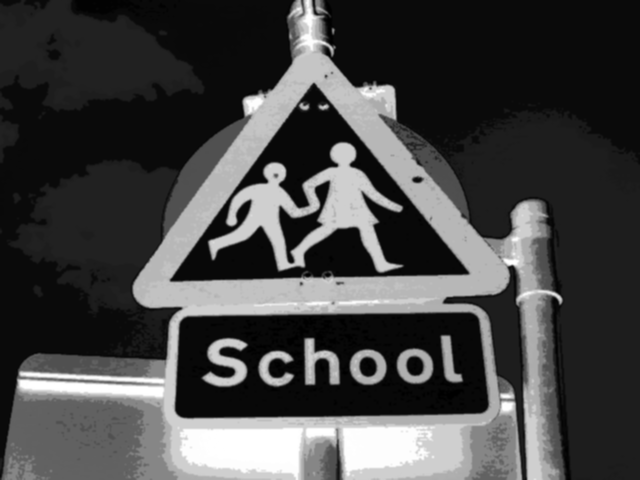} \ &
\includegraphics[width=0.07\linewidth]{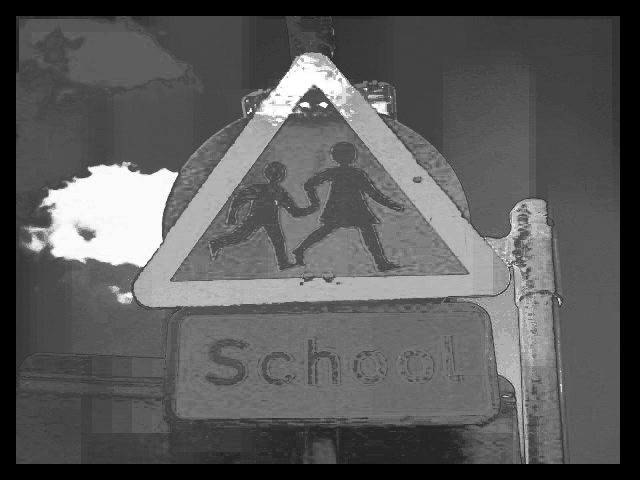} \ &
\includegraphics[width=0.07\linewidth]{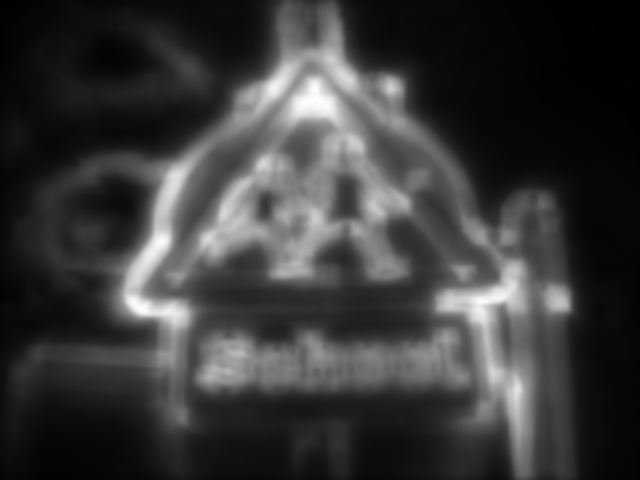} \ &
\includegraphics[width=0.07\linewidth]{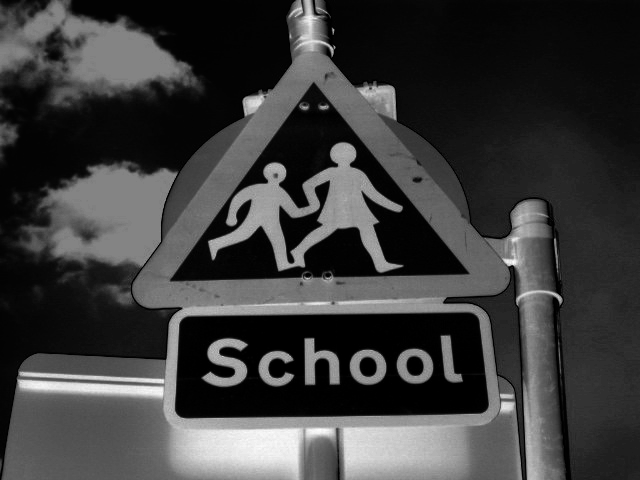} \ &
\includegraphics[width=0.07\linewidth]{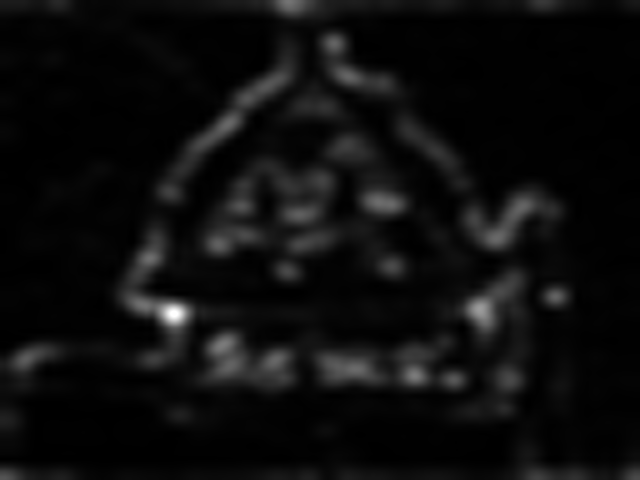} \ &
\includegraphics[width=0.07\linewidth]{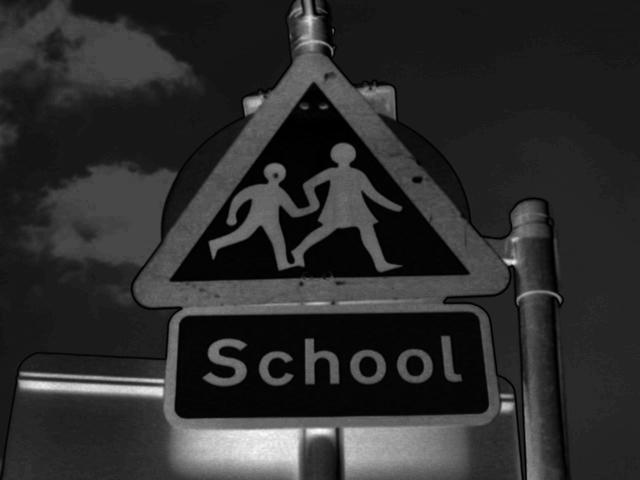} \ &
\includegraphics[width=0.07\linewidth]{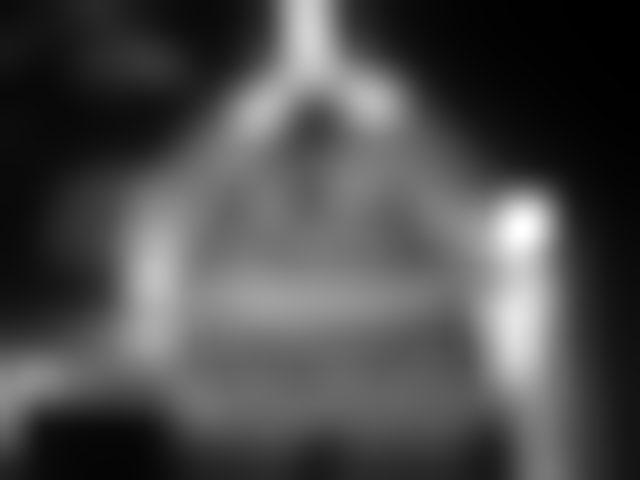} \\
\includegraphics[width=0.07\linewidth]{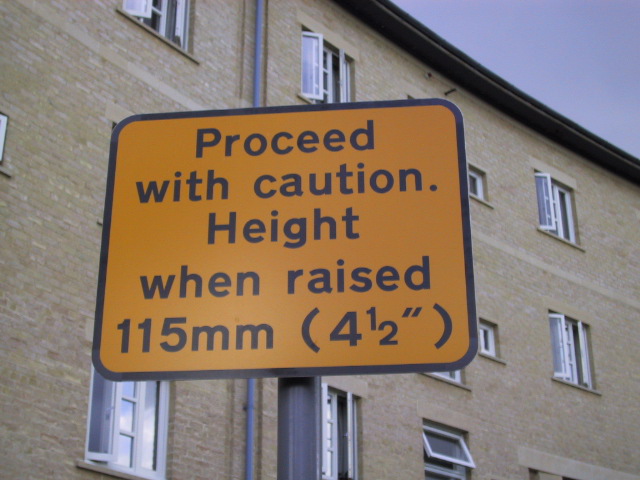} \ &
\includegraphics[width=0.07\linewidth]{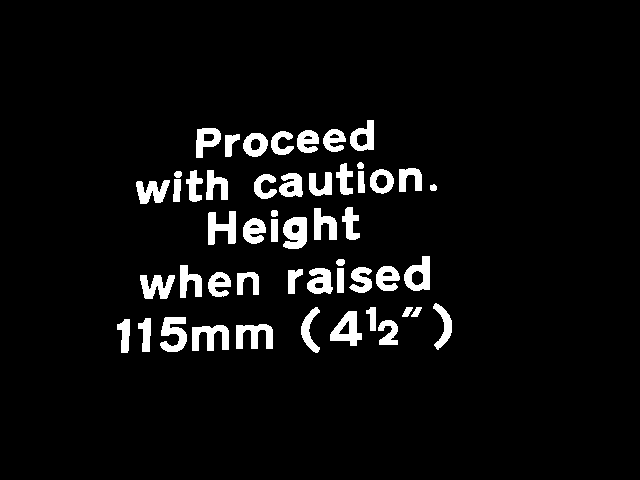} \ &
\includegraphics[width=0.07\linewidth]{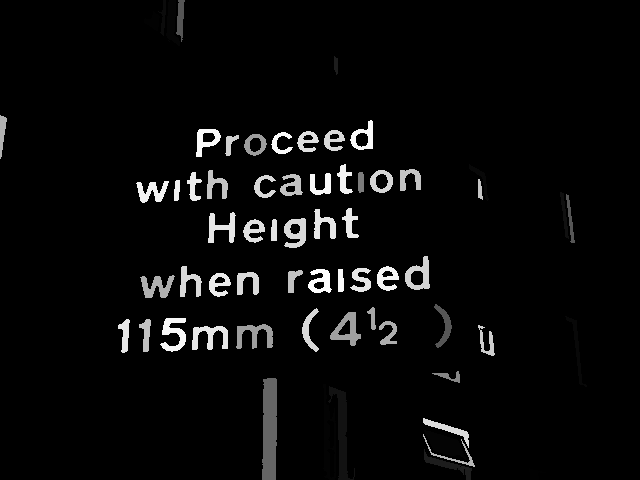} \ &
\includegraphics[width=0.07\linewidth]{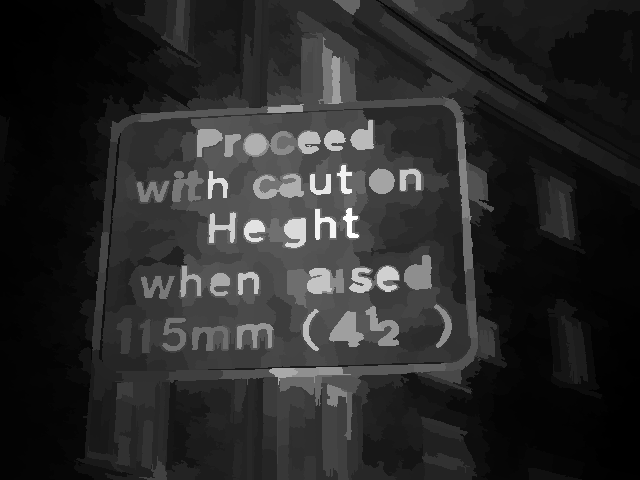} \ &
\includegraphics[width=0.07\linewidth]{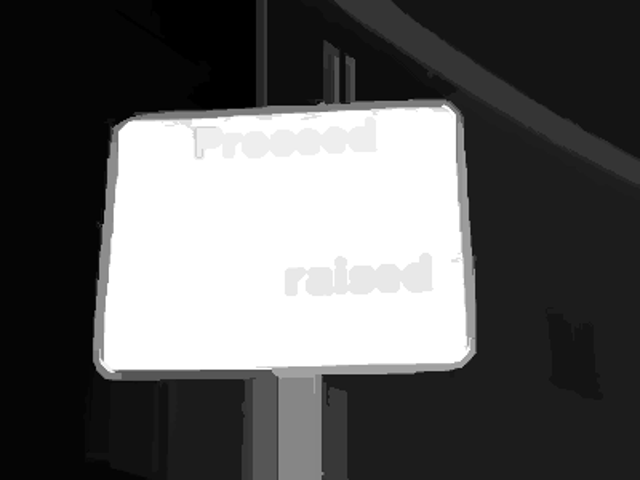} \ &
\includegraphics[width=0.07\linewidth]{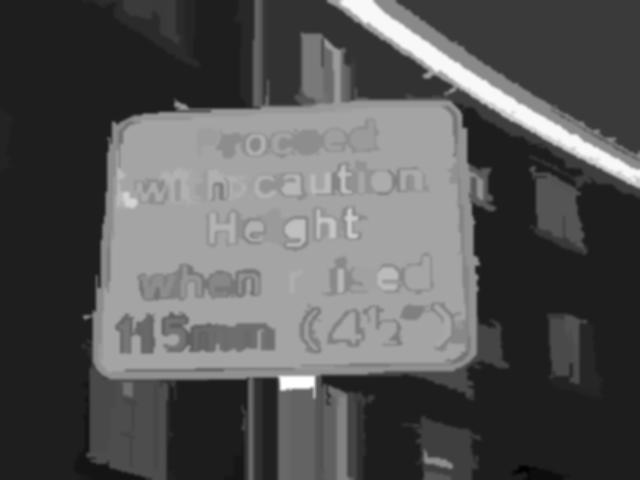} \ &
\includegraphics[width=0.07\linewidth]{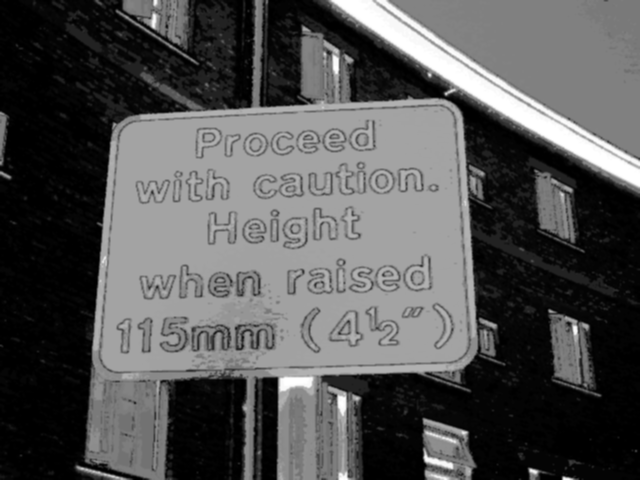} \ &
\includegraphics[width=0.07\linewidth]{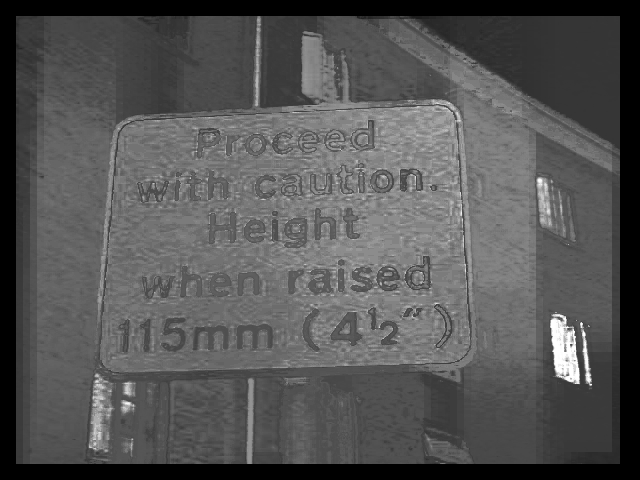} \ &
\includegraphics[width=0.07\linewidth]{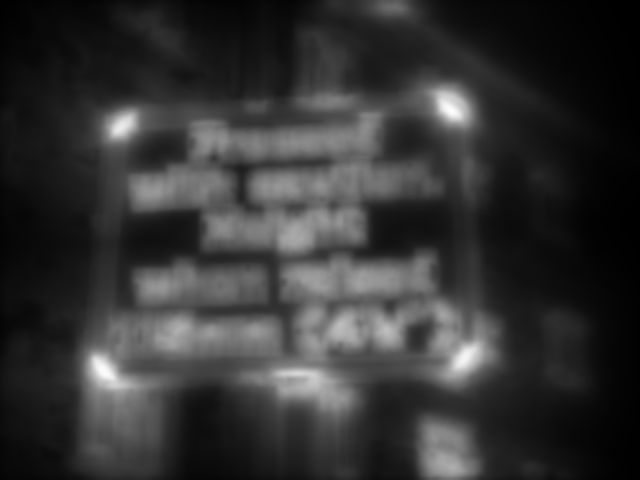} \ &
\includegraphics[width=0.07\linewidth]{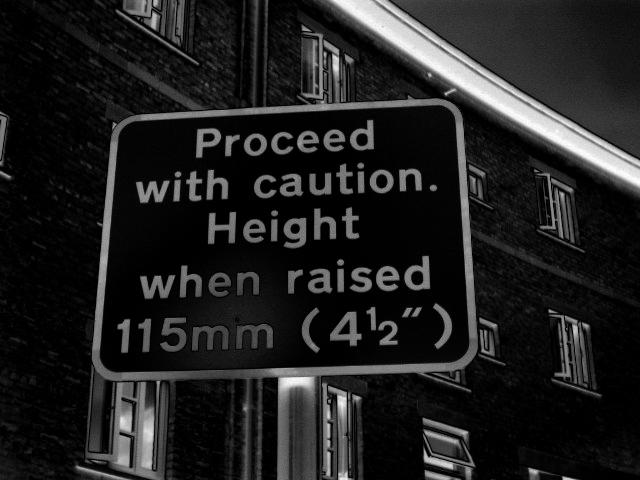} \ &
\includegraphics[width=0.07\linewidth]{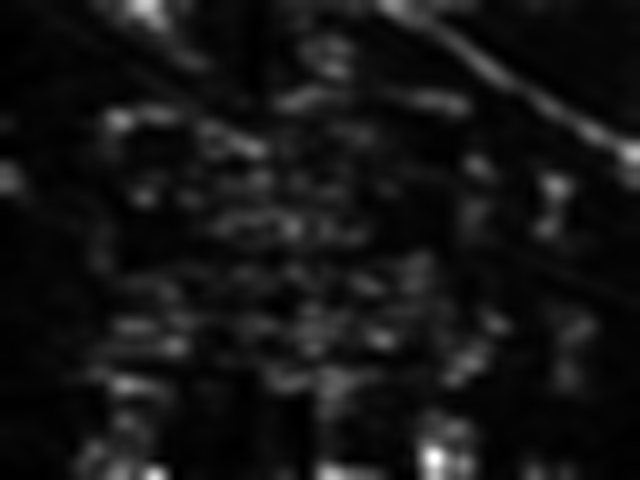} \ &
\includegraphics[width=0.07\linewidth]{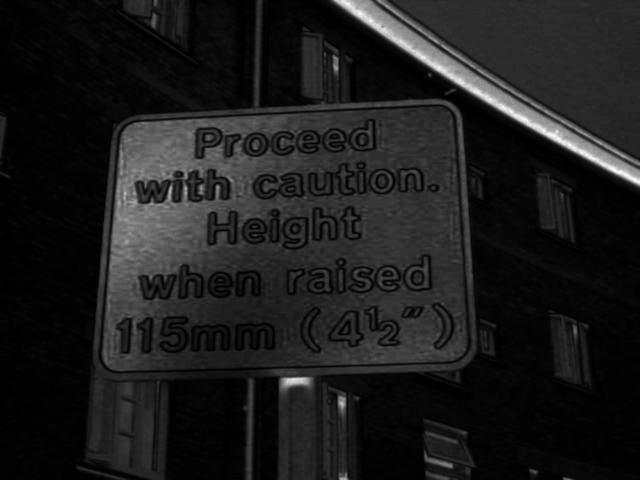} \ &
\includegraphics[width=0.07\linewidth]{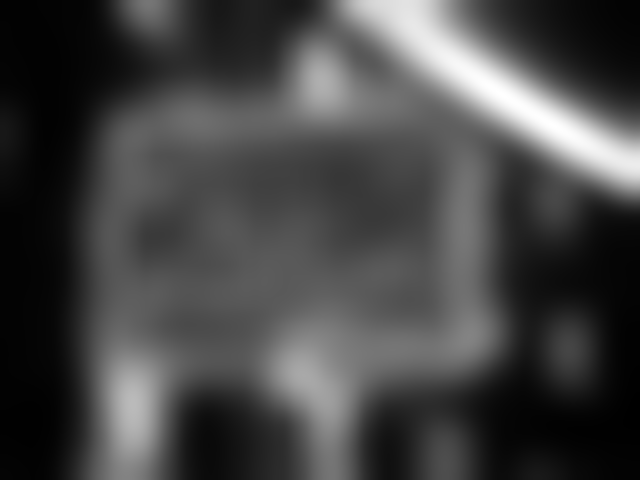} \\
\includegraphics[width=0.07\linewidth]{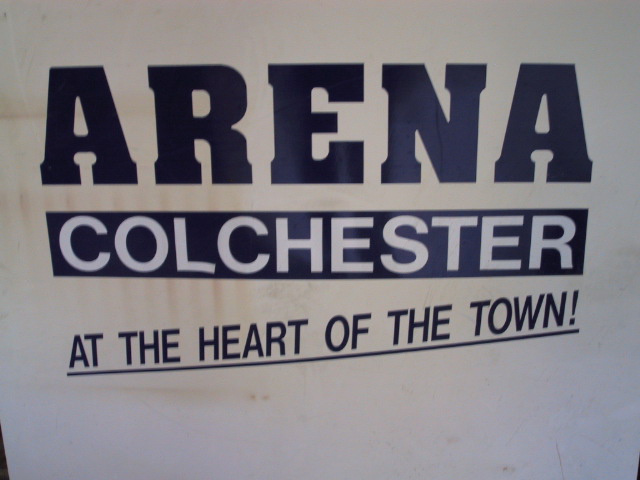} \ &
\includegraphics[width=0.07\linewidth]{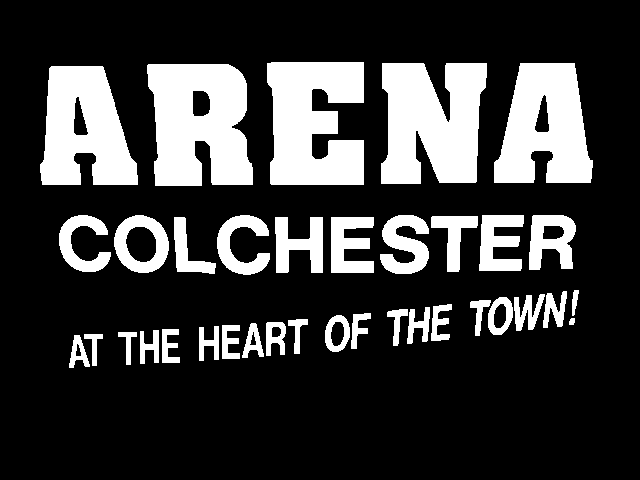} \ &
\includegraphics[width=0.07\linewidth]{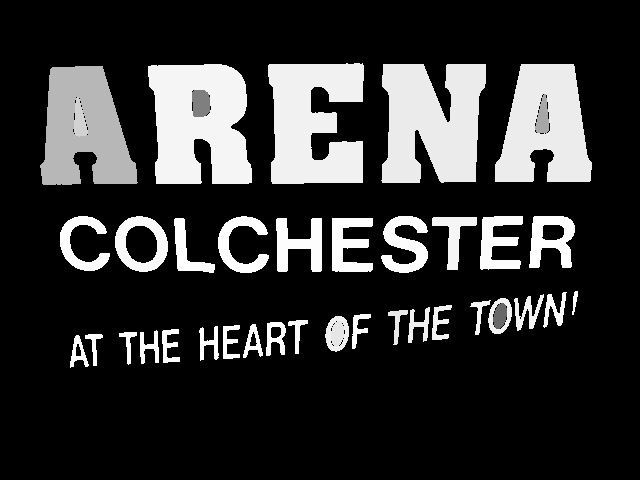} \ &
\includegraphics[width=0.07\linewidth]{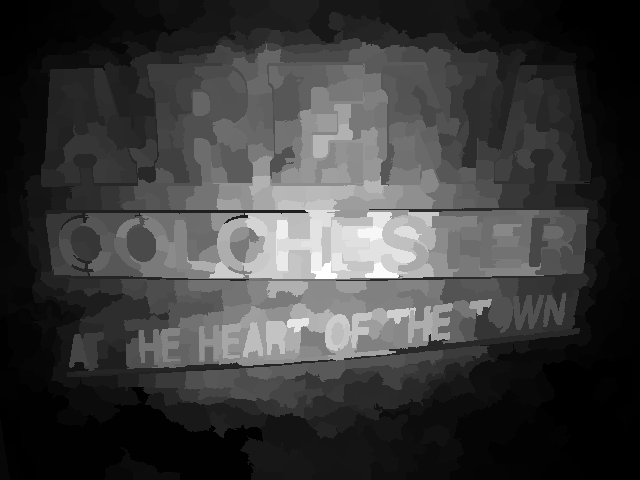} \ &
\includegraphics[width=0.07\linewidth]{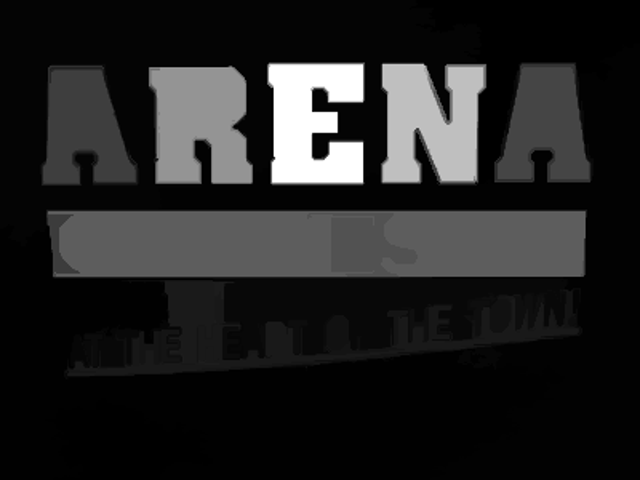} \ &
\includegraphics[width=0.07\linewidth]{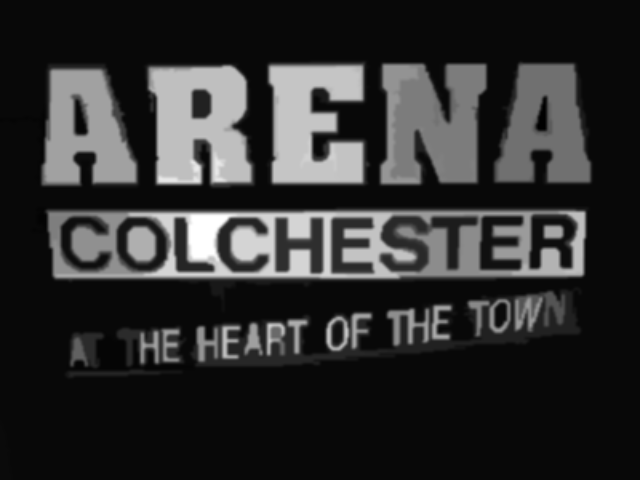} \ &
\includegraphics[width=0.07\linewidth]{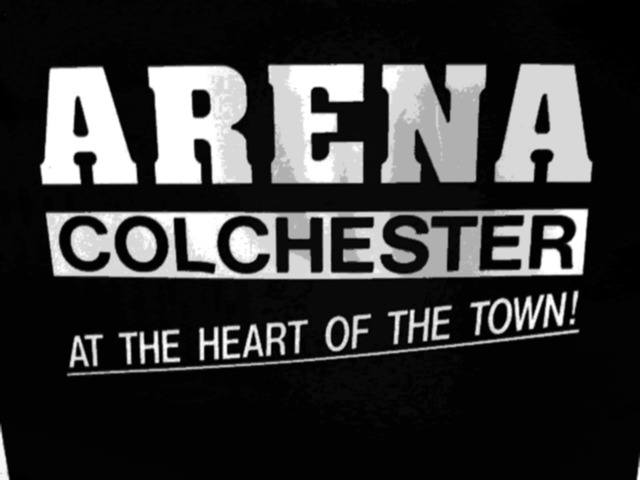} \ &
\includegraphics[width=0.07\linewidth]{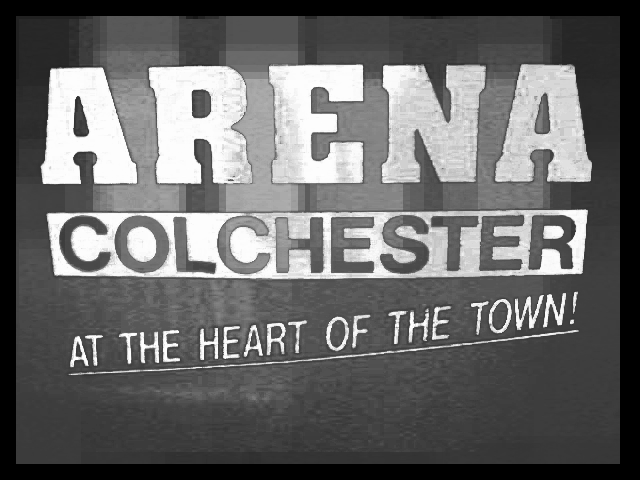} \ &
\includegraphics[width=0.07\linewidth]{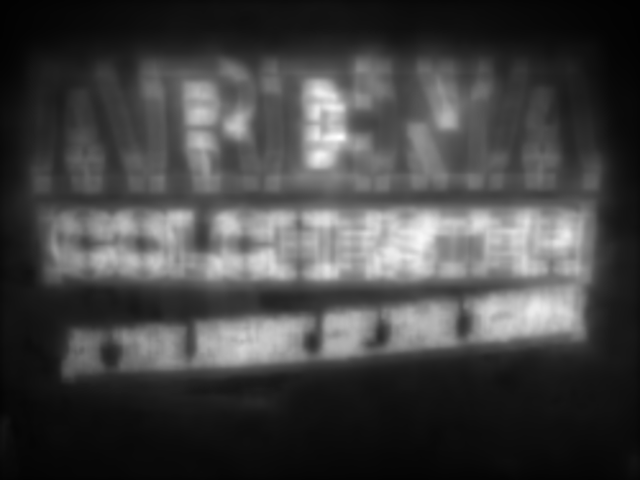} \ &
\includegraphics[width=0.07\linewidth]{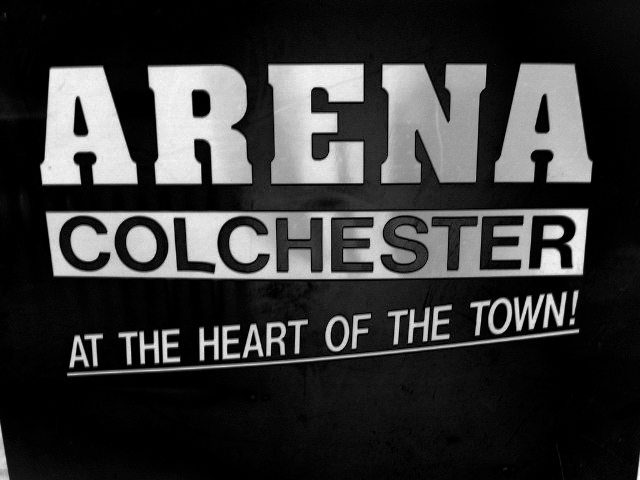} \ &
\includegraphics[width=0.07\linewidth]{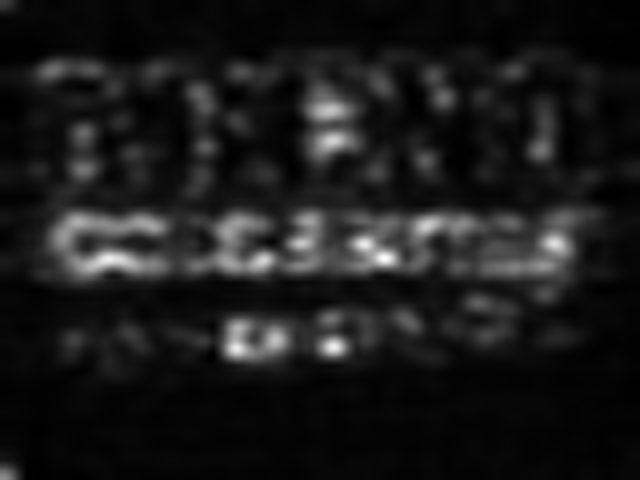} \ &
\includegraphics[width=0.07\linewidth]{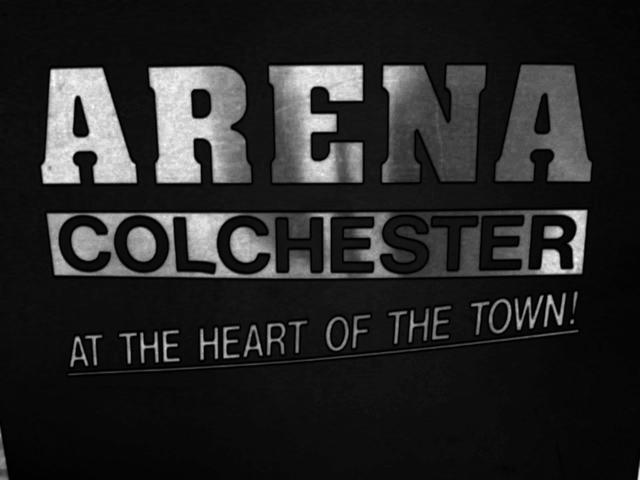} \ &
\includegraphics[width=0.07\linewidth]{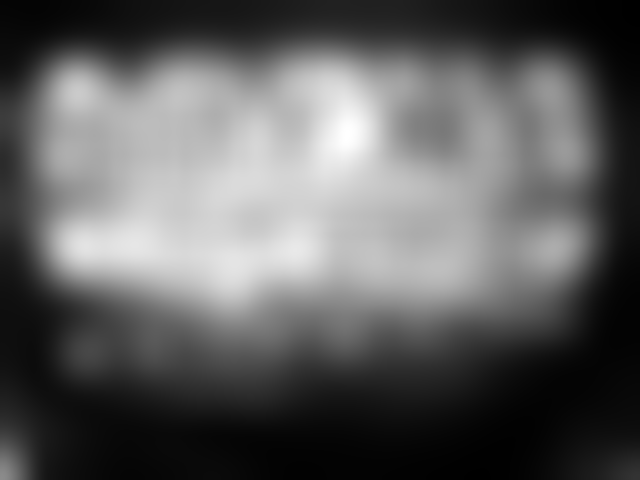} \\
\includegraphics[width=0.07\linewidth]{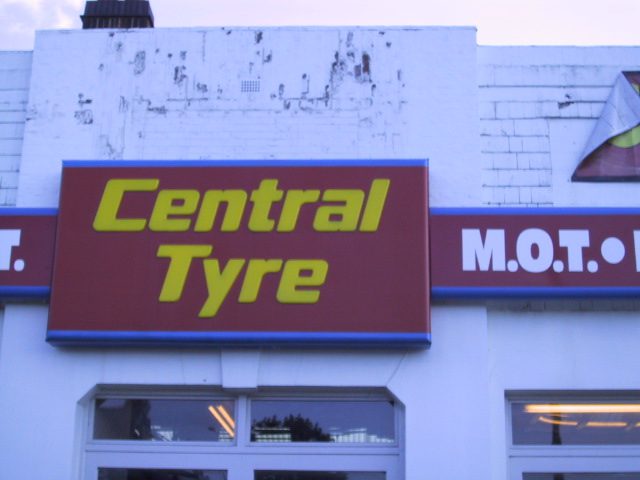} \ &
\includegraphics[width=0.07\linewidth]{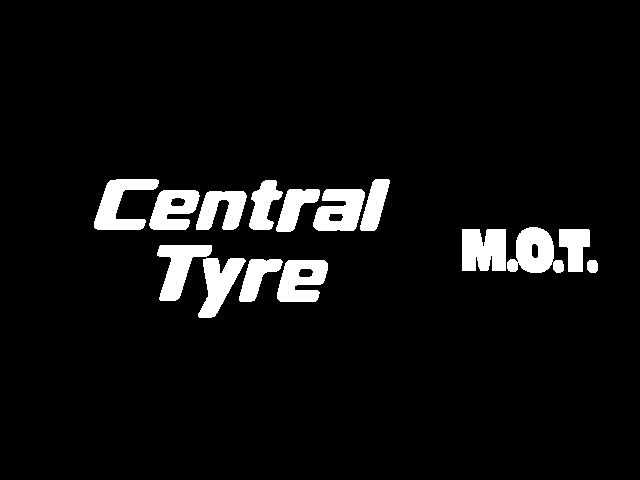} \ &
\includegraphics[width=0.07\linewidth]{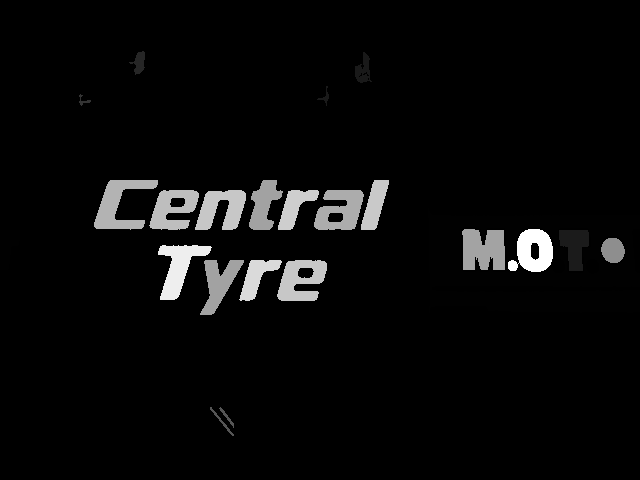} \ &
\includegraphics[width=0.07\linewidth]{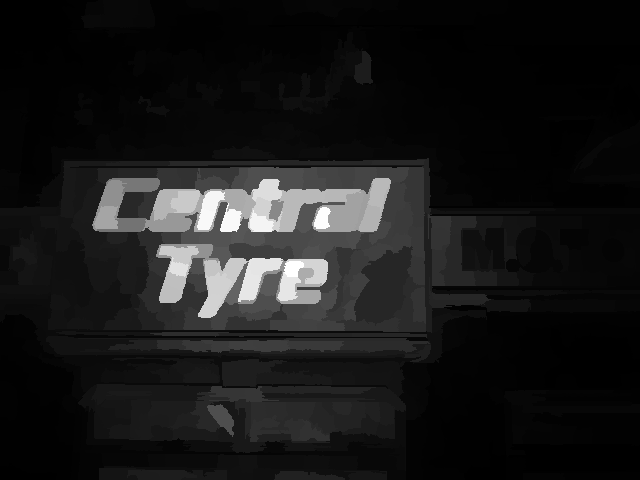} \ &
\includegraphics[width=0.07\linewidth]{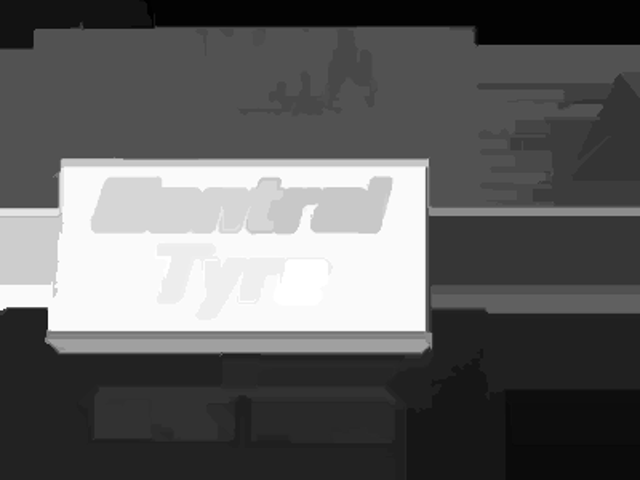} \ &
\includegraphics[width=0.07\linewidth]{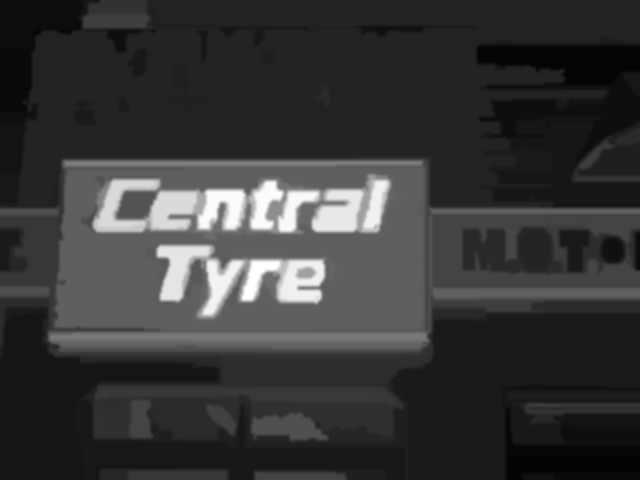} \ &
\includegraphics[width=0.07\linewidth]{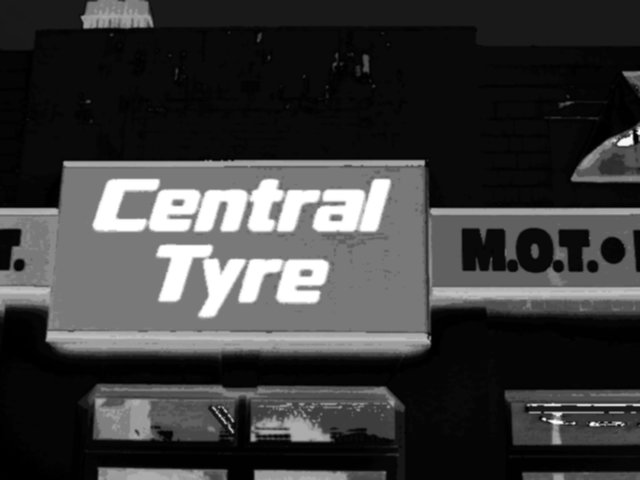} \ &
\includegraphics[width=0.07\linewidth]{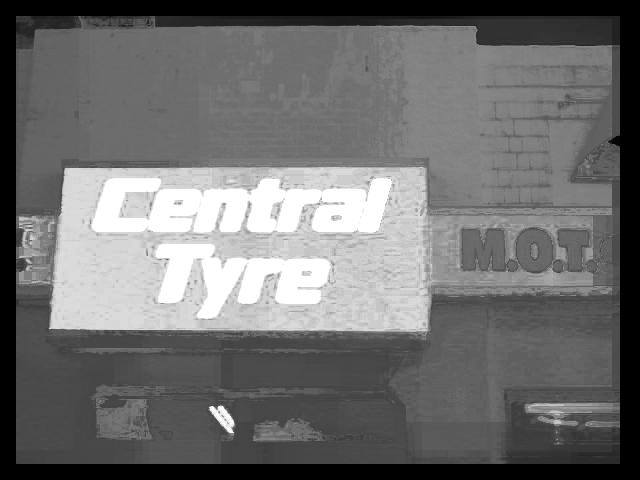} \ &
\includegraphics[width=0.07\linewidth]{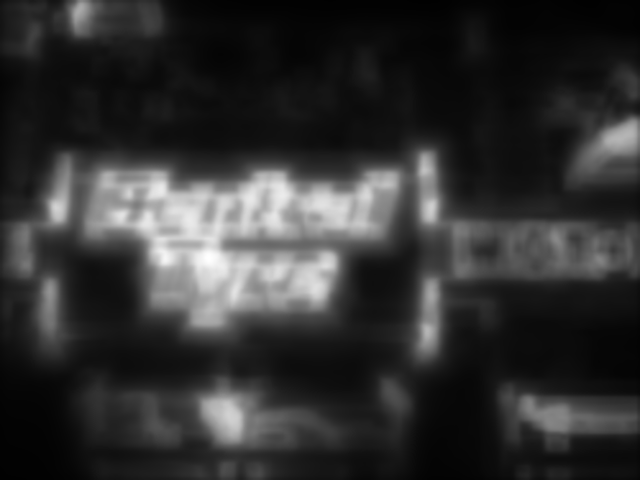} \ &
\includegraphics[width=0.07\linewidth]{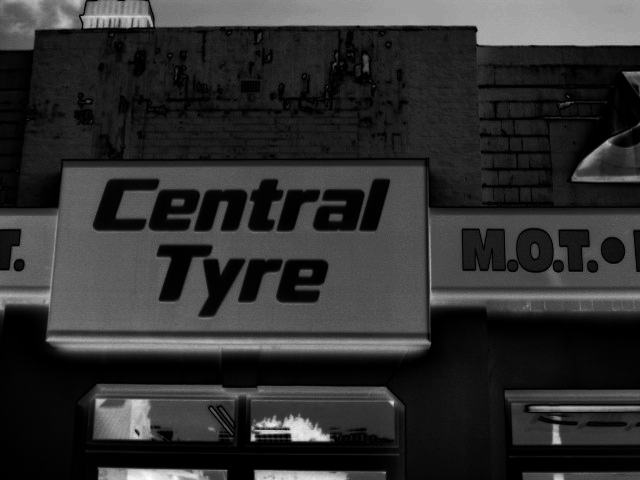} \ &
\includegraphics[width=0.07\linewidth]{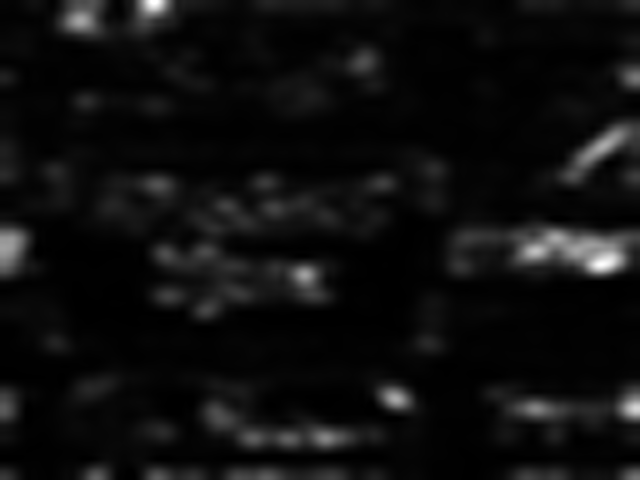} \ &
\includegraphics[width=0.07\linewidth]{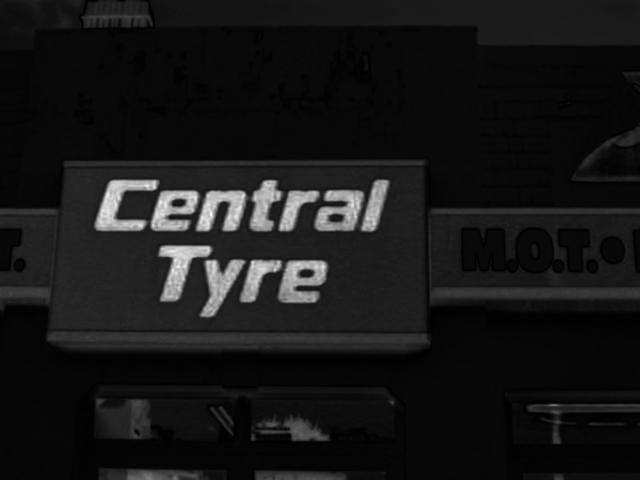} \ &
\includegraphics[width=0.07\linewidth]{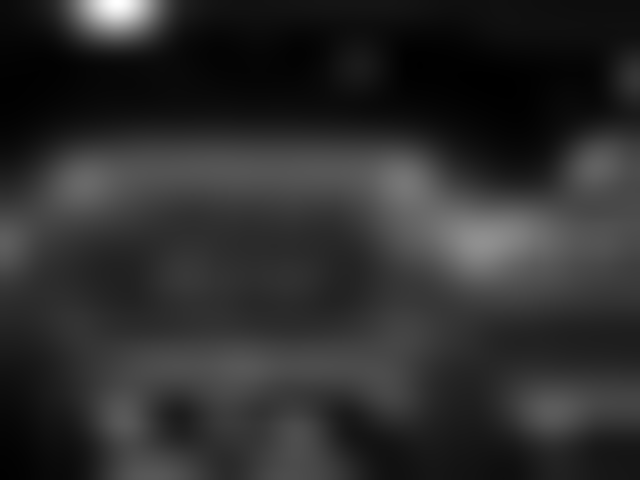} \\
\includegraphics[width=0.07\linewidth]{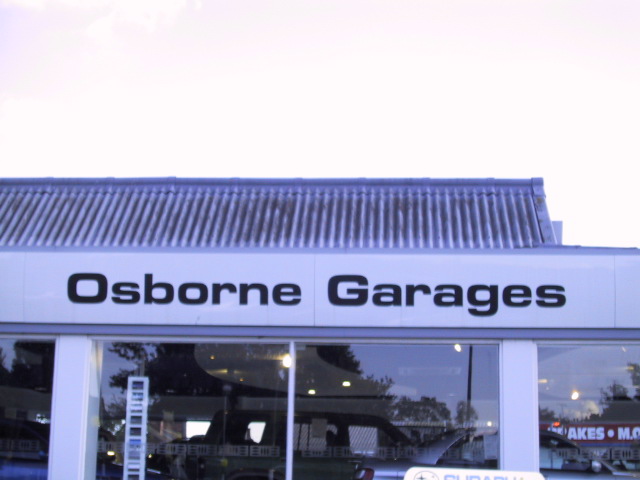} \ &
\includegraphics[width=0.07\linewidth]{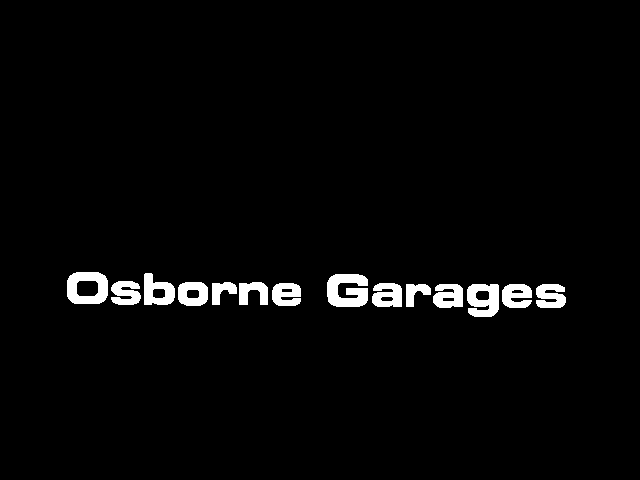} \ &
\includegraphics[width=0.07\linewidth]{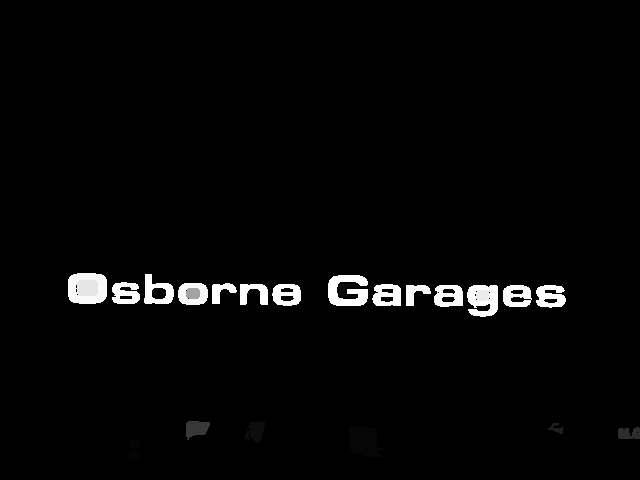} \ &
\includegraphics[width=0.07\linewidth]{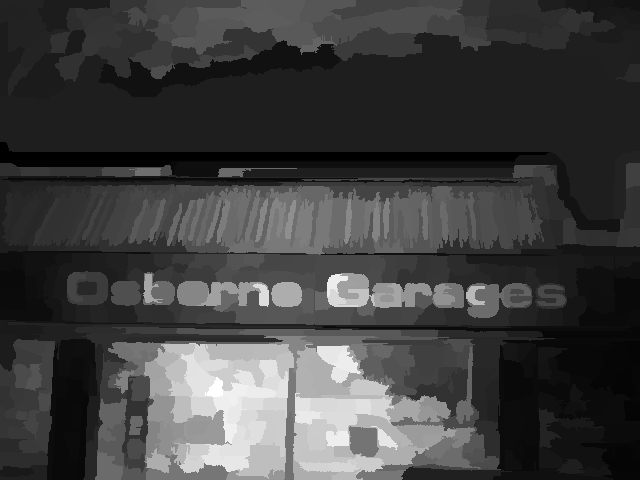} \ &
\includegraphics[width=0.07\linewidth]{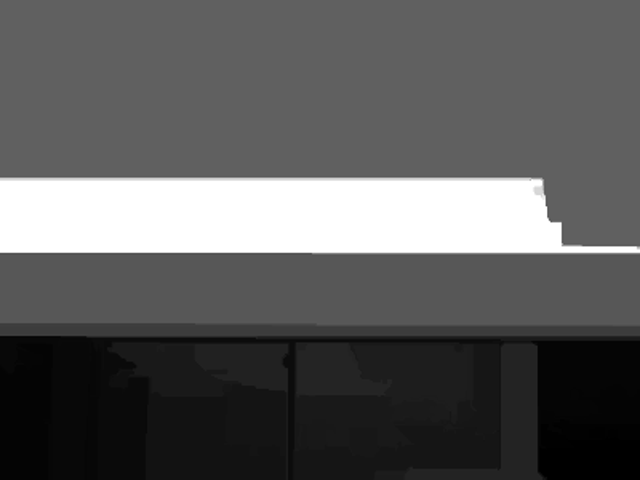} \ &
\includegraphics[width=0.07\linewidth]{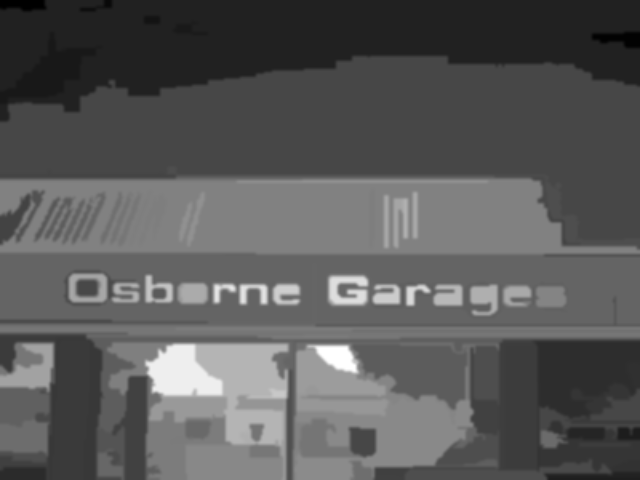} \ &
\includegraphics[width=0.07\linewidth]{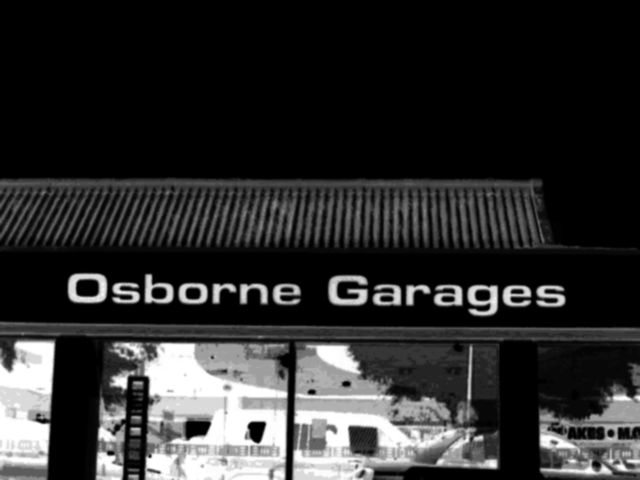} \ &
\includegraphics[width=0.07\linewidth]{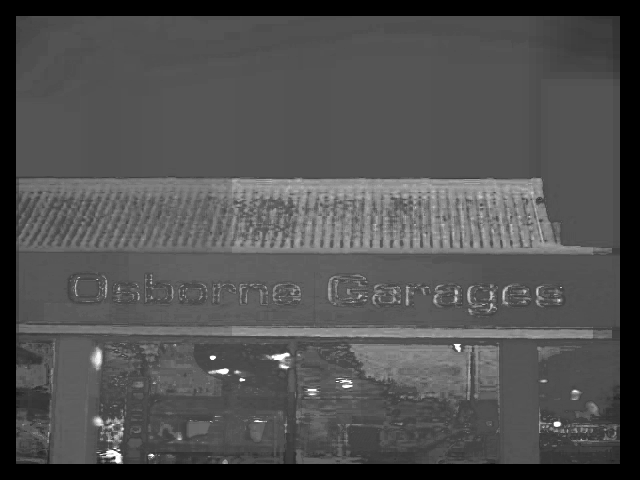} \ &
\includegraphics[width=0.07\linewidth]{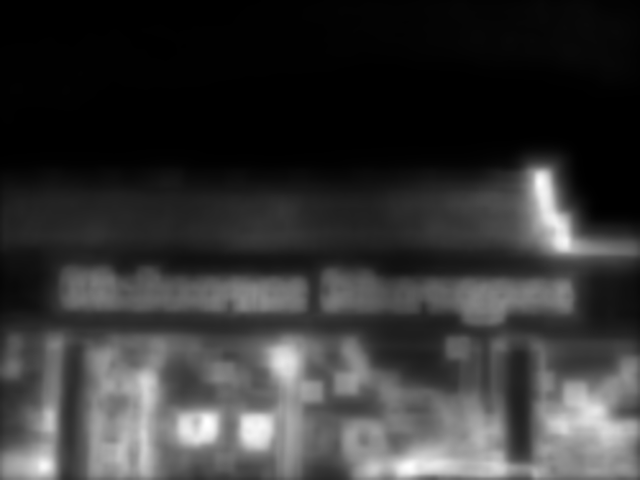} \ &
\includegraphics[width=0.07\linewidth]{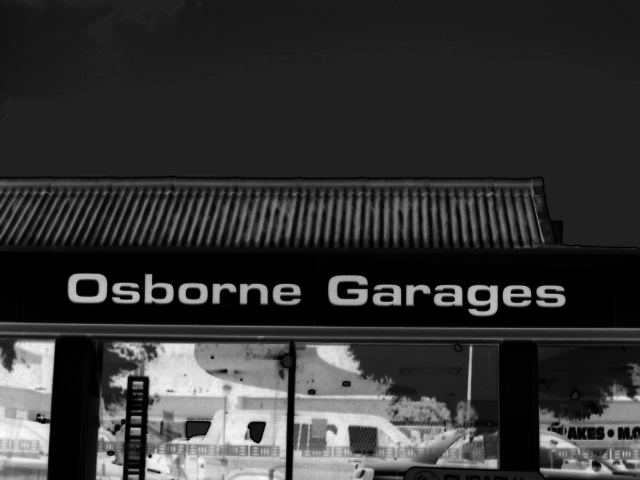} \ &
\includegraphics[width=0.07\linewidth]{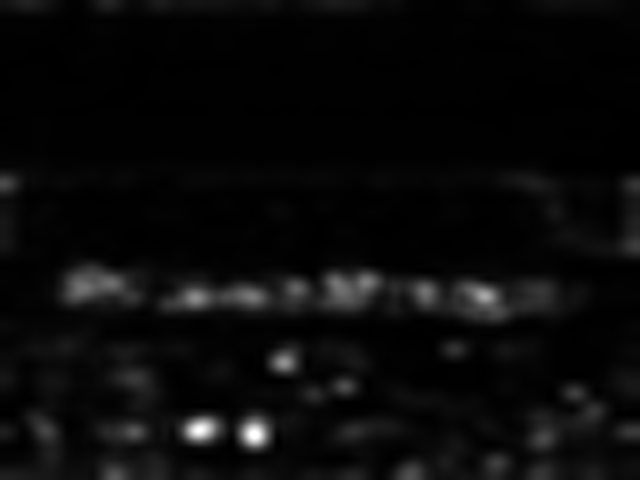} \ &
\includegraphics[width=0.07\linewidth]{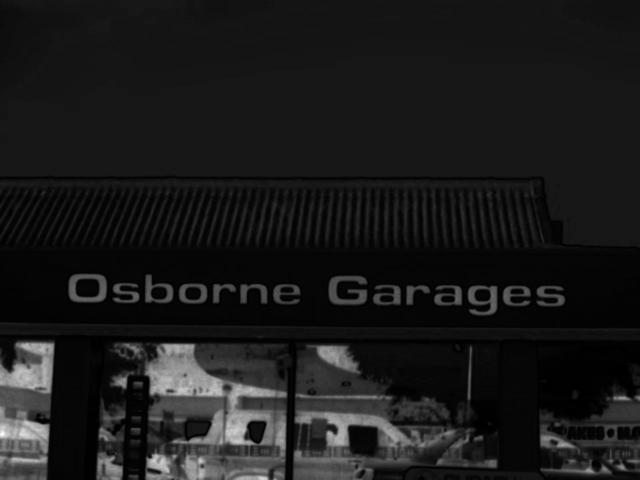} \ &
\includegraphics[width=0.07\linewidth]{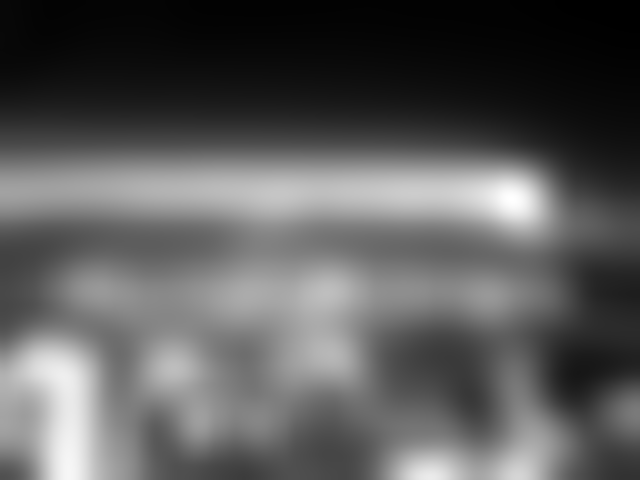} \\
\includegraphics[width=0.07\linewidth]{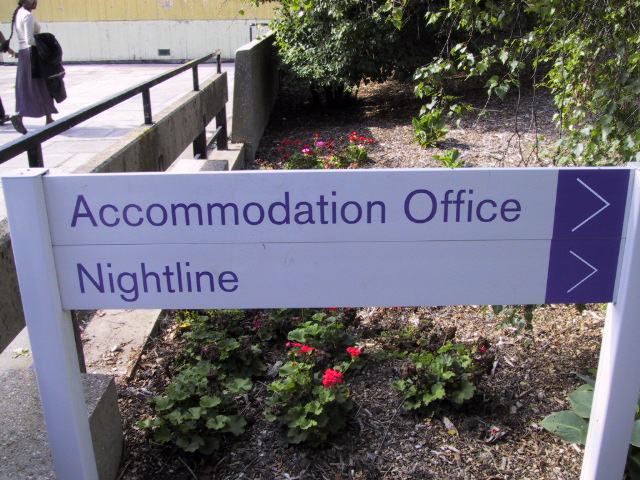} \ &
\includegraphics[width=0.07\linewidth]{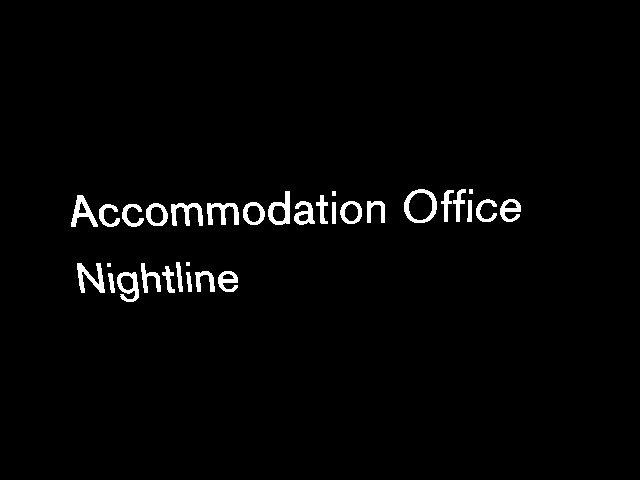} \ &
\includegraphics[width=0.07\linewidth]{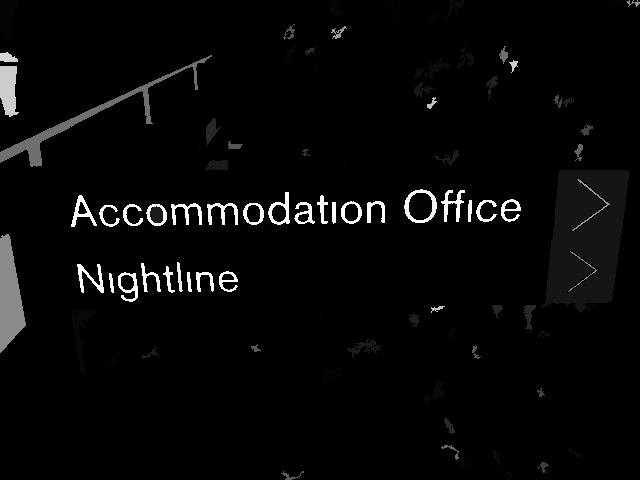} \ &
\includegraphics[width=0.07\linewidth]{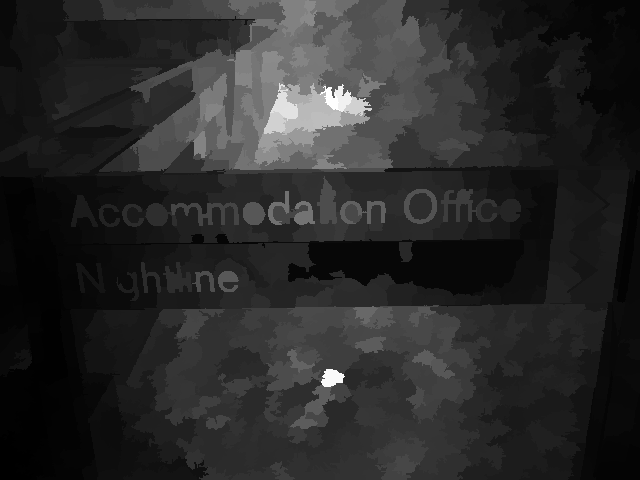} \ &
\includegraphics[width=0.07\linewidth]{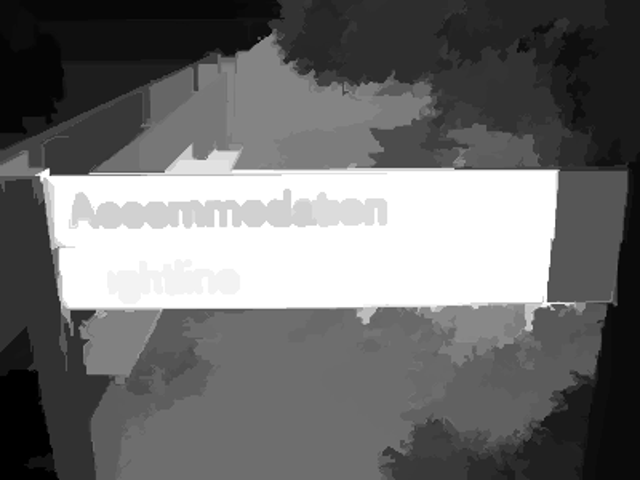} \ &
\includegraphics[width=0.07\linewidth]{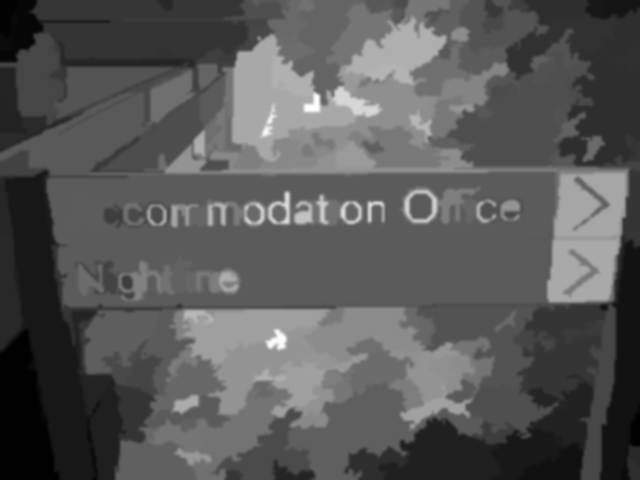} \ &
\includegraphics[width=0.07\linewidth]{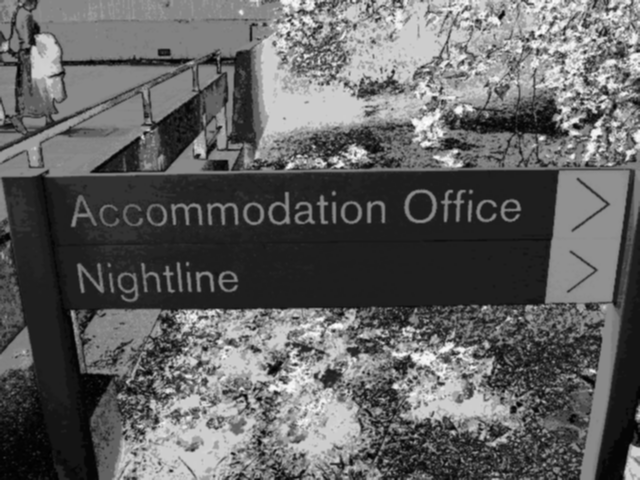} \ &
\includegraphics[width=0.07\linewidth]{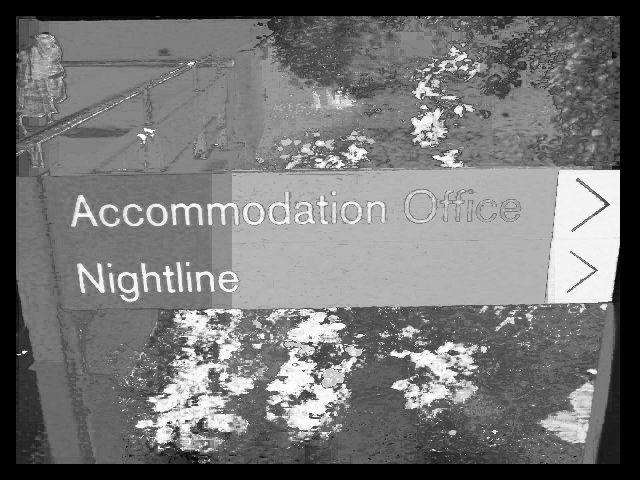} \ &
\includegraphics[width=0.07\linewidth]{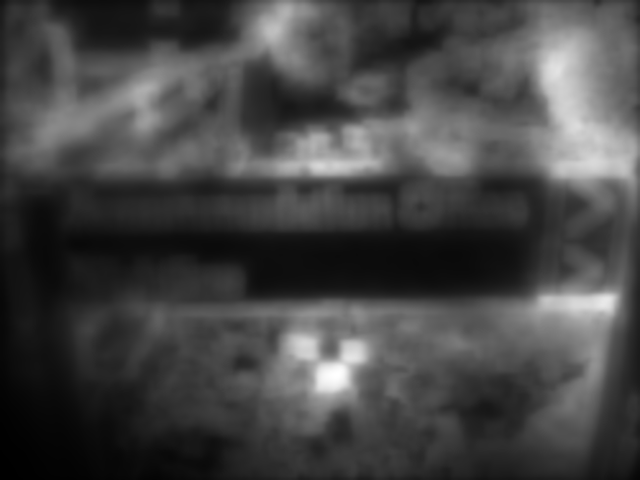} \ &
\includegraphics[width=0.07\linewidth]{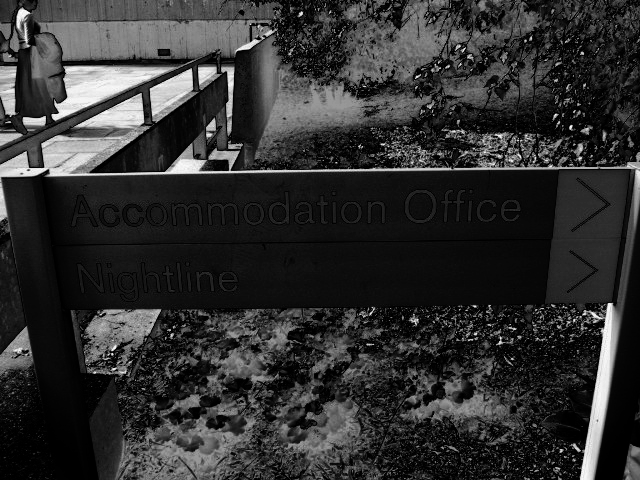} \ &
\includegraphics[width=0.07\linewidth]{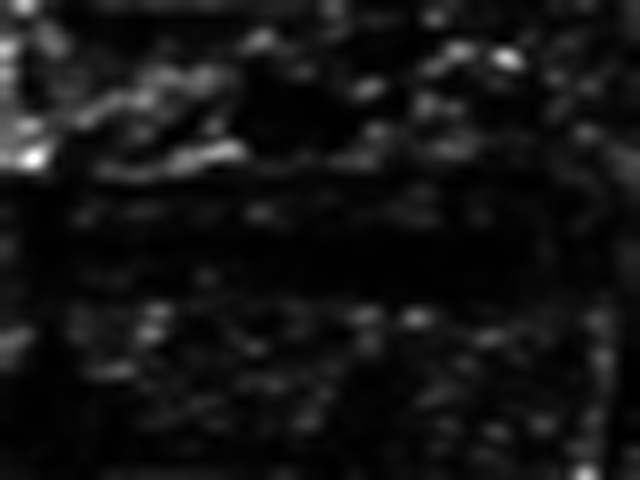} \ &
\includegraphics[width=0.07\linewidth]{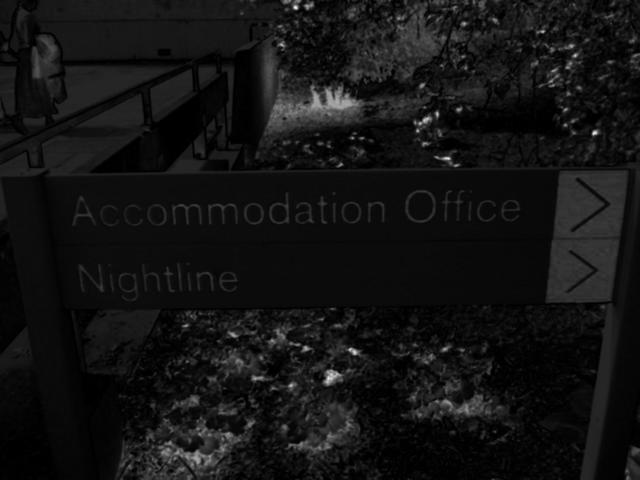} \ &
\includegraphics[width=0.07\linewidth]{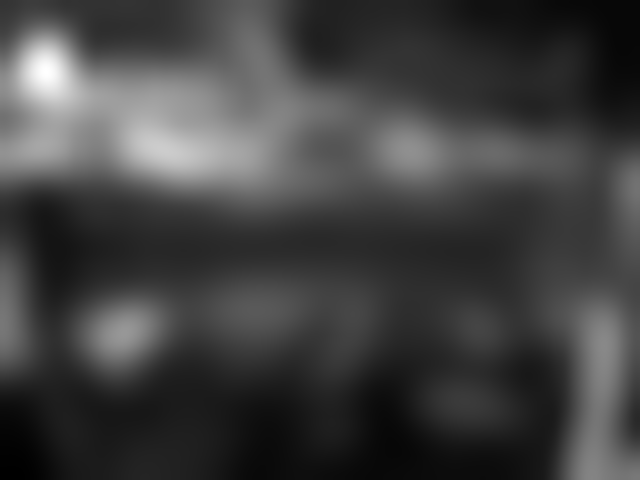} \\
\end{tabular}
\caption{Visual comparison of saliency maps. Clearly, the proposed method highlights
characters as salient regions whereas state-of-the-art saliency detection algorithms may be attracted by other stuff in the scene.}
\label{fig:CharacternessMap}
\end{figure*}

As it shows in Fig.~\ref{fig:PRcurve},
in terms of the PR curve, all existing saliency detection models, including three best saliency detection models in~\cite{DBLP:conf/eccv/BorjiSI12} achieve low precision rate (below 0.5) in most cases when the recall rate is fixed.
However, the proposed characterness model produces significantly better results, indicating that our model is more suitable for the measurement of characterness.
The straight line segment of our PR curve (when recall rates ranging from 0.67 to 1) is attributed to the fact only foreground regions extracted by eMSER are considered as character candidates, thus background regions always have a zero saliency value.
It can also be observed from the PR curve that in our scenario, two best existing saliency detection models are RC and LR.

Precision, recall and F-measure
computed via adaptive threshold are illustrated in the right sub-figure of Fig.~\ref{fig:PRcurve}. Our result significantly outperforms other saliency detection models
in all three criteria, which indicates that our approach consistently produces results closer to ground truth.

Table \ref{tal:overlapratio} illustrates the performance of all approaches measured by VOC overlap score.
As it shows, our result is almost twice that of the best saliency detection model LR on this task.

Fig.~\ref{fig:CharacternessMap} shows some saliency maps of all approaches.
It is observed that our approach has obtained visually more feasible results than other approaches have: it usually gives high saliency values to characters while suppressing
non-characters, whereas the state-of-the-art saliency detection models may be attracted by other objects in the natural scene (\emph{e.g.}, sign boards are also considered as salient objects in CB).

In summary, both quantitative and qualitative evaluation demonstrate that the proposed
characterness model significantly outperforms saliency detection approaches on this task.

\section{Proposed Scene Text Detection Approach Evaluation}
\label{sec:SceneTextDetectionEvaluation}
In this section, we evaluate our scene text detection approach as a whole.
Same as previous work on scene text detection, we use the detected bounding boxes to evaluate performance and compare with state-of-the-art approaches.
Compared with Sec. \ref{sec:CharacternessEvaluation} in which only 119 images are utilized to learn the distribution of cues, all images with pixel-level ground truth (229 images) are adopted here, thus
the distribution is closer to the real scene.

From the large body of work on scene text detection, we compare our result with some state-of-the-art approaches, including
TD method~\cite{DBLP:conf/cvpr/Yao}, Epshtein's method~\cite{DBLP:conf/cvpr/EpshteinOW10}, Li's method~\cite{DBLP:conf/icpr/LiL12,DBLP:conf/icip/Li13}, Yi's method~\cite{DBLP:journals/tip/YiT11}, Meng's method~\cite{DBLP:conf/das/MengS12},
Neumann's method~\cite{DBLP:conf/accv/NeumannM10,DBLP:conf/cvpr/NeumannM10}, Zhang's method~\cite{DBLP:conf/icpr/ZhangK10a} and some approaches presented in the ICDAR competitions.
Note that the bandwidth of mean shift clustering in the text line
formulation step was set to 2.2 in all experiments.

\subsection{Datasets}
We have conducted comprehensive experimental evaluation on three
publicly available datasets.
Two of them are from the benchmark ICDAR robust reading competition held in different years, namely ICDAR 2003~\cite{DBLP:conf/icdar/LucasPSTWY03} and ICDAR 2011~\cite{DBLP:conf/icdar/ShahabSD11a}.
ICDAR 2003
dataset contains 258 images for
training and 251 images for testing.
This dataset was also adopted in the ICDAR 2005~\cite{DBLP:conf/icdar/Lucas05} competition.
ICDAR 2011 dataset contains two folds of data, one for training with 229
images, and the other one for testing with 255 images.
To evaluate the effectiveness of the proposed algorithm on text in arbitrary orientations, we also adopted the Oriented Scene Text Database (OSTD)~\cite{DBLP:journals/tip/YiT11} in our experiments.
The dataset set contains 89 images with text lies in in arbitrary orientations.

\subsection{Evaluation criteria}
According to literature review, precision, recall and f-measure are the most popularly adopted criteria used to evaluate scene text detection approaches.
However, definition of the three criteria are slightly different across datasets.

In the ICDAR 2003 and 2005 competition, precision and recall are computed by finding the best match between each detected bounding boxes $|D|$ and each ground truth $|G|$. In this sense, only one-to-one matches are taken into account.
To overcome this unfavorable fact, ICDAR 2011 competition adopts the DetEval software~\cite{DBLP:journals/ijdar/WolfJ06} which supports one-to-one matches,
one-to-many matches and many-to-one matches.
For the OSTD dataset, we use the original definition of precision and recall from the authors~\cite{DBLP:journals/tip/YiT11}, which are based on computing the size of overlapping areas between $|D|$ and $|G|$.
In all three datasets, f-measure is always defined as the harmonic mean of precision and recall.
\subsection{eMSER versus MSER}
Since the proposed characterness cues are computed on regions, the extraction of informative regions is a prerequisite for the robustness of our approach.
To demonstrate that the modified eMSER algorithm improves the performance,
we compare it with the original MSER algorithm on the ICDAR 2011 dataset.
For fair comparison, when learning the distribution of cues on negative samples, we use MSER rather than eMSER to harvest negative samples and then compute the three cues.
Other parts of our approach remain fixed.

Using the MSER algorithm achieves a recall of 66\%, a precision of 67\% and an f-measure of 66\%.
In comparison, when the eMSER is adopted, the precision rate is boosted significantly (80\%), leading to an improved f-measure (70\%).
This is owing to that eMSER is capable of preserving shape of regions, whereas regions extracted by MSER are more likely to be blurred which makes cues less effective.

\subsection{Evaluation of characterness cues}
\begin{table}[t]
\centering
\caption{Evaluation of characterness cues on the ICDAR 2011 dataset.}
\label{tal:cues}
\scalebox{1.2}{
\begin{tabular}{l||c|c|c}
\hline
Cues & precision & recall & f-measure\\
\hline
SW & 0.71 & 0.63 & 0.67\\
\hline
PD & 0.64 & 0.63 & 0.63\\
\hline
eHOG & 0.58 & 0.65 & 0.61\\
\hline
SW+PD & 0.78 & 0.63 & 0.68\\
\hline
SW+eHOG & 0.74 & 0.63 & 0.68\\
\hline
PD+eHOG & 0.73 & 0.63 & 0.67\\
\hline
{\bf SW+PD+eHOG} & \bf{0.80} & \bf{0.62} & \bf{0.70}\\
\hline
Baseline & 0.53 & 0.67 & 0.59\\
\hline
\end{tabular}}
\end{table}
The proposed characterness cues (\emph{i.e.} SW, PD and eHOG) are critical to the characternss model and the final text detection result.
In order to show that they are effective in distinguishing characters and non-characters, we evaluate different combinations of the cues
on the ICDAR 2011 dataset.
Table~\ref{tal:cues} shows the evaluation via precision, recall and f-measure.
Clearly, the table shows an upward trend in performance with increasing number of cues.
Note that the baseline method in Table~\ref{tal:cues} corresponds to the result obtained by directly preforming text line formulation after candidate region extraction.

As it shows in Table~\ref{tal:cues}, the performance of the proposed approach is generally poorer when only one cue is adopted.
However, the f-measures are still much higher than the baseline method, which indicates that individual cues are effective.
We also notice that the SW cue shows the best f-measure when individual cue is considered.
This indicates that characters and non-characters are much easier to be separated by using the SW cue.
From Table~\ref{tal:cues}, we can easily conclude that the order of discriminability of individual cues (from high to low) is: SW, PD and eHOG.

The performance of the proposed approach is boosted by a large extent
(about 5\% in f-measure) when two cues are combined, which attributes to the significant increase in precision.

As expected, the pest performance is achieved when all cues are combined.
Although there is a slightly drop in recall rate, precision rate (80\%) is significantly higher than all other combinations, thus the f-measure is the best.

\subsection{Comparison with other approaches}
\begin{table}
\caption{Results on ICDAR 2003 dataset.}
\label{tal:2003}
\centering
\scalebox{1.2}{
\begin{tabular}{ l||c|c|c }
\hline
method & precision & recall & f-measure \\
\hline
{\bf Ours} & {\bf 0.79} & {\bf 0.64} & {\bf 0.71}\\
\hline
Kim \cite{koo2013scene} & 0.78 & 0.65 & 0.71\\
\hline
TD-Mixture \cite{DBLP:conf/cvpr/Yao} & 0.69 & 0.66 & 0.67\\
\hline
Yi \cite{DBLP:journals/tip/YiT12} & 0.73 & 0.67 & 0.66\\
\hline
Epshtein \cite{DBLP:conf/cvpr/EpshteinOW10} & 0.73 & 0.60 & 0.66\\
\hline
Li \cite{DBLP:conf/icip/Li13} & 0.62 & 0.65 & 0.63\\
\hline
Yi \cite{DBLP:journals/tip/YiT11} & 0.71 & 0.62 & 0.62\\
\hline
Becker \cite{DBLP:conf/icdar/Lucas05} & 0.62 & 0.67 & 0.62\\
\hline
Meng \cite{DBLP:conf/das/MengS12} & 0.66 & 0.57 & 0.61\\
\hline
Li \cite{DBLP:conf/icpr/LiL12} & 0.59 & 0.59 & 0.59\\
\hline
Chen \cite{DBLP:conf/icdar/Lucas05} & 0.60 & 0.60 & 0.58\\
\hline
Neumann \cite{DBLP:conf/accv/NeumannM10} & 0.59 & 0.55 & 0.57\\
\hline
Zhang \cite{DBLP:conf/icpr/ZhangK10a} & 0.67 & 0.46 & 0.55\\
\hline
Ashida & 0.55 & 0.46 & 0.50\\
\hline
\end{tabular}}
\end{table}

\begin{table}
\centering
\caption{Results on ICDAR 2011 dataset.}
\label{tal:2011}
\scalebox{1.2}{
\begin{tabular}{ l||c|c|c }
\hline
method & precision & recall & f-measure \\
\hline
Kim~\cite{koo2013scene} & 0.81 & 0.69 & 0.75\\
\hline
{\bf Ours} & {\bf 0.80} & {\bf 0.62} & {\bf 0.70}\\
\hline
Neumann~\cite{DBLP:conf/cvpr/NeumannM10} & 0.73 & 0.65 & 0.69\\
\hline
Li \cite{DBLP:conf/icip/Li13} & 0.63 & 0.68 & 0.65\\
\hline
Yi & 0.67 & 0.58 & 0.62\\
\hline
TH-TextLoc & 0.67 & 0.58 & 0.62\\
\hline
Li \cite{DBLP:conf/icpr/LiL12} & 0.59 & 0.62 & 0.61\\
\hline
Neumann & 0.69 & 0.53 & 0.60\\
\hline
TDM\_IACS & 0.64 & 0.54 & 0.58\\
\hline
LIP6-Retin & 0.63 & 0.50 & 0.56\\
\hline
KAIST AIPR & 0.60 & 0.45 & 0.51\\
\hline
ECNU-CCG & 0.35 & 0.38 & 0.37\\
\hline
Text Hunter & 0.50 & 0.26 & 0.34\\
\hline
\end{tabular}}
\end{table}

\begin{figure*}[t]
\centering
\begin{tabular}{@{}c@{}c@{}c@{}c@{}c@{}c@{}c@{}c}
\includegraphics[width=24mm,height=18mm]{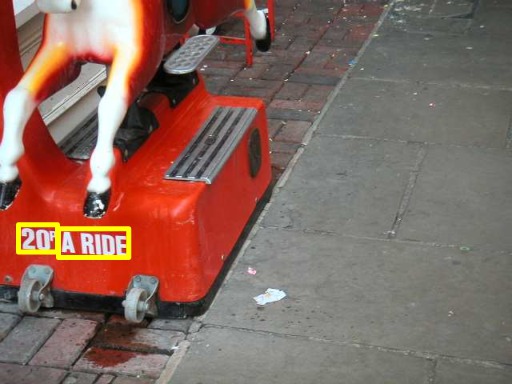} \ &
\includegraphics[width=24mm,height=18mm]{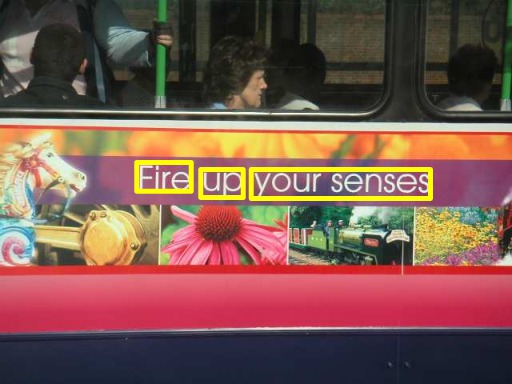} \ &
\includegraphics[width=24mm,height=18mm]{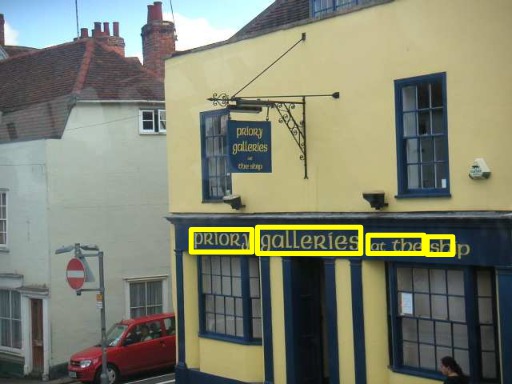} \ &
\includegraphics[width=24mm,height=18mm]{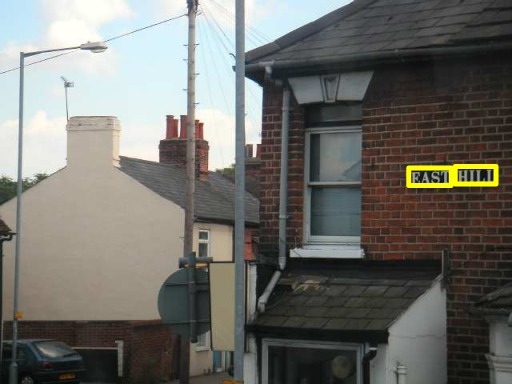} \ &
\includegraphics[width=24mm,height=18mm]{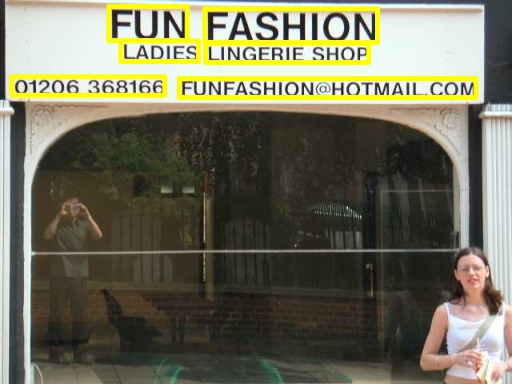} \ &
\includegraphics[width=24mm,height=18mm]{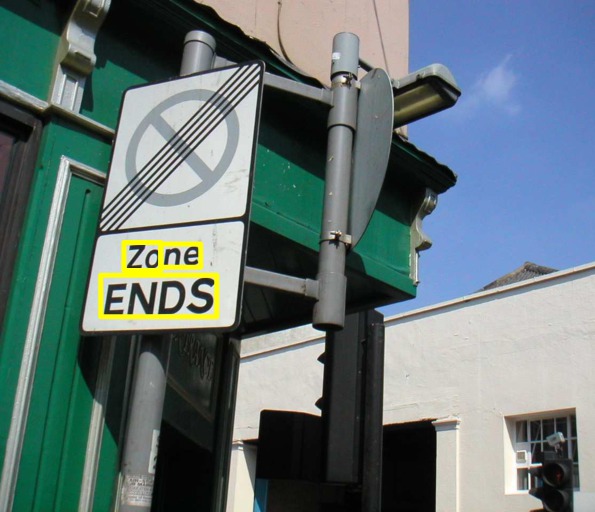} \ &
\includegraphics[width=24mm,height=18mm]{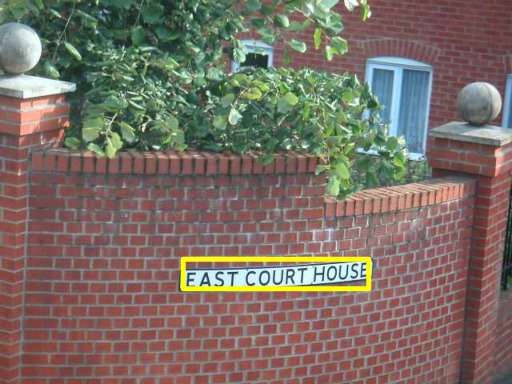} \ \\
\includegraphics[width=24mm,height=18mm]{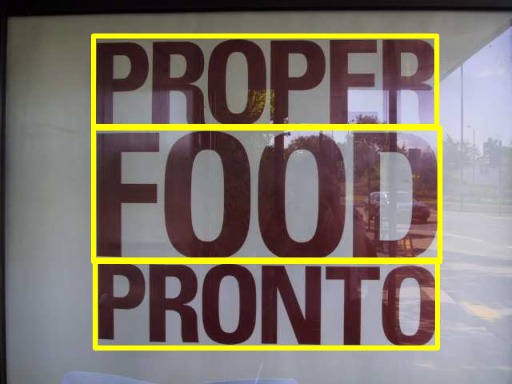} \ &
\includegraphics[width=24mm,height=18mm]{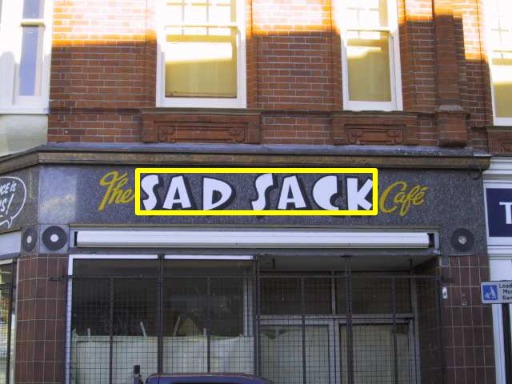} \ &
\includegraphics[width=24mm,height=18mm]{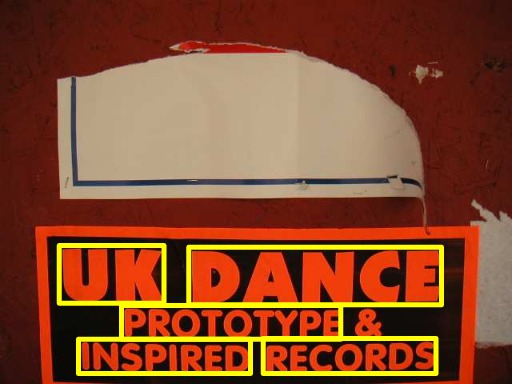} \ &
\includegraphics[width=24mm,height=18mm]{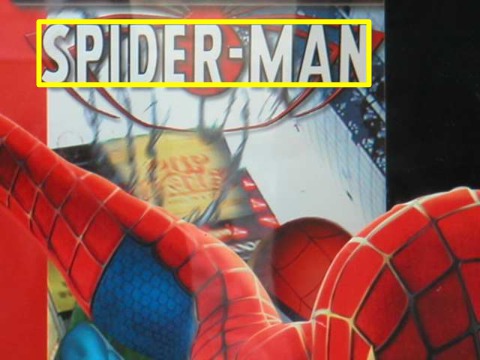} \ &
\includegraphics[width=24mm,height=18mm]{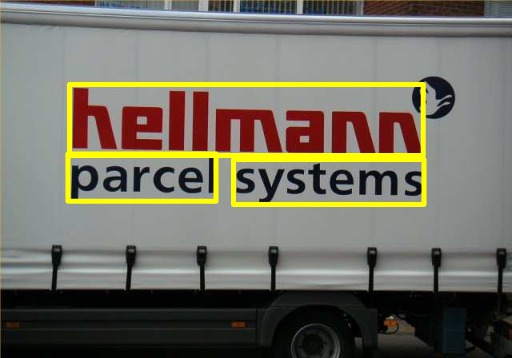} \ &
\includegraphics[width=24mm,height=18mm]{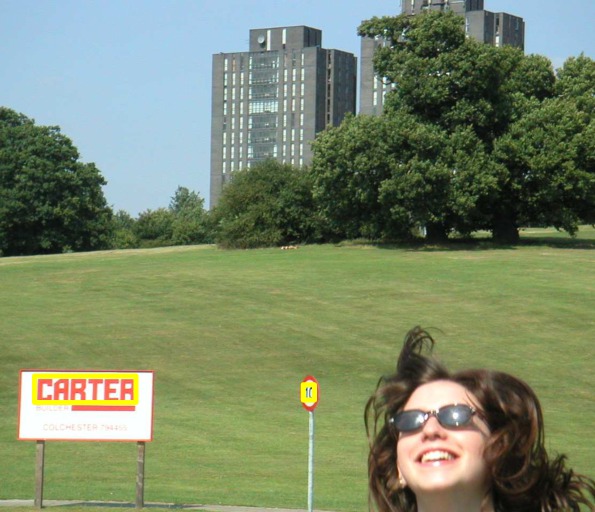} \ &
\includegraphics[width=24mm,height=18mm]{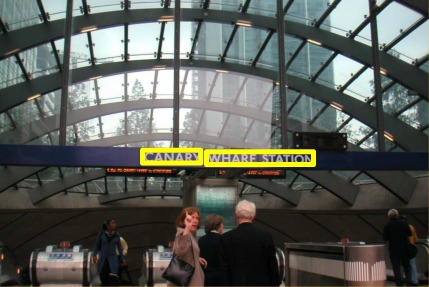} \ \\
\includegraphics[width=24mm,height=18mm]{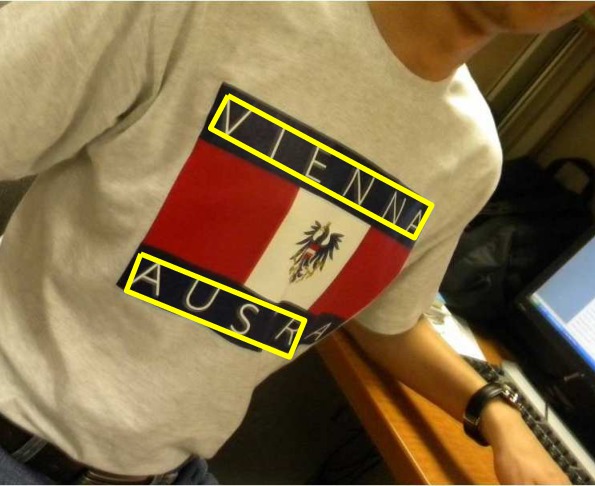} \ &
\includegraphics[width=24mm,height=18mm]{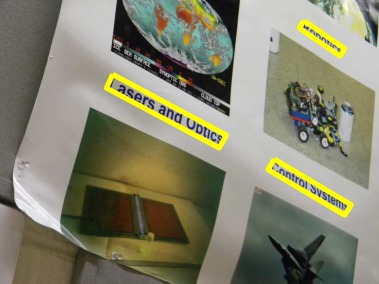} \ &
\includegraphics[width=24mm,height=18mm]{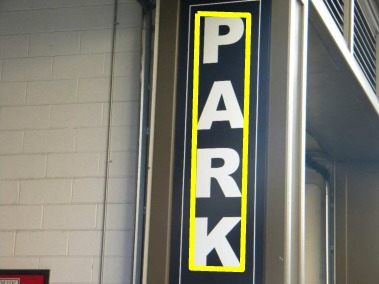} \ &
\includegraphics[width=24mm,height=18mm]{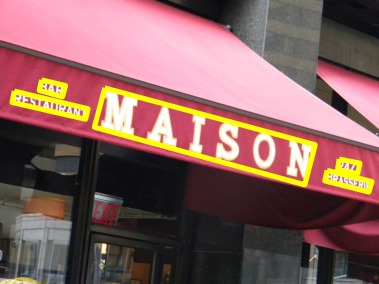} \ &
\includegraphics[width=24mm,height=18mm]{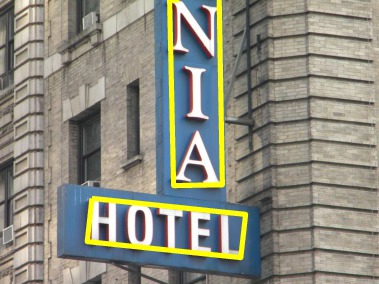} \ &
\includegraphics[width=24mm,height=18mm]{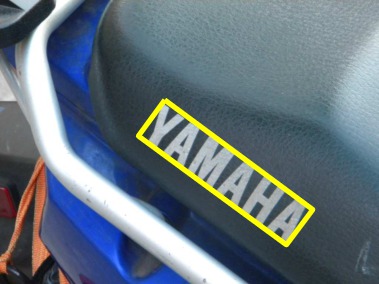} \ &
\includegraphics[width=24mm,height=18mm]{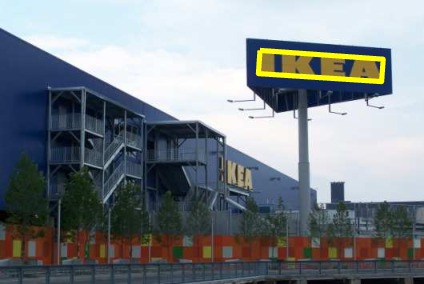} \ \\
\includegraphics[width=24mm,height=18mm]{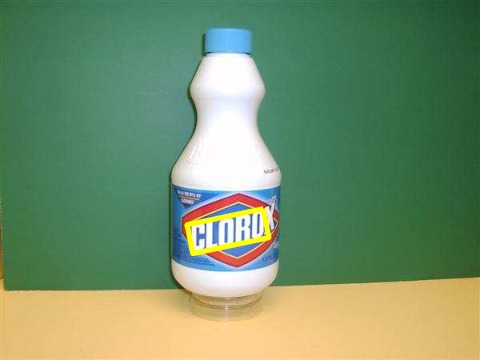} \ &
\includegraphics[width=24mm,height=18mm]{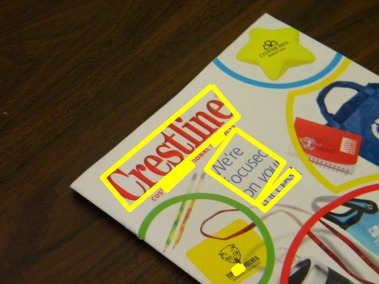} \ &
\includegraphics[width=24mm,height=18mm]{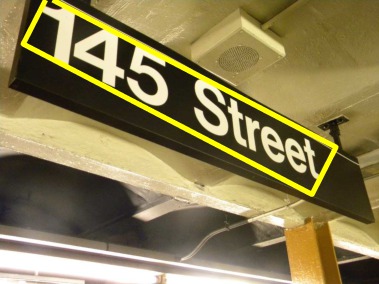} \ &
\includegraphics[width=24mm,height=18mm]{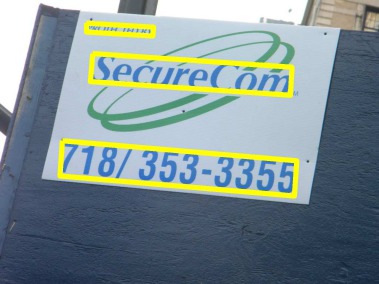} \ &
\includegraphics[width=24mm,height=18mm]{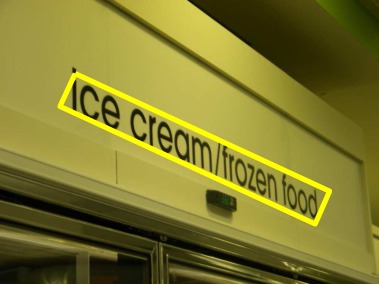} \ &
\includegraphics[width=24mm,height=18mm]{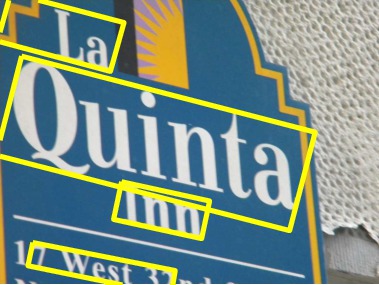} \ &
\includegraphics[width=24mm,height=18mm]{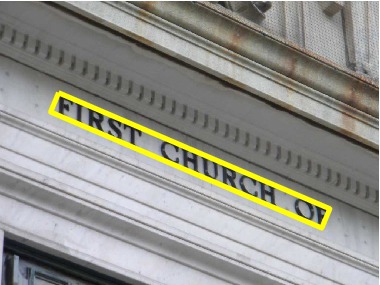} \ \\
\end{tabular}
\caption{Sample outputs of our method on the ICDAR datasets (top two rows) and OSTD dataset (bottom two rows). Detected text are in yellow rectangles. }
\label{fig:samples}
\end{figure*}

Table~\ref{tal:2003} and Table~\ref{tal:2011} show the performance of our approach on two benchmark datasets (\emph{i.e.} ICDAR 2003 and 2011), along with the
performance of other state-of-the-art scene text detection algorithms.
Note that methods without reference correspond to those presented in each competition.

On the ICDAR 2003 dataset, our method achieves significantly better precision (79\%) than all other approaches. Besides, our recall rate (64\%) is above the average, thus our f-measure (71\%) is superior than others.
Although supervised learning (random forest) is adopted in TD-Mixture~\cite{DBLP:conf/cvpr/Yao}, its precision (69\%) is much lower than
ours (79\%), which indicates the strong discriminability of the Bayesian classifier which is based on fusion of characterness cues.

On the ICDAR 2011 dataset, our method achieves a precision of 80\%, a recall of 62\%, and an f-measure of 70\%.
In terms of precision, our rate (80\%) is only 1\% lower than that of Kim's method~\cite{koo2013scene} (81\%) which is based on two times of supervised learning.
Besides, we report the best performance amongst all region-based approaches.

Our method achieves a precision of 72\%, a recall of 60\% and an f-measure of 61\% on the OSTD dataset~\cite{DBLP:journals/tip/YiT11} whcih outperforms Yi's method \cite{DBLP:journals/tip/YiT11}, with an improvement of 6\% in f-measure.

Fig.~\ref{fig:samples} shows some sample outputs of our method with detected text bounded by yellow rectangles.
As it shows, our method can handle several text variations, including color, orientation and size.
The proposed method also works well in a wide range of challenging conditions,
such as strong light, cluttered scenes, flexible surfaces and so forth.

\begin{figure}[t]
\centering
\begin{tabular}{@{}c@{}c@{}c}
\includegraphics[width=24mm,totalheight=18mm]{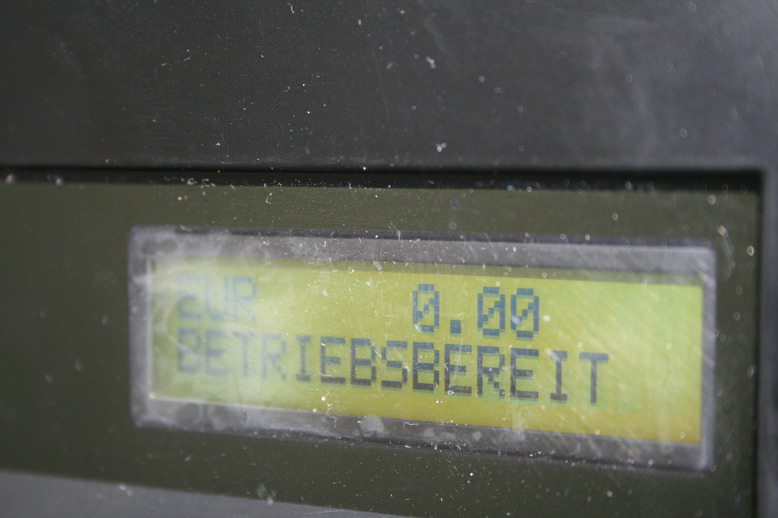} \ &
\includegraphics[width=24mm,height=18mm]{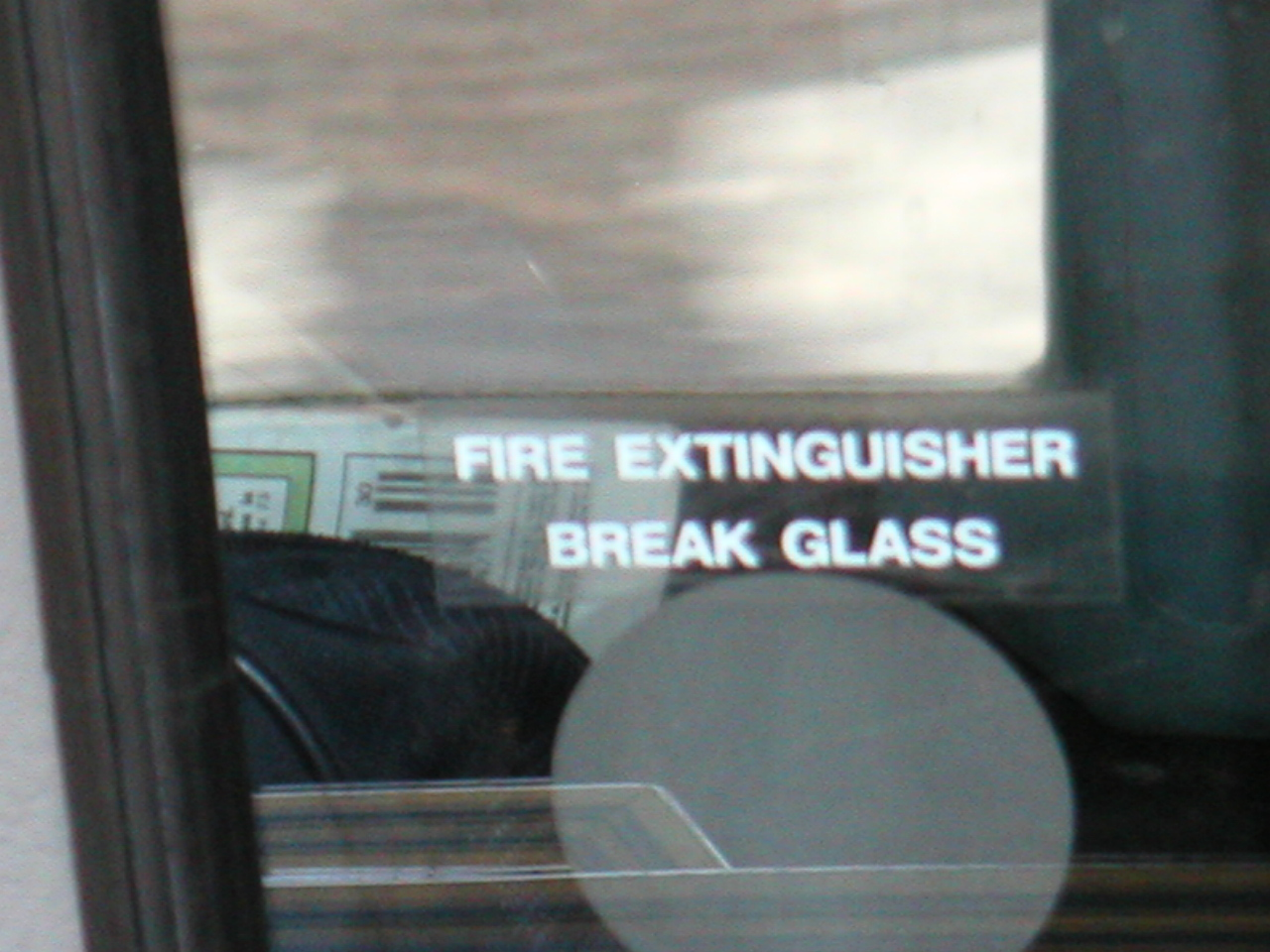} \ &
\includegraphics[width=24mm,totalheight=18mm]{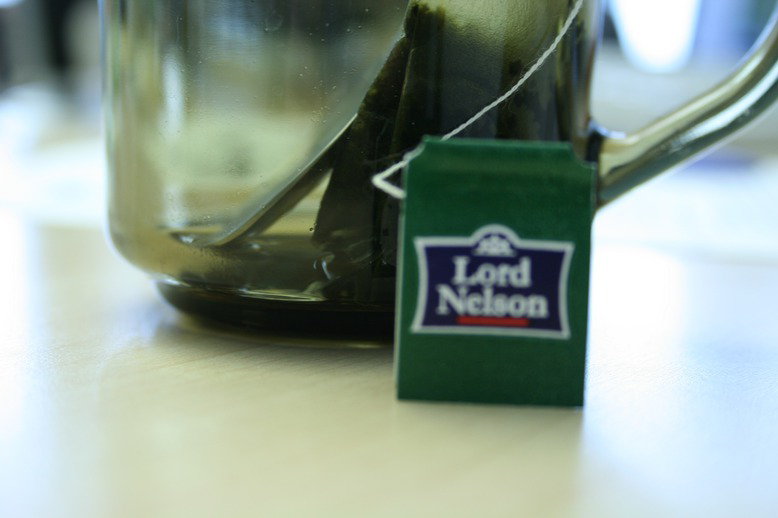} \ \\
\includegraphics[width=24mm,height=18mm]{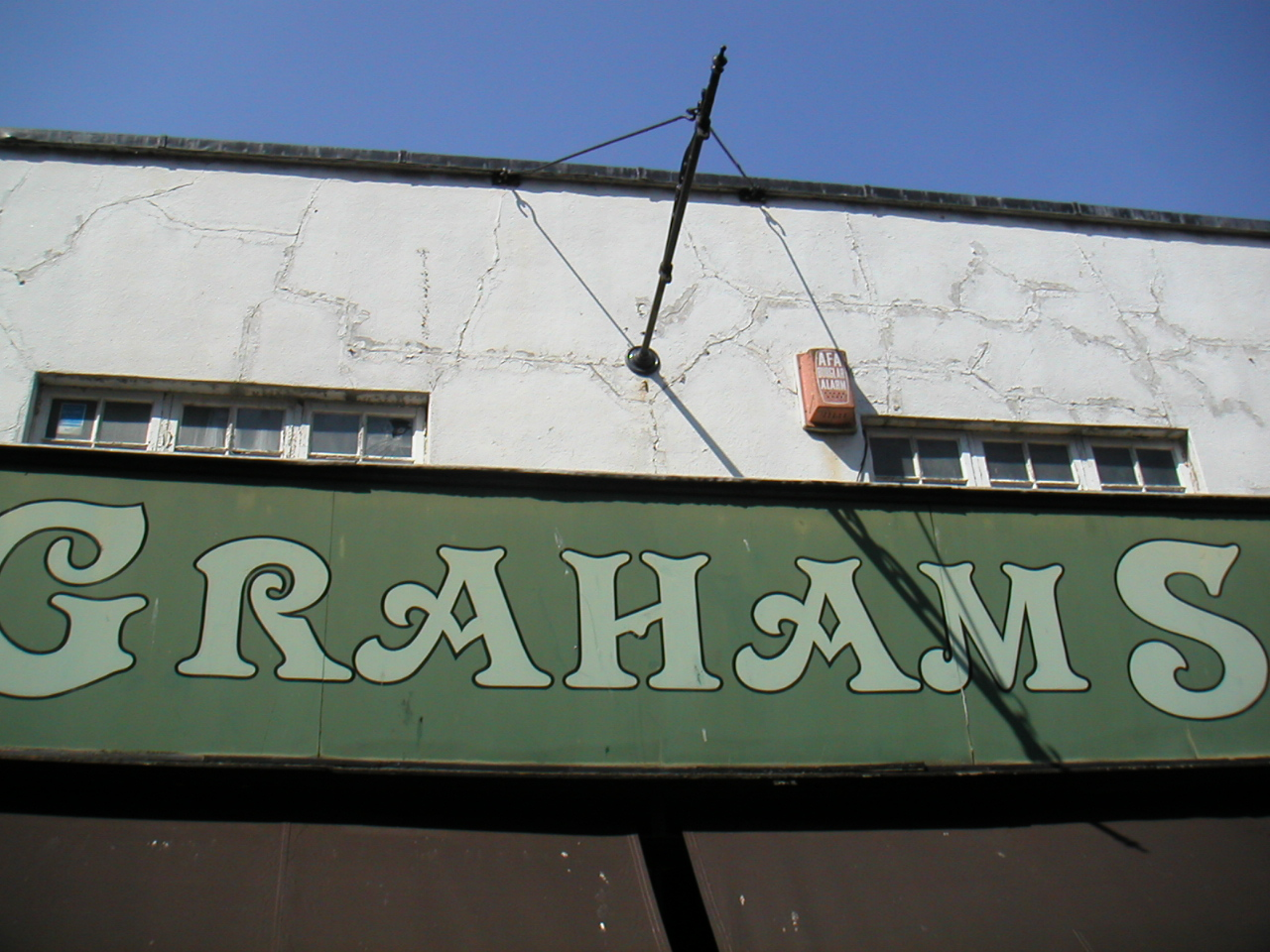} \ &
\includegraphics[width=24mm,height=18mm]{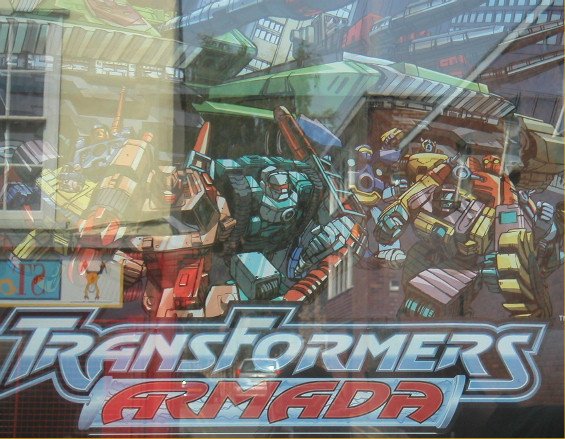} \ &
\includegraphics[width=24mm,height=18mm]{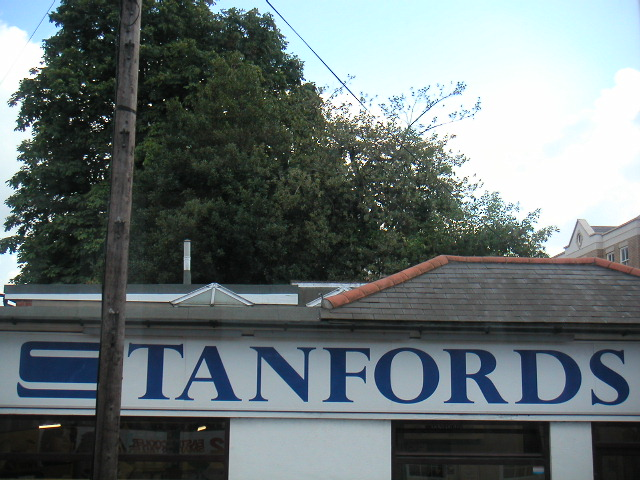} \ \\
\end{tabular}
\caption{False negatives of our approach. Clearly, we show two types
     of characters that our approach fails to detect:
    (i) characters in extremely blur and low resolutions (top row),
    (ii) characters in uncommon fonts (bottom row). }
\label{fig:undetected}
\end{figure}

In terms of failure cases (see Fig.~\ref{fig:undetected}), there are two culprits of false negatives.
Firstly, the candidate region extraction step misses some characters with very low resolution.
Furthermore, some characters in uncommon fonts are likely to have low characterness score, thus likely to be labeled as non-characters in the character labeling model.
This problem may be solved by enlarging the training sets to get more accurate distribution of characterness cues.
On the other hand, most false positives stem from non-characters whose distribution of cues is similar to that of characters.

\section{Conclusion}
In this work, we have proposed a scene text detection approach based
on measuring `characterness'.
The proposed characterness model reflects the probability of extracted regions belonging to character, which is constructed via fusion of novel characterness cues in the Bayesian framework.
We have demonstrated that this model significantly outperforms the state-of-the-art saliency detection approaches in the task of measuring the `characterness' of text.
In the character labeling model, by constructing a standard graph, not only characterness score of individual regions is considered, similarity between regions is also adopted as the pairwise potential.
Compared with state-of-the-art scene text detection approaches, we have shown that our method is able to achieve more accurate and robust results of scene text detection.

\section*{Acknowledgment}
This work is in part supported by
ARC Grants FT120100969 and LP130100156;
the UTS FEIT Industry and Innovation Project Scheme.
\ifCLASSOPTIONcaptionsoff
  \newpage
\fi
\bibliographystyle{IEEEtran}
\bibliography{IEEEabrv,strings}

\begin{thebibliography}{10}
\providecommand{\url}[1]{#1}
\csname url@samestyle\endcsname
\providecommand{\newblock}{\relax}
\providecommand{\bibinfo}[2]{#2}
\providecommand{\BIBentrySTDinterwordspacing}{\spaceskip=0pt\relax}
\providecommand{\BIBentryALTinterwordstretchfactor}{4}
\providecommand{\BIBentryALTinterwordspacing}{\spaceskip=\fontdimen2\font plus
\BIBentryALTinterwordstretchfactor\fontdimen3\font minus
  \fontdimen4\font\relax}
\providecommand{\BIBforeignlanguage}[2]{{%
\expandafter\ifx\csname l@#1\endcsname\relax
\typeout{** WARNING: IEEEtran.bst: No hyphenation pattern has been}%
\typeout{** loaded for the language `#1'. Using the pattern for}%
\typeout{** the default language instead.}%
\else
\language=\csname l@#1\endcsname
\fi
#2}}
\providecommand{\BIBdecl}{\relax}
\BIBdecl

\bibitem{itti1998model}
L.~Itti, C.~Koch, and E.~Niebur, ``A model of saliency-based visual attention
  for rapid scene analysis,'' \emph{{IEEE} Trans. Pattern Anal. Mach. Intell.},
  vol.~20, no.~11, pp. 1254--1259, 1998.

\bibitem{DBLP:conf/nips/BruceT05}
N.~D.~B. Bruce and J.~K. Tsotsos, ``Saliency based on information
  maximization,'' in \emph{Proc. Adv. Neural Inf. Process. Syst.}, 2005.

\bibitem{DBLP:conf/nips/HarelKP06}
J.~Harel, C.~Koch, and P.~Perona, ``Graph-based visual saliency,'' in
  \emph{Proc. Adv. Neural Inf. Process. Syst.}, 2006, pp. 545--552.

\bibitem{hou2007saliency}
X.~Hou and L.~Zhang, ``Saliency detection: A spectral residual approach,'' in
  \emph{Proc. IEEE Conf. Comp. Vis. Patt. Recogn.}, 2007, pp. 1--8.

\bibitem{DBLP:conf/iccv/JuddEDT09}
T.~Judd, K.~A. Ehinger, F.~Durand, and A.~Torralba, ``Learning to predict where
  humans look,'' in \emph{Proc. IEEE Int. Conf. Comp. Vis.}, 2009, pp.
  2106--2113.

\bibitem{DBLP:conf/mm/ZhaiS06}
Y.~Zhai and M.~Shah, ``Visual attention detection in video sequences using
  spatiotemporal cues,'' in \emph{ACM Multimedia}, 2006, pp. 815--824.

\bibitem{achanta2009frequency}
R.~Achanta, S.~Hemami, F.~Estrada, and S.~Susstrunk, ``Frequency-tuned salient
  region detection,'' in \emph{Proc. IEEE Conf. Comp. Vis. Patt. Recogn.},
  2009, pp. 1597--1604.

\bibitem{DBLP:conf/eccv/WeiWZ012}
Y.~Wei, F.~Wen, W.~Zhu, and J.~Sun, ``Geodesic saliency using background
  priors,'' in \emph{Proc. Eur. Conf. Comp. Vis.}, 2012, pp. 29--42.

\bibitem{DBLP:conf/iccv/FengWTZS11}
J.~Feng, Y.~Wei, L.~Tao, C.~Zhang, and J.~Sun, ``Salient object detection by
  composition,'' in \emph{Proc. IEEE Int. Conf. Comp. Vis.}, 2011, pp.
  1028--1035.

\bibitem{cheng2011global}
M.-M. Cheng, G.-X. Zhang, N.~J. Mitra, X.~Huang, and S.-M. Hu, ``Global
  contrast based salient region detection,'' in \emph{Proc. IEEE Conf. Comp.
  Vis. Patt. Recogn.}, 2011, pp. 409--416.

\bibitem{liu2011learning}
T.~Liu, Z.~Yuan, J.~Sun, J.~Wang, N.~Zheng, X.~Tang, and H.-Y. Shum, ``Learning
  to detect a salient object,'' \emph{{IEEE} Trans. Pattern Anal. Mach.
  Intell.}, vol.~33, no.~2, pp. 353--367, 2011.

\bibitem{DBLP:conf/iccv/KleinF11}
D.~A. Klein and S.~Frintrop, ``Center-surround divergence of feature statistics
  for salient object detection,'' in \emph{Proc. IEEE Int. Conf. Comp. Vis.},
  2011, pp. 2214--2219.

\bibitem{DBLP:conf/cvpr/AlexeDF10}
B.~Alexe, T.~Deselaers, and V.~Ferrari, ``What is an object?'' in \emph{Proc.
  IEEE Conf. Comp. Vis. Patt. Recogn.}, 2010, pp. 73--80.

\bibitem{perazzi2012saliency}
F.~Perazzi, P.~Krahenbuhl, Y.~Pritch, and A.~Hornung, ``Saliency filters:
  Contrast based filtering for salient region detection,'' in \emph{Proc. IEEE
  Conf. Comp. Vis. Patt. Recogn.}, 2012, pp. 733--740.

\bibitem{shen2012unified}
X.~Shen and Y.~Wu, ``A unified approach to salient object detection via low
  rank matrix recovery,'' in \emph{Proc. IEEE Conf. Comp. Vis. Patt. Recogn.},
  2012, pp. 853--860.

\bibitem{jiang2011automatic}
H.~Jiang, J.~Wang, Z.~Yuan, T.~Liu, N.~Zheng, and S.~Li, ``Automatic salient
  object segmentation based on context and shape prior,'' in \emph{Proc. Brit.
  Mach. Vis. Conf.}, vol.~3, no.~4, 2011, p.~7.

\bibitem{DBLP:conf/cvpr/GofermanZT10}
S.~Goferman, L.~Zelnik-Manor, and A.~Tal, ``Context-aware saliency detection,''
  in \emph{Proc. IEEE Conf. Comp. Vis. Patt. Recogn.}, 2010, pp. 2376--2383.

\bibitem{DBLP:conf/iccv/ChangLCL11}
K.-Y. Chang, T.-L. Liu, H.-T. Chen, and S.-H. Lai, ``Fusing generic objectness
  and visual saliency for salient object detection,'' in \emph{Proc. IEEE Int.
  Conf. Comp. Vis.}, 2011, pp. 914--921.

\bibitem{DBLP:conf/eccv/RahtuKSH10}
E.~Rahtu, J.~Kannala, M.~Salo, and J.~Heikkil{\"a}, ``Segmenting salient
  objects from images and videos,'' in \emph{Proc. Eur. Conf. Comp. Vis.},
  2010, pp. 366--379.

\bibitem{DBLP:conf/cvpr/DingXY11}
Y.~Ding, J.~Xiao, and J.~Yu, ``Importance filtering for image retargeting,'' in
  \emph{Proc. IEEE Conf. Comp. Vis. Patt. Recogn.}, 2011, pp. 89--96.

\bibitem{sun2011scale}
J.~Sun and H.~Ling, ``Scale and object aware image retargeting for thumbnail
  browsing,'' in \emph{Proc. IEEE Int. Conf. Comp. Vis.}, 2011, pp. 1511--1518.

\bibitem{DBLP:conf/cvpr/SharmaJS12}
G.~Sharma, F.~Jurie, and C.~Schmid, ``Discriminative spatial saliency for image
  classification,'' in \emph{Proc. IEEE Conf. Comp. Vis. Patt. Recogn.}, 2012,
  pp. 3506--3513.

\bibitem{DBLP:conf/iccv/WangXZH11}
L.~Wang, J.~Xue, N.~Zheng, and G.~Hua, ``Automatic salient object extraction
  with contextual cue,'' in \emph{Proc. IEEE Int. Conf. Comp. Vis.}, 2011, pp.
  105--112.

\bibitem{cerf2009faces}
M.~Cerf, E.~P. Frady, and C.~Koch, ``Faces and text attract gaze independent of
  the task: Experimental data and computer model,'' \emph{J. Vis.}, vol.~9,
  no.~12, 2009.

\bibitem{DBLP:conf/icpr/SunLS10}
Q.~Sun, Y.~Lu, and S.~Sun, ``A visual attention based approach to text
  extraction,'' in \emph{Proc. IEEE Int. Conf. Patt. Recogn.}, 2010, pp.
  3991--3995.

\bibitem{DBLP:conf/das/ShahabSDU12}
A.~Shahab, F.~Shafait, A.~Dengel, and S.~Uchida, ``How salient is scene text?''
  in \emph{Proc. IEEE Int. Workshop. Doc. Anal. Syst.}, 2012, pp. 317--321.

\bibitem{DBLP:conf/das/MengS12}
Q.~Meng and Y.~Song, ``Text detection in natural scenes with salient region,''
  in \emph{Proc. IEEE Int. Workshop. Doc. Anal. Syst.}, 2012, pp. 384--388.

\bibitem{DBLP:conf/icdar/UchidaSKF11}
S.~Uchida, Y.~Shigeyoshi, Y.~Kunishige, and Y.~Feng, ``A keypoint-based
  approach toward scenery character detection,'' in \emph{Proc. IEEE Int. Conf.
  Doc. Anal. and Recogn.}, 2011, pp. 819--823.

\bibitem{DBLP:journals/ijcv/HoiemEH11}
D.~Hoiem, A.~A. Efros, and M.~Hebert, ``Recovering occlusion boundaries from an
  image,'' \emph{Int. J. Comp. Vis.}, vol.~91, no.~3, pp. 328--346, 2011.

\bibitem{DBLP:conf/cvpr/ArbelaezHGGBM12}
P.~Arbelaez, B.~Hariharan, C.~Gu, S.~Gupta, L.~D. Bourdev, and J.~Malik,
  ``Semantic segmentation using regions and parts,'' in \emph{Proc. IEEE Conf.
  Comp. Vis. Patt. Recogn.}, 2012.

\bibitem{DBLP:conf/cvpr/PrestLCSF12}
A.~Prest, C.~Leistner, J.~Civera, C.~Schmid, and V.~Ferrari, ``Learning object
  class detectors from weakly annotated video,'' in \emph{Proc. IEEE Conf.
  Comp. Vis. Patt. Recogn.}, 2012, pp. 3282--3289.

\bibitem{DBLP:journals/pami/BoykovVZ01}
Y.~Boykov, O.~Veksler, and R.~Zabih, ``Fast approximate energy minimization via
  graph cuts,'' \emph{{IEEE} Trans. Pattern Anal. Mach. Intell.}, vol.~23,
  no.~11, pp. 1222--1239, 2001.

\bibitem{DBLP:conf/icdar/LeeLLYK11}
J.-J. Lee, P.-H. Lee, S.-W. Lee, A.~L. Yuille, and C.~Koch, ``Adaboost for text
  detection in natural scene,'' in \emph{Proc. IEEE Int. Conf. Doc. Anal. and
  Recogn.}, 2011, pp. 429--434.

\bibitem{DBLP:conf/cvpr/ChenY04}
X.~Chen and A.~L. Yuille, ``Detecting and reading text in natural scenes,'' in
  \emph{Proc. IEEE Conf. Comp. Vis. Patt. Recogn.}, 2004, pp. 366--373.

\bibitem{DBLP:conf/cvpr/EpshteinOW10}
B.~Epshtein, E.~Ofek, and Y.~Wexler, ``Detecting text in natural scenes with
  stroke width transform,'' in \emph{Proc. IEEE Conf. Comp. Vis. Patt.
  Recogn.}, 2010, pp. 2963--2970.

\bibitem{DBLP:conf/icpr/ZhangK10a}
J.~Zhang and R.~Kasturi, ``Text detection using edge gradient and graph
  spectrum,'' in \emph{Proc. IEEE Int. Conf. Patt. Recogn.}, 2010, pp.
  3979--3982.

\bibitem{DBLP:conf/accv/ZhangK10}
------, ``Character energy and link energy-based text extraction in scene
  images,'' in \emph{Proc. Asian Conf. Comp. Vis.}, 2010, pp. 308--320.

\bibitem{DBLP:journals/tip/YiT11}
C.~Yi and Y.~Tian, ``Text string detection from natural scenes by
  structure-based partition and grouping,'' \emph{{IEEE} Trans. Image Proc.},
  vol.~20, no.~9, pp. 2594--2605, 2011.

\bibitem{DBLP:conf/icip/ChenTSCGG11}
H.~Chen, S.~S. Tsai, G.~Schroth, D.~M. Chen, R.~Grzeszczuk, and B.~Girod,
  ``Robust text detection in natural images with edge-enhanced maximally stable
  extremal regions,'' in \emph{Proc. IEEE Int. Conf. Image Process.}, 2011, pp.
  2609--2612.

\bibitem{DBLP:conf/icpr/LiL12}
Y.~Li and H.~Lu, ``Scene text detection via stroke width,'' in \emph{Proc. IEEE
  Int. Conf. Patt. Recogn.}, 2012, pp. 681--684.

\bibitem{DBLP:conf/icip/Li13}
Y.~Li, C.~Shen, W.~Jia, and A.~van~den Hengel, ``Leveraging surrounding context
  for scene text detection,'' in \emph{Proc. IEEE Int. Conf. Image Process.},
  2013.

\bibitem{DBLP:conf/accv/NeumannM10}
L.~Neumann and J.~Matas, ``A method for text localization and recognition in
  real-world images,'' in \emph{Proc. Asian Conf. Comp. Vis.}, 2010, pp.
  770--783.

\bibitem{DBLP:conf/cvpr/NeumannM10}
------, ``Real-time scene text localization and recognition,'' in \emph{Proc.
  IEEE Conf. Comp. Vis. Patt. Recogn.}, 2012, pp. 3538--3545.

\bibitem{DBLP:journals/tip/PanHL11}
Y.-F. Pan, X.~Hou, and C.-L. Liu, ``A hybrid approach to detect and localize
  texts in natural scene images,'' \emph{{IEEE} Trans. Image Proc.}, vol.~20,
  no.~3, pp. 800--813, 2011.

\bibitem{DBLP:conf/cvpr/Yao}
C.~Yao, X.~Bai, W.~Liu, Y.~Ma, and Z.~Tu, ``Detecting texts of arbitrary
  orientations in natural images,'' in \emph{Proc. IEEE Conf. Comp. Vis. Patt.
  Recogn.}, 2012, pp. 1083--1090.

\bibitem{DBLP:journals/tip/YiT12}
C.~Yi and Y.~Tian, ``Localizing text in scene images by boundary clustering,
  stroke segmentation, and string fragment classification,'' \emph{{IEEE}
  Trans. Image Proc.}, vol.~21, no.~9, pp. 4256--4268, 2012.

\bibitem{koo2013scene}
H.~Koo and D.~Kim, ``Scene text detection via connected component clustering
  and non-text filtering.'' \emph{{IEEE} Trans. Image Proc.}, vol.~22, no.~6,
  pp. 2296--2305, 2013.

\bibitem{DBLP:conf/bmvc/MatasCUP02}
J.~Matas, O.~Chum, M.~Urban, and T.~Pajdla, ``Robust wide baseline stereo from
  maximally stable extremal regions,'' in \emph{Proc. Brit. Mach. Vis. Conf.},
  2002, pp. 384--393.

\bibitem{DBLP:conf/cvpr/DonoserB06}
M.~Donoser and H.~Bischof, ``Efficient maximally stable extremal region (mser)
  tracking,'' in \emph{Proc. IEEE Conf. Comp. Vis. Patt. Recogn.}, 2006, pp.
  553--560.

\bibitem{DBLP:conf/iccv/ForssenL07}
P.-E. Forss{\'e}n and D.~G. Lowe, ``Shape descriptors for maximally stable
  extremal regions,'' in \emph{Proc. IEEE Int. Conf. Comp. Vis.}, 2007, pp.
  1--8.

\bibitem{DBLP:conf/icdar/NeumannM11}
L.~Neumann and J.~Matas, ``Text localization in real-world images using
  efficiently pruned exhaustive search,'' in \emph{Proc. IEEE Int. Conf. Doc.
  Anal. and Recogn.}, 2011, pp. 687--691.

\bibitem{DBLP:conf/accv/Tsai}
S.~Tsai, V.~Parameswaran, J.~Berclaz, R.~Vedantham, R.~Grzeszczuk, and
  B.~Girod, ``Design of a text detection system via hypothesis generation and
  verification,'' in \emph{Proc. Asian Conf. Comp. Vis.}, 2012.

\bibitem{DBLP:journals/ijcv/MikolajczykTSZMSKG05}
K.~Mikolajczyk, T.~Tuytelaars, C.~Schmid, A.~Zisserman, J.~Matas,
  F.~Schaffalitzky, T.~Kadir, and L.~J.~V. Gool, ``A comparison of affine
  region detectors,'' \emph{Int. J. Comp. Vis.}, vol.~65, no. 1-2, pp. 43--72,
  2005.

\bibitem{DBLP:conf/eccv/HeST10}
K.~He, J.~Sun, and X.~Tang, ``Guided image filtering,'' in \emph{Proc. Eur.
  Conf. Comp. Vis.}, 2010, pp. 1--14.

\bibitem{DBLP:conf/bmvc/Ali12}
N.~B. Ali Mosleh~and and A.~B. Hamza, ``Image text detection using a
  bandlet-based edge detector and stroke width transform,'' in \emph{Proc.
  Brit. Mach. Vis. Conf.}, 2012, pp. 1--12.

\bibitem{DBLP:conf/accv/Pan12}
J.~Pan, Y.~Chen, B.~Anderson, P.~Berkhin, and T.~Kanade, ``Effectively
  leveraging visual context to detect texts in natural scenes,'' in \emph{Proc.
  Asian Conf. Comp. Vis.}, 2012.

\bibitem{DBLP:conf/icpr/LiuS08a}
Z.~Liu and S.~Sarkar, ``Robust outdoor text detection using text intensity and
  shape features,'' in \emph{Proc. IEEE Int. Conf. Patt. Recogn.}, 2008, pp.
  1--4.

\bibitem{DBLP:conf/cvpr/DalalT05}
N.~Dalal and B.~Triggs, ``Histograms of oriented gradients for human
  detection,'' in \emph{Proc. IEEE Conf. Comp. Vis. Patt. Recogn.}, 2005, pp.
  886--893.

\bibitem{DBLP:conf/iccv/BoykovJ01}
Y.~Boykov and M.-P. Jolly, ``Interactive graph cuts for optimal boundary and
  region segmentation of objects in n-d images,'' in \emph{Proc. IEEE Int.
  Conf. Comp. Vis.}, 2001, pp. 105--112.

\bibitem{DBLP:conf/cvpr/KuttelF12}
D.~K{\"u}ttel and V.~Ferrari, ``Figure-ground segmentation by transferring
  window masks,'' in \emph{Proc. IEEE Conf. Comp. Vis. Patt. Recogn.}, 2012,
  pp. 558--565.

\bibitem{DBLP:conf/eccv/BorjiSI12}
A.~Borji, D.~N. Sihite, and L.~Itti, ``Salient object detection: A benchmark,''
  in \emph{Proc. Eur. Conf. Comp. Vis.}, 2012, pp. 414--429.

\bibitem{RosenfeldW11}
A.~Rosenfeld and D.~Weinshall, ``Extracting foreground masks towards object
  recognition,'' in \emph{Proc. IEEE Int. Conf. Comp. Vis.}, 2011, pp.
  1371--1378.

\bibitem{DBLP:conf/icdar/LucasPSTWY03}
S.~M. Lucas, A.~Panaretos, L.~Sosa, A.~Tang, S.~Wong, and R.~Young, ``Icdar
  2003 robust reading competitions,'' in \emph{Proc. IEEE Int. Conf. Doc. Anal.
  and Recogn.}, 2003, pp. 682--687.

\bibitem{DBLP:conf/icdar/ShahabSD11a}
A.~Shahab, F.~Shafait, and A.~Dengel, ``Icdar 2011 robust reading competition
  challenge 2: Reading text in scene images,'' in \emph{Proc. IEEE Int. Conf.
  Doc. Anal. and Recogn.}, 2011, pp. 1491--1496.

\bibitem{DBLP:conf/icdar/Lucas05}
S.~M. Lucas, ``Text locating competition results,'' in \emph{Proc. IEEE Int.
  Conf. Doc. Anal. and Recogn.}, 2005, pp. 80--85.

\bibitem{DBLP:journals/ijdar/WolfJ06}
C.~Wolf and J.-M. Jolion, ``Object count/area graphs for the evaluation of
  object detection and segmentation algorithms,'' \emph{Int. J. Doc. Anal.
  Recogn.}, vol.~8, no.~4, pp. 280--296, 2006.

\end{thebibliography}
\end{document}